\RequirePackage{fix-cm}
\documentclass[twocolumn]{svjour3}
\smartqed
\usepackage{bm}
\usepackage{graphicx}
\usepackage{newtxtext, newtxmath}
\usepackage{natbib}
\usepackage{balance}
\usepackage{amsmath}
\usepackage[colorlinks=true,citecolor=blue]{hyperref}
\usepackage{xcolor}
\usepackage{latexsym}
\usepackage{makecell}
\usepackage{diagbox}
\usepackage{adjustbox}

\usepackage{caption}
\usepackage{subcaption}
\usepackage{soul}
\usepackage{booktabs}
\usepackage{multirow}
\usepackage{stfloats}
\usepackage{threeparttable}
\newcommand{\revise}[1]{\textcolor{black}{#1}}

\usepackage{url}
\usepackage{array}
\usepackage{ragged2e}
\usepackage{enumitem}
\usepackage{microtype}
\PassOptionsToPackage{hyphens}{url}\usepackage{hyperref}

\hypersetup{
urlcolor=blue
}

\usepackage{graphicx}
\usepackage{algorithm, algorithmic}
\usepackage{xcolor}
\usepackage{colortbl}
\usepackage{xurl}

\renewcommand{\and}{\unskip\hspace{4em}}

\journalname{International Journal of Computer Vision}

\begin{document}

\title{Optimizing Multi-Modality Trackers via Significance-Regularized Tuning}

\author{Zhiwen Chen, Jinjian Wu, Zhiyu Zhu, Yifan Zhang, Guangming Shi, Junhui Hou}

\institute{
This work was supported in part by the NSFC Excellent Young Scientists Fund 62422118, in part by the Hong Kong RGC under Grants RFS2627-1S02, R1018-25F, 11219324, and 11219422, in part by the Hong Kong ITC under Grant ITS/164/23, and in part by Shanghai Pujiang Program (25PJA042);\\
\and
Zhiwen Chen and Junhui Hou are with the Department of Computer Science, City University of Hong Kong, Hong Kong SAR, China. \email{ zhiwen.chen@cityu.edu.hk; jh.hou@cityu.edu.hk}\\
\and
Jinjian Wu and Guangming Shi are with the School of Artificial Intelligence, Xidian University, China. \email{jinjian.wu@mail.xidian.edu.cn, gmshi@xidian.edu.cn;}\\
\and
Zhiyu Zhu is with City University of Hong Kong (Dongguan), Dongguan, China, and also with the Department of Computer Science, City University of Hong Kong, Hong Kong SAR, China. \email{zhiyuzhu2-c@my.cityu.edu.hk}\\
\and
Yifan Zhang is with the School of Mechatronic Engineering and Automation, Shanghai University, Shanghai, China, and also with the Department of Computer Science, City University of Hong Kong. \email{yfzhang@shu.edu.cn;}\\
\and 
Corresponding authors: Jinjian Wu and Zhiyu Zhu
}

\date{Received: date / Accepted: date}

\maketitle
\begin{abstract}
This paper tackles the critical challenge of optimizing multi-modality trackers by adapting RGB-pre-trained trackers. Existing fine-tuning paradigms oscillate between excessive flexibility and over-restriction, leading to suboptimal stability--plasticity trade-offs. To mitigate this dilemma, we propose a novel significance-regularized fine-tuning framework (SRFT), which delicately refines parameter updates by incorporating parameter significance. Through a comprehensive investigation of the transition from pre-trained to multi-modality contexts, we identify that parameters associated with
source-knowledge preservation and target-domain adaptation are the primary drivers of this issue. Specifically, we first \revise{characterize the local curvature of the source
tracking objective around the} pre-trained weights to measure prior significance for preserving pre-trained knowledge. Subsequently, we estimate transfer significance to capture target-domain adaptation demands during fine-tuning. By incorporating these complementary significance terms as unified regularization, SRFT markedly enhances transferability across modalities. Extensive experiments demonstrate the
effectiveness of SRFT across various multi-modality tracking benchmarks, with controlled ablations further validating the contribution
of significance regularization. The source code and models are publicly available at \url{https://github.com/zhiwen-xdu/SRTrack}.

\keywords{Multi-modality tracking \and Cross-modal transfer \and Parameter significance \and Regularized tuning}
\end{abstract}

\begin{figure}[t]
\centering
\centerline{\includegraphics[width=0.48\textwidth]{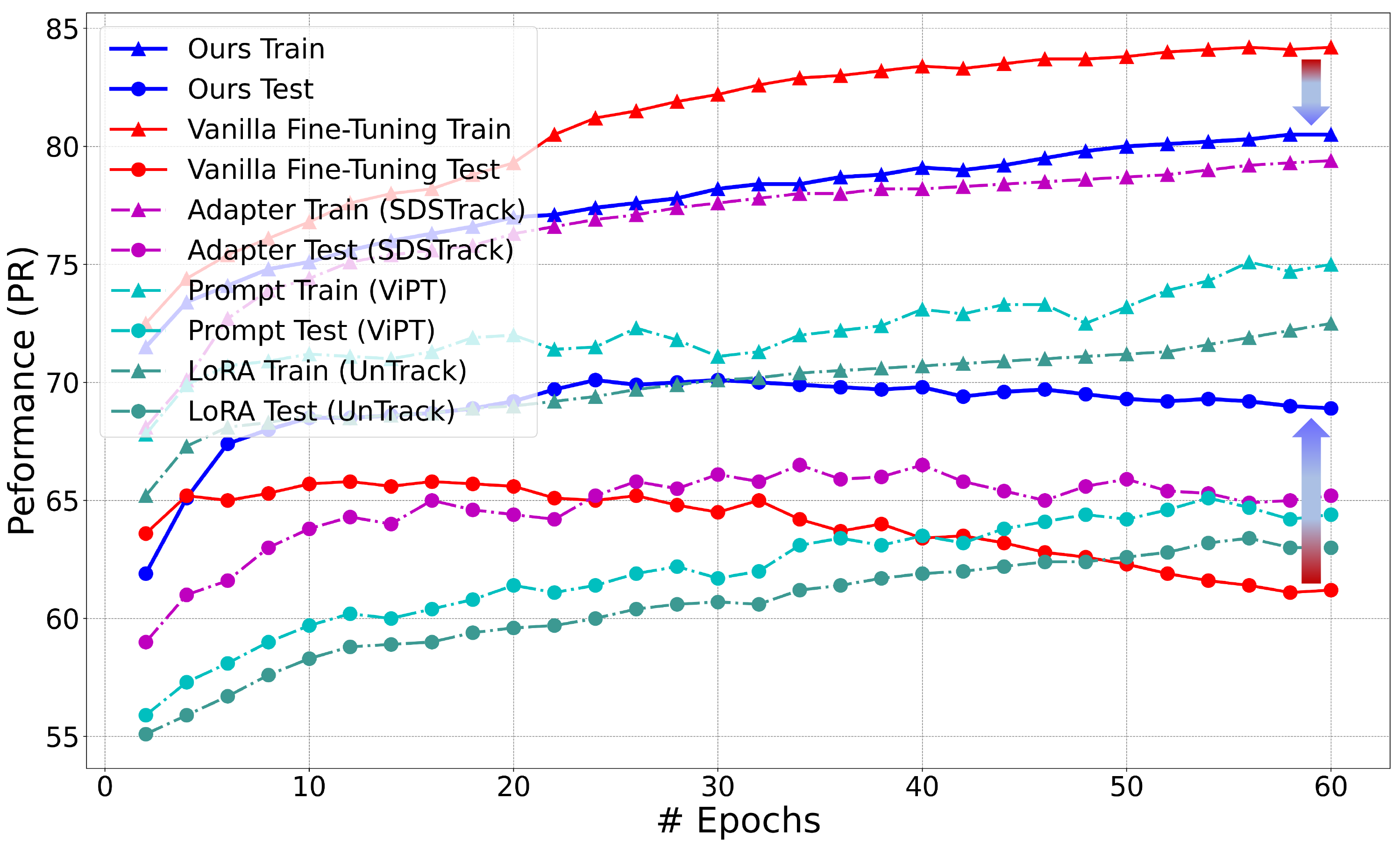}}
\vskip -0.1in
\caption{\textbf{Optimization trajectory analysis on LasHeR.} This plot contrasts the training and testing dynamics of different tuning paradigms. As visualized by the arrows, our method effectively mitigates the misfitting issue and enhances multi-modality trackers with superior generalization and stability.}
\label{fig:training-trajectory}
\vskip -0.1in
\end{figure}

\section{Introduction} \label{sec1}
Object tracking, a fundamental task of visual perception, has witnessed remarkable advancements over the past decades~\citep{hong2024onetracker,xie2024diffusiontrack,zheng2024nettrack,cai2024hiptrack}. Despite the promising results, RGB-based trackers often struggle under challenging conditions, such as extreme illumination, motion blur, and occlusions. Therefore, multi-modality tracking with more comprehensive sensory signals (e.g., event, depth, thermal) has garnered growing interest. With the popularity of the data-driven methods in the object tracking community, both training data scale and model capacity have grown substantially in recent years~\citep{ye2022joint,lin2022swintrack,chen2023seqtrack,chen2025sutrack}. A prevailing paradigm is to adapt trackers pre-trained on large-scale RGB
datasets to downstream auxiliary modalities through cross-modal fine-tuning
or transfer learning, thereby leveraging well-structured pre-trained
representations to improve performance and accelerate convergence.


Concretely, some existing approaches have explored the \textit{full fine-tuning} (FFT) paradigm \citep{wang2023visevent,zhu2023cross,Sun_2025_CVPR}, where multi-modality trackers are initialized with RGB-based weights and subsequently optimized for task-specific objectives. Nevertheless, while FFT offers maximum flexibility and swiftly adapts to the target domain, substantial domain gaps, coupled with limited scale of auxiliary modalities, hinder the retention of pre-trained knowledge during transfer, often inducing severe \textbf{overfitting (i.e., a widened train-test gap and deteriorating test performance)}. In contrast to full fine-tuning, recent research has shifted toward \textit{parameter efficient fine-tuning} (PEFT) \citep{hu2021lora,jia2022visual,chen2022adaptformer}. PEFT keeps the majority of pre-trained parameters frozen, updating only a small fraction of modality-specific ones to retain prior knowledge. Several methods fall under this   umbrella~\citep{zhu2023visual,hou2024sdstrack,wu2024single}, including prompt tuning, visual adapter, etc. Although effective, PEFT-based methods impose rigid constraints on the primary model weights, resulting in \textbf{underfitting (i.e., a restricted upper bound on training performance)} when handling the large distribution drifts. In summary, existing fine-tuning paradigms fluctuate between excessive flexibility and over-restriction, both contributing to \textbf{misfitting} (i.e., overfitting and underfitting) and a sub-optimal stability--plasticity trade-off between pre-trained knowledge and downstream adaptation (as clearly illustrated in Fig.~\ref{fig:training-trajectory}).


In this work, we endeavor to mitigate the misfitting dilemma in cross-modal tracker adaptation by strategically regularizing the learning process. By analyzing parameter responses spanning from pre-training to fine-tuning phases, we identify that pronounced parameter significance, in its various aspects, is the key factor of degraded prior generalizability and transfer adaptability. To mitigate this, we propose a \textit{significance-regularized fine-tuning} (SRFT) framework that adaptively regulates parameter updates to boost model transfer with precision. Specifically, we optimize multi-modality trackers from the following perspectives. 
\begin{enumerate}
\item \textbf{\textit{Formulating Prior Significance}}. We commence by investigating the \revise{local source-loss curvature} around pre-trained trackers, utilizing it as a critical indicator to characterize prior generalizability for downstream fine-tuning. We then estimate the corresponding operation-level prior significance through Rayleigh-quotient probing of dominant curvature responses.
\item \textbf{\textit{Modeling Transfer Significance}}. We analyze the sparse and heterogeneous transfer gradients arising during
downstream adaptation and formulate parameter-wise transfer significance
from readily available gradient signals to capture target-domain adaptation
demands and guide adaptive parameter updates.
\item \textbf{\textit{Significance Regularized Tuning}}. We integrate prior and transfer significance to adaptively modulate parameter
updates, balancing source-knowledge preservation and target-domain adaptation
throughout fine-tuning. This mechanism facilitates seamless fine-tuning across various multi-modality tracking tasks, continuously enhancing the model during the training phase.
\end{enumerate}

Our method strategically guides the cross-domain fine-tuning process to optimize downstream multi-modality tracking tasks. Extensive experimental results showcase our method achieves new state-of-the-art results across three multi-modality tracking tasks (RGB-Event, RGB-Depth, RGB-Thermal) and seven benchmarks, spanning diverse pre-trained tracking models. Comprehensive ablation studies and parameter significance measurements confirm the effectiveness of the significance-aware regularization fine-tuning strategy. In summary, the main contributions of this paper are: 
\begin{itemize}
\item We revisit the misfitting issue arising when adapting RGB-pre-trained
trackers to multi-modality tracking tasks and propose SRFT, a novel training-time regularization framework to indicate better transferability, which is \revise{complementary} to the existing FFT and PEFT configurations; 
\item we formulate the prior significance \revise{from source-loss curvature and transfer
significance from target-conditioned adaptation signals}, and integrate them to
guide significance-aware parameter updates, thereby facilitating the generalization and adaptability of trackers;
\item we conduct comprehensive experiments covering three multi-modality tracking tasks, seven  benchmarks, and diverse pre-trained trackers, demonstrating the
effectiveness and general applicability of SRFT.
\end{itemize}

The remainder of the paper is organized as follows. Section~\ref{sec2} reviews existing literature on multi-modality trackers and cross-modal transfer learning methods. In Section~\ref{sec3}, we rigorously formulate prior and transfer parameter significance from both pre-training and fine-tuning perspectives, followed by the implementation of significance-regularized tuning. Section~\ref{sec4} presents extensive experiments to demonstrate the effectiveness of our method, along with comprehensive ablation studies to analyze the impact of different components and designs. Finally, Section~\ref{sec5} concludes the paper.

\section{Related Work} \label{sec2}
\subsection{Multi-Modality Object Tracking}
Object tracking involves localizing an object across frames given its initial appearance~\citep{wei2023autoregressive,bai2024artrackv2,lin2024tracking,zheng2024odtrack}. However, the vulnerability of RGB-only trackers under adverse conditions has driven the development of multi-modality tracking, where auxiliary cues complement the intrinsic deficiencies of visible imagery~\citep{zhang2023glenet}. For instance, event cameras provide robust dynamic information under extreme motion or lighting, enabling RGB-event fusion for high-speed and low-dynamic tracking~\citep{zhu2022learning,chen2024crossei,Wang_2024_CVPR,zhang2024revisiting,wang2025mamba,wang2025towards}. Depth offers geometric priors that help handle occlusion and background clutter, and has been integrated to improve robustness in crowded scenes~\citep{lukezic2019cdtb,qian2021dal,yan2021depthtrack,10819455}. Thermal infrared imaging captures thermodynamic signatures independent of visible light, proves effective in low-illumination. \citep{zhang2021jointly,hui2023bridging,GDSTrack_2025_IJCAI,Xiang_2025_CVPR} demonstrated that fusing thermal and RGB data can yield more reliable appearance representations. 

In summary, contemporary methods emphasized effective multi-modality interaction and fusion~\citep{zhang2024universal,zhang2024revisiting,chen2024unifiedsequencetosequencelearningsingle,tan2025you,tan2024xtrack,fengcstrack}. With the emergence of large-scale RGB datasets and universal backbones (e.g., vision transformer), pre-trained trackers~\citep{ye2022joint,wu2023dropmae,chen2025sutrack} have demonstrated remarkable generalization across diverse scenarios. These advances have shifted the paradigm toward transferring pre-trained models for designing high-performance multi-modality trackers. Consequently, multi-modality tracking is increasingly driven by reusing RGB-induced semantic priors, particularly when annotated multi-modality data are scarce or noisy. In this work, we target this imperative by optimizing the adaptation of pre-trained trackers for efficacious cross-modal transfer learning.


\subsection{Cross-modal Transfer Learning}
To adapt pre-trained models for multi-modality tracking, two main transfer learning strategies have recently emerged. Some works adhere to the full fine-tuning (FFT) paradigm~\citep{tang2022revisiting,wang2023visevent,zhu2023cross,Sun_2025_CVPR}, where pre-trained models serve as weight initializations and are entirely re-trained on the target tasks. These methods require a shared or compact cross-modal feature space to inherit the generalization capability of the original model. Representatively, SUTrack~\citep{chen2025sutrack} employed a single vision transformer with unified input representations for multiple modalities, avoiding task-specific customization. While effective, one primary dilemma may be innate to FFT: the contradiction between the paucity of large-scale downstream datasets and the huge appetites for cross-domain transfer. This mismatch frequently leads to catastrophic forgetting of the pre-trained knowledge and severe overfitting to the limited target data. 

To alleviate this, profiting from the affluent experience of natural language processing and computer vision communities~\citep{jia2022visual,chen2022adaptformer,hu2021lora}, the field has pivoted toward parameter-efficient fine-tuning (PEFT). The core principle of PEFT is to keep the majority of pre-trained weights frozen while tuning only a minimal number of additional parameters dedicated to the new task. By lightly fine-tuning, these methods aim to preserve the generalization while mitigating overfitting. For example, ProTrack~\citep{yang2022prompting} and ViPT~\citep{zhu2023visual} modulated RGB features by introducing trainable auxiliary tokens into attention layers. Similarly, methods like BAT~\citep{cao2024bi} and SDSTrack~\citep{hou2024sdstrack} inserted lightweight adapter modules between attention layers for cross-modal shift compensation. More recently, unified trackers like UnTrack~\citep{wu2024single} and OneTracker~\citep{hong2024onetracker} employed LoRA and prompt tuning to seamlessly integrate multiple modalities, enabling effective unification across diverse inputs. Despite their effectiveness, PEFT methods suffer from intrinsic limitations. First, the rigid constraints can lead to underfitting, limiting the model's capacity to handle vast distribution drifts. Conversely, the randomly initialized add-on modules lack alignment with the pre-trained knowledge, creating an optimization gap that predisposes the model to overfitting. Consequently, achieving robust and reliable fine-tuning for multi-modality tracking remains an elusive goal.

\noindent \textbf{\textit{Remark.}} Beyond fine-tuning strategies in multi-modality tracking, we also review sensitivity-aware fine-tuning paradigms from adjacent domains to
clarify the distinction from our method. Recent studies have exploited parameter sensitivity for efficient fine-tuning (SPT) and continual learning (CL). Notably, our method differs fundamentally from these approaches in both objective and mechanism. Specifically, SPT~\citep{he2023sensitivity} and CDRA-SPT~\citep{chen2025sensitivity} leveraged sensitivity to guide structural adaptation for efficiency, e.g., sparse parameter updates or dynamic rank allocation. MACL~\citep{wang2024model} employed sensitivity to mitigate forgetting in sequential tasks by minimizing loss variance. Crucially, these approaches typically operate with an exclusive focus on the target context. Diverging from this, SRFT adopts an update-regularized paradigm: we define a hybrid parameter significance spanning pre-trained and cross-modal domains to balance prior generalization with target adaptability. This feature renders our method distinct from existing FFT, PEFT, and SPT methods.

\begin{figure}[t]
\begin{center}
\centerline{\includegraphics[width=0.405\textwidth]{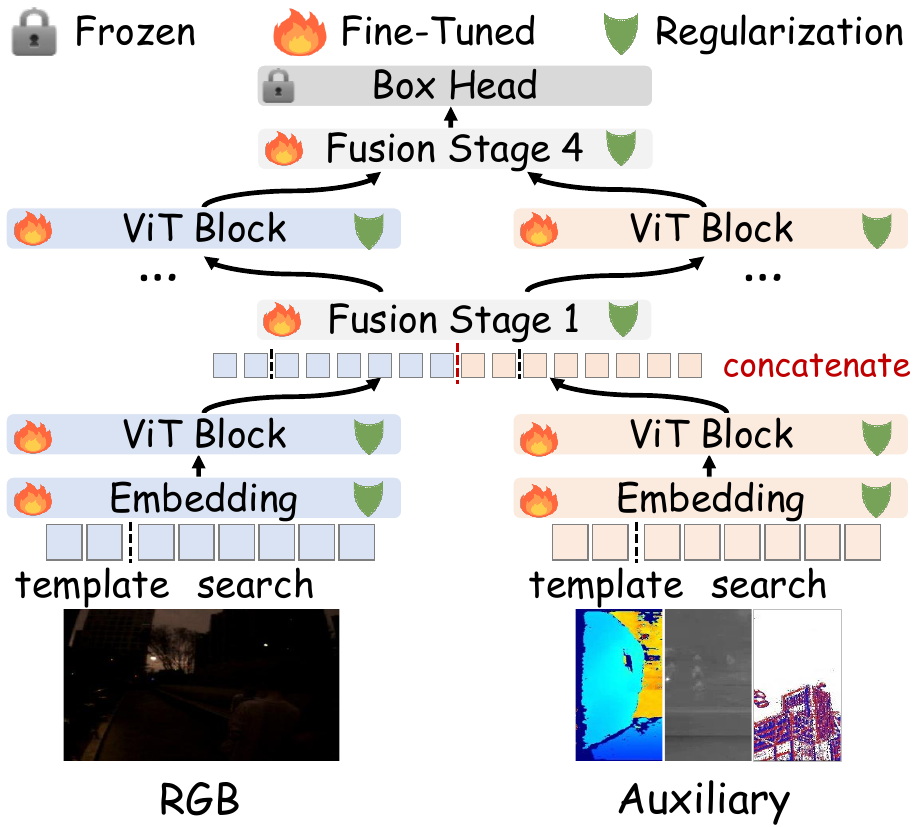}}
\vskip -0.05in
\caption{\textbf{Network architecture of our multi-modality trackers.} All modules are initialized with the weights of a pre-trained RGB tracker. We only fine-tune backbones and fusion blocks with significance-aware regularization terms.}
\label{fig:architecture}
\end{center}
\vskip -0.3in
\end{figure}

\section{Proposed Method} \label{sec3}
Learning generalized and coherent representations is crucial for adapting RGB-based models to multi-modality trackers. To unlock the full potential of pre-trained trackers, we revisit the core principles governing the cross-modal transfer process. In Section~\ref{sec:preliminaries}, we first present the architectural design of our multi-modality trackers and highlight the key challenges associated with full fine-tuning. Next, Section~\ref{sec:significance} identifies and quantifies pronounced prior and transfer parameter significance that characterize pre-trained knowledge and cross-domain adaptation. Finally, in Section~\ref{sec:srtuning}, we incorporate these significance measurements into the optimization phase, enabling adaptive and dynamic modulation of parameter updates.

\subsection{Preliminaries} \label{sec:preliminaries}
\noindent \textbf{\textit{Network Architecture of Multi-modality Trackers.}} Fig.~\ref{fig:architecture} depicts the architecture of our multi-modality trackers, in which all modules are initialized with the weights of a pre-trained RGB tracker. For the multi-modality tracker, RGB and auxiliary inputs are first fed to the embedding layer to generate the corresponding template and search tokens. Then, symmetric transformer backbones (e.g., ViT or its variants) handle feature extraction and interaction. Without involving customized multi-modality fusion modules, we repurpose certain ViT blocks (e.g., layers 2, 5, 8, and 11) for multi-stage fusion by concatenating multi-modality template and search tokens. Finally, the fused features are fed into the box head to estimate the object state. To retain modal-agnostic object association knowledge, the pre-trained box head is utilized and kept frozen. \textit{Additional model details can be found in \textbf{Appendix~\ref{sec:a-1}}.}

\vspace{0.3cm}
\noindent \textbf{\textit{Learning Process of Cross-Domain Transfer.}} We consider $f_\theta(\cdot): X \rightarrow Y$ as a multi-modality tracker with parameters $\theta \in \mathbb{R}^{|\theta|}$, with $|\theta|$ representing the total number of the parameters $\theta$. The model takes a task-specific dataset $D=\{X,Y\}=\left\{\left(x_i, y_i\right)\right\}_{i=1}^M$ to yield the optimal $\theta$, where $x_i$ represents a multi-modality input pair and $y_i$ is the corresponding object bounding box. In this work, we focus on efficiently transferring the models that have been pre-trained on a source domain $D_0$ (e.g., source RGB tracking data) to downstream domains $D_t$ (e.g., auxiliary modalities or degraded versions of images), where typically $\left|D_0\right| \gg\left|D_t\right|$. Let $\theta_0$ represent the pre-trained parameters, which are optimal for the source domain $D_0$, and these parameters are used as the weight initialization. After fine-tuning the model on the target domain $D_t$, the model parameters become $\theta_t$. Given a task-specific loss function $\mathcal{L}$, the \textbf{vanilla transfer objective} is formulated as follows:
\begin{equation}
\theta_t=\underset{\theta}{\operatorname{\arg \min}} \mathcal{L}\left(\theta \mid D_t\right), \quad {s.t.} \quad \theta^{(0)}=\theta_0,
\label{eq:vanilla}
\end{equation} 

\begin{figure}[t]
\centering
\centerline{\includegraphics[width=0.49\textwidth]{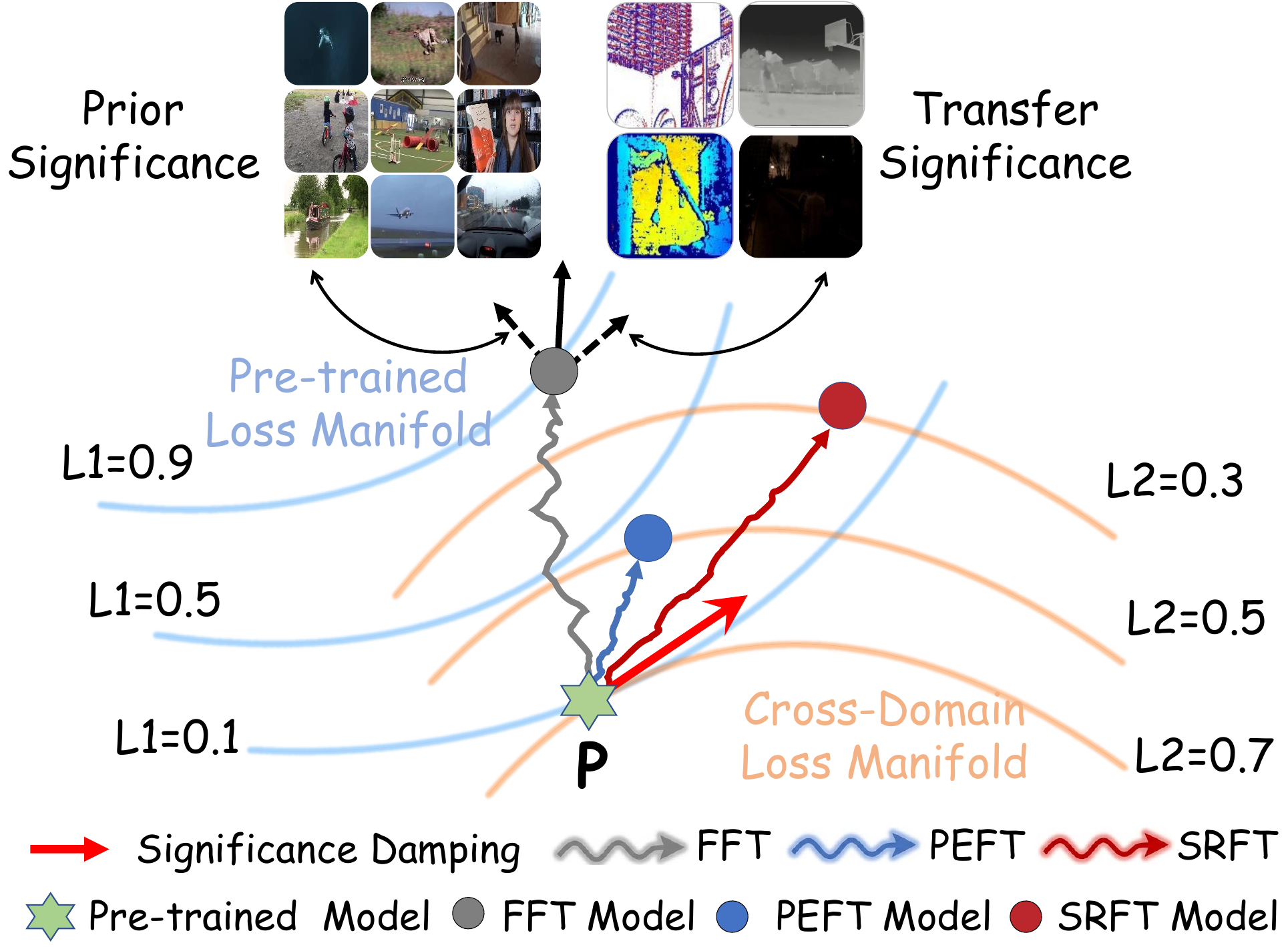}}
\caption{\textbf{Loss-landscape schematic across fine-tuning paradigms.} \textbf{\textit{FFT}} allows unrestricted parameter movement, risking severe pre-trained knowledge forgetting (deteriorating L1). While \textbf{\textit{PEFT}} restricts updates to additional parameters, this limits performance on new domains (stalled L2). In contrast, \textbf{\textit{SRFT}} \revise{regulates update mobility according to parameter significance, indicating a better stability--plasticity trade-off.}}
\label{fig:significance}
\vskip -0.15in
\end{figure}
\noindent However, directly optimizing $\theta_t$ as per Eq.~(\ref{eq:vanilla}) leads to severe overfitting and unstable tuning, as evidenced by the pronounced train-test performance gap and oscillations in training dynamics observed in Fig.~\ref{fig:training-trajectory}. This occurs because the model tends to over-adapt to the unstable target domain $D_t$, while neglecting the generalization capabilities learned from the source domain $D_0$. These challenges motivate a more principled transfer learning approach, as described next.

\subsection{Modeling Significance for Multi-Modality Tracking} \label{sec:significance}
Effective cross-domain fine-tuning jointly requires stability
(preserving pre-trained knowledge) and plasticity (adapting to the
target domain)~\citep{zheng2025towards,zhou2025revisiting}. We model these two requirements through complementary parameter
significance: \textbf{\textit{prior significance}}, characterizing
parameter importance to pre-trained knowledge, and
\textbf{\textit{transfer significance}}, capturing the evolving
parameter sensitivity to target-domain adaptation. Their combination provides a parameter-wise signal for regulating update mobility.

\vspace{0.1cm}
\noindent \textbf{\textit{Loss-Parameter Manifold.}}
As shown in Fig.~\ref{fig:significance}, for a given loss level $c$, the parameter
level set
$\mathcal{M}_{c}
=
\{\theta\in\mathbb{R}^{|\theta|}:\mathcal{L}(\theta)=c\}$
locally forms a codimension-one manifold~\citep{bengio2013representation,fefferman2016testing,
meilua2024manifold} at regular points where
$\nabla_{\theta}\mathcal{L}(\theta)\neq 0$. \revise{This manifold view provides an intuitive description of how parameter
movement during downstream fine-tuning affects the pre-trained
tracking objective. Since different perturbation
directions can induce markedly different source-loss variations, preserving
pre-trained knowledge requires accounting for this non-uniform local
sensitivity. We therefore characterize such local
source-loss sensitivity and use it to
formulate \textbf{\textit{prior significance}}.}


\subsubsection{\textbf{Safeguarding Generalization via Prior Significance}} 
In this context, we first analyze \revise{the local source-loss response to parameter perturbations} around the pre-trained solution and then design an efficient operation-wise curvature estimation scheme to quantify prior significance.

\vspace{0.1cm}
\noindent \textbf{\textit{Exploring Local Source-Loss Sensitivity.}}
We first define the joint empirical risks over the source and downstream tasks:
\begin{equation}
\mathcal{L}\left(\theta \mid D_u\right)
=
\mathcal{L}\left(\theta \mid D_t\right)
+
\beta \mathcal{L}\left(\theta \mid D_0\right),
\label{eq:joint_risk}
\end{equation}

\noindent where $D_u = D_0 \bigcup D_t$, $\beta>0$ balances the source and downstream objectives, and $\mathcal{L}\left(\theta \mid D_0\right)$ regularizes the downstream objective $\mathcal{L}\left(\theta \mid D_t\right)$ toward the pre-trained solution. The pre-trained weights $\theta_0$ minimize $\mathcal{L}\left(\theta \mid D_0\right)$. When $\theta$ is in the vicinity of $\theta_0$, let $\Delta\theta=\theta-\theta_0$. The source loss can then be locally approximated by the second-order Taylor expansion:
\begin{equation}
\begin{aligned}
\mathcal{L}\left(\theta \mid D_0\right)
&=
\mathcal{L}\left(\theta_0 \mid D_0\right)
+
\Delta\theta^T
\nabla_{\theta}\mathcal{L}\left(\theta_0 \mid D_0\right)
\\
&\quad+
\frac{1}{2}
\Delta\theta^T
\mathcal{H}^{(\theta_0)}
\Delta\theta
+
\mathcal{O}\left(\left\|\Delta\theta\right\|_2^3\right),
\end{aligned}
\label{eq:expansion}
\end{equation}
\noindent where $\nabla \mathcal{L}\left(\theta_0\right) \approx \mathbf{0}$, and \revise{$\mathcal{H}^{\left(\theta_0\right)} \in \mathbb{R}^{|\theta| \times|\theta|}$ denotes
the source-loss Hessian $\frac{\partial^2 \mathcal{L}(\theta \mid D_0)}
{\partial \theta_i \partial \theta_j}$ evaluated at $\theta_0$.}\footnote{\revise{For a probabilistic model with a corresponding negative-log-likelihood objective, the Fisher information equals the negative expected Hessian of the log-likelihood only under the model-distribution expectation and the usual regularity conditions~\citep{amari2019fisher,kunstner2019limitations}. The composite tracking objective in Eq.~(\ref{eq:loss}) is not assumed} \revise{to satisfy this identity. We therefore formulate SRFT directly using the source-loss Hessian; Fisher/empirical-Fisher quantities are used only in the corresponding comparison baselines.}}

Eq.~(\ref{eq:expansion}) shows that the source-loss increase induced by parameter displacement provides a local proxy for disrupting pre-trained tracking knowledge. Hence the local second-order source-loss response is
\begin{equation}
\varepsilon_{\mathrm{gen}}
=
\mathcal{L}(\theta \mid D_0)
-
\mathcal{L}(\theta_0 \mid D_0)
\approx
\frac{1}{2}\Delta\theta^{\top}
\mathcal{H}^{\left(\theta_0\right)}
\Delta\theta,
\label{eq:source_loss_variation}
\end{equation}
\noindent Thus, $\mathcal{H}^{\left(\theta_0\right)}$ characterizes the local sensitivity of the source tracking objective to parameter displacement. Perturbations along directions with large positive curvature can induce larger source-loss increases and therefore pose a higher risk of disrupting pre-trained knowledge. Accordingly, dominant positive-curvature responses provide the source-sensitivity signal to regulate update mobility during downstream adaptation.

\begin{figure}[t]
\centering
\centerline{\includegraphics[width=0.40\textwidth]{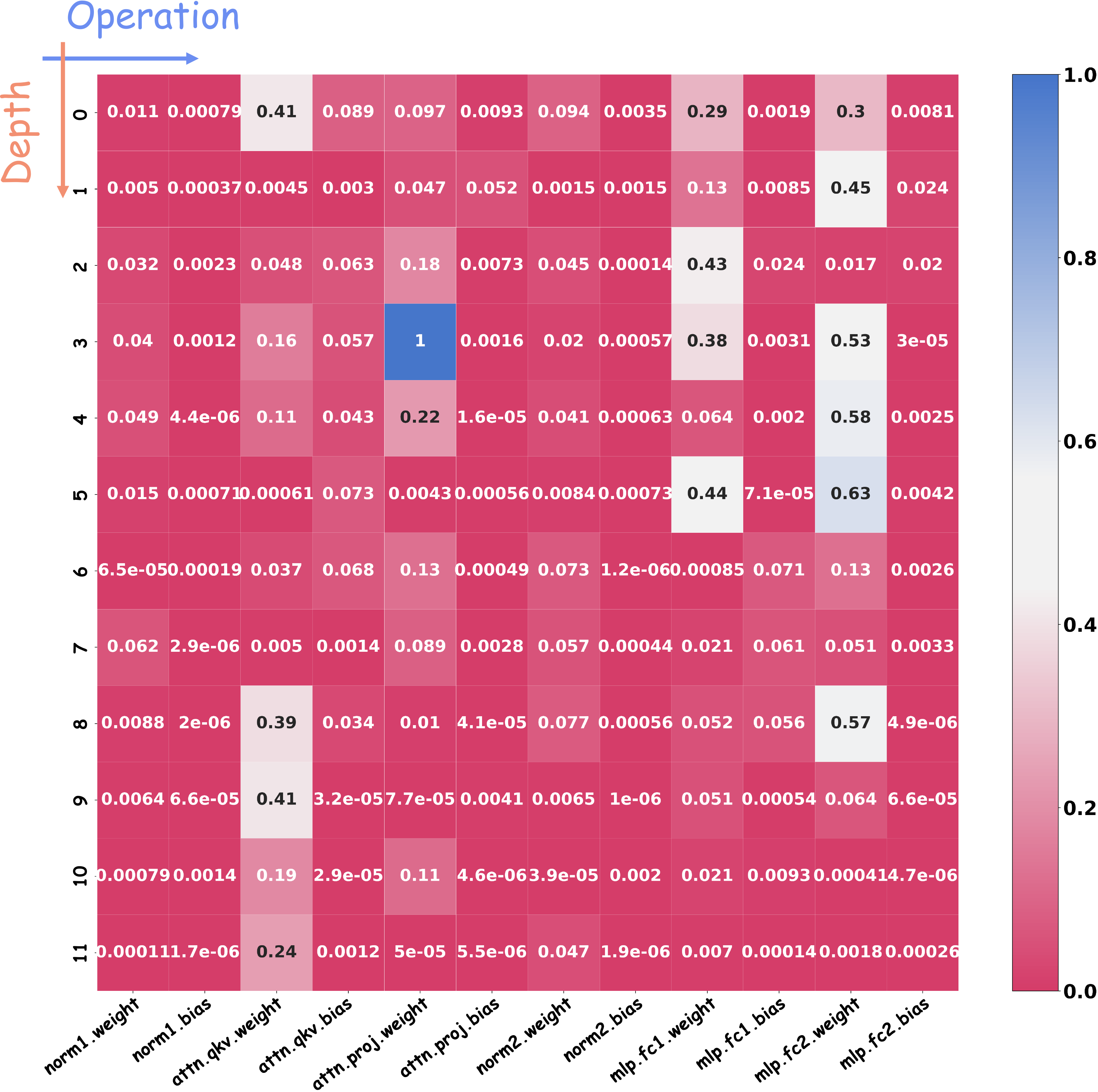}}
\vskip -0.1in
\caption{\textbf{Operation-wise prior parameter significance} estimated from the source-loss curvature of the pre-trained OSTrack on the source datasets. A larger prior significance indicates stronger local source-objective sensitivity and a higher risk of source-loss increase under parameter perturbations.}
\label{fig:prior_significance}
\end{figure}

\noindent \textbf{\textit{Operation-Wise Curvature Approximation.}}
Directly computing $\mathcal{H}^{(\theta_0)}$ for a tracking model is intractable due to its $\mathcal{O}(|\theta|^2)$ complexity. We therefore estimate dominant curvature responses at the operation level without explicitly forming or decomposing the full source-loss Hessian. Analogous to loss-landscape analysis, large positive curvature indicates ``steep'' and source-sensitive directions, whereas small curvature suggests weaker source-loss responses and greater local update mobility. Thus, dominant curvature responses provide a compact characterization of operation-level prior significance.

Specifically, we partition the model parameters into $N$ disjoint operation groups,
$\theta=\{\theta^1,\ldots,\theta^N\}$, e.g., MLPs of FFNs and attention QKVs, and adopt the operation-wise group-diagonal approximation $
\mathcal{H}_{\mathrm{bd}}^{(\theta_0)}
=
\operatorname{diag}
\left(
\mathcal{H}^{(\theta_0^1)},
\ldots,
\mathcal{H}^{(\theta_0^N)}
\right),
$
where
$\mathcal{H}^{(\theta_0^j)}
\in
\mathbb{R}^{|\theta^j|\times|\theta^j|}$
denotes the source-loss Hessian block of operation $j$. This approximation retains intra-operation curvature while omitting, rather than assuming zero, cross-operation curvature.

\revise{For a perturbation $\Delta\theta$ restricted to operation $j$, i.e.,
$\Delta\theta=[0,\ldots,(\Delta\theta^j)^T,\ldots,0]^T$,
the Hessian quadratic form exactly reduces to $
\Delta\theta^T
\mathcal{H}^{(\theta_0)}
\Delta\theta
=
(\Delta\theta^j)^T
\mathcal{H}^{(\theta_0^j)}
\Delta\theta^j.
$
Thus, $\mathcal{H}^{(\theta_0^j)}$ directly characterizes the local source-loss sensitivity to perturbations within operation $j$, providing the basis for its prior significance. Since the source-loss Hessian may be indefinite, these operation-wise quantities characterize relative curvature sensitivity rather than reconstruct the complete loss geometry.}

Empirical studies~\cite{ghorbani2019investigation,rame2022fishr} suggest that deep-network loss curvature is often dominated by a limited number of strong responses followed by a relatively flat spectral tail. This motivates summarizing the dominant curvature of each operation rather than reconstructing its complete curvature block. For theoretical characterization, let the exact eigendecomposition of the $j$-th operation-wise Hessian block be
$
\mathcal{H}^{(\theta_0^j)}
=
V^j\Lambda^j(V^j)^T
$
with
$
\Lambda^j
=
\operatorname{diag}
(\lambda_1^j,\ldots,\lambda_{|\theta^j|}^j),
$
where $V^j$ is orthogonal and
$
\lambda_1^j\geq\lambda_2^j\geq\cdots\geq\lambda_{|\theta^j|}^j
$
are the exact Hessian eigenvalues. These spectral quantities are used only for theoretical analysis; practical SRFT does not recover the eigenvalues or eigenvectors of $\mathcal{H}^{(\theta_0^j)}$.

For any non-zero direction $v^j$, the Rayleigh quotient is
\begin{equation}
\mathcal{R}
\left(
\mathcal{H}^{(\theta_0^j)},v^j
\right)
=
\frac{
(v^j)^T
\mathcal{H}^{(\theta_0^j)}
v^j
}{
(v^j)^T v^j
}.
\label{eq:rayleigh}
\end{equation}
By the Rayleigh--Ritz theorem~\citep{li2015rayleigh}, its maximum equals $\lambda_1^j$ and is attained in the corresponding leading eigenspace. Hence, larger positive Rayleigh quotients indicate stronger positive-curvature responses of the pre-trained source objective, motivating Rayleigh-quotient probing without explicitly recovering the eigenspace.

\vspace{0.1cm}
\begin{proposition}[\textbf{\textit{Error Induced by Operation-Wise Curvature Scalarization.}}]
Let
$\mathcal{H}^{(\theta_0^j)}
=
V^j\Lambda^j(V^j)^T$
denote the exact operation-wise source-loss Hessian block associated with operation $j$. Define its group-isotropic scalarization as
\begin{equation}
\widetilde{\mathcal{H}}^{(\theta_0^j)}
=
\gamma^j I_{|\theta^j|},
\quad
\text{with}
\quad
\gamma^j
=
\frac{1}{K}\sum_{i=1}^K\lambda_i^j,
\label{eq:curvature_scalarization}
\end{equation}
where
$
\widetilde{\mathcal{H}}^{(\theta_0)}
=
\operatorname{diag}
(
\widetilde{\mathcal{H}}^{(\theta_0^1)},
\ldots,
\widetilde{\mathcal{H}}^{(\theta_0^N)}
)
$
denotes the group-isotropic scalarization of
$\mathcal{H}_{\mathrm{bd}}^{(\theta_0)}$.
Then the following hold:

\begin{enumerate}
\revise{
\item \textbf{Exact Scalarization Error.}
For each operation $j$,
$
\big\|
\mathcal{H}^{(\theta_0^j)}
-
\widetilde{\mathcal{H}}^{(\theta_0^j)}
\big\|_F^2
=
K\operatorname{Var}(\lambda_{1:K}^j)
+
\sum_{i=K+1}^{|\theta^j|}
(\lambda_i^j-\gamma^j)^2,
$
where
$
\operatorname{Var}(\lambda_{1:K}^j)
=
K^{-1}\sum_{i=1}^{K}(\lambda_i^j-\gamma^j)^2.
$}
\revise{The first term captures the dispersion of the retained top-$K$ eigenvalues, while the second measures their mean mismatch with the remaining spectrum.}

\revise{
\item \textbf{Bounded Source-Loss Response Error.}
For any
$\Delta\theta\in\mathbb{R}^{|\theta|}$,
let
$
\varepsilon_{\mathrm{gen}}(A)
=
\frac{1}{2}\Delta\theta^T A\Delta\theta.
$
Then
$
\big|
\varepsilon_{\mathrm{gen}}(\mathcal{H}_{\mathrm{bd}}^{(\theta_0)})
-
\varepsilon_{\mathrm{gen}}(\widetilde{\mathcal{H}}^{(\theta_0)})
\big|
\leq
\tfrac{1}{2}\|\Delta\theta\|_2^2
\max_{1\leq j\leq N}
\big\{
\lambda_1^j-\gamma^j,\,
\gamma^j-\lambda_{|\theta^j|}^j
\big\},
$
where the second term accounts for the lower end of the spectrum, including possible negative curvature.}
\end{enumerate}
\end{proposition}

\begin{proof}
    See \textit{\textbf{Appendix~\ref{sec:a-2-1}}}.
\end{proof}

\revise{The proposition characterizes the information loss from group-isotropic scalarization rather than implying preservation of the within-operation eigenspace. Accordingly, $\widetilde{\mathcal{H}}^{(\theta_0)}$ remains isotropic within operations while retaining inter-operation curvature heterogeneity. The source-loss response bound isolates the scalarization error of $\mathcal{H}_{\mathrm{bd}}^{(\theta_0)}$ from that due to omitted cross-operation curvature. Since SRFT uses operation-level sensitivity to regulate update mobility within first-order fine-tuning, this scalarization provides an appropriate abstraction for significance-aware optimization.}

\vspace{0.1cm}
\noindent \textbf{\textit{Prior Significance Measurement.}}
\revise{The theoretical quantities $\lambda_i^j$ and $\gamma^j$ characterize the exact spectral properties of $\mathcal{H}^{(\theta_0^j)}$. Since explicitly constructing and diagonalizing each Hessian block is computationally expensive, we propose an efficient estimator of its dominant directional curvature responses based on symmetric finite-difference Rayleigh-quotient probes~\citep{leveque1998finite}.}

For operation $j$, we sample
$v_r^j\in\mathbb{R}^{|\theta^j|}$
from an isotropic Gaussian distribution,
$v_r^j\sim\mathcal{N}(0,I)$,
and construct the normalized group-wise perturbation
$
\epsilon_r^j
=
\rho
\|\theta_0^j\|_2
\,v_r^j
\big/
\|v_r^j\|_2,
$
where $\rho=1\times10^{-5}$ sets the perturbation radius relative to the corresponding parameter-group norm. When applied to the full parameter vector, $\epsilon_r^j$ is zero outside operation $j$. The observed directional curvature response of the $r$-th probe is
\begin{equation}
\begin{aligned}
q_r^j
&=
\frac{
(\epsilon_r^j)^T
\mathcal{H}^{(\theta_0^j)}
\epsilon_r^j
}{
\|\epsilon_r^j\|_2^2
}
\\
&\approx
\frac{
\mathcal{L}(\theta_0+\epsilon_r^j)
-
2\mathcal{L}(\theta_0)
+
\mathcal{L}(\theta_0-\epsilon_r^j)
}{
\|\epsilon_r^j\|_2^2
}.
\end{aligned}
\label{eq:curvature_probe}
\end{equation}
Here, $\mathcal{L}(\theta_0)\triangleq\mathcal{L}(\theta_0 | D_0)$, and $\epsilon_r^j$ is zero-padded outside operation $j$ in the full parameter space. \revise{The quantity $q_r^j$ is an \textbf{observed Rayleigh-quotient curvature response}, rather than an \textbf{exact eigenvalue} of $\mathcal{H}^{(\theta_0^j)}$. The symmetric finite difference suppresses first-order effects and directly estimates local directional curvature without explicitly forming the Hessian.}

\revise{By repeating the probing over different directions, we obtain a set of observed curvature responses for each operation and retain the $K$ largest positive responses,
$
q_{(1)}^j
\geq
q_{(2)}^j
\geq
\cdots
\geq
q_{(K)}^j
$, where $K$ denotes the retained curvature-response budget. Their mean defines the operation-level prior significance:
\begin{equation}
s_j^p
=
\frac{1}{K}
\sum_{i=1}^{K}
q_{(i)}^j.
\label{eq:prior_significance}
\end{equation}
Eq.~(\ref{eq:prior_significance}) provides the practical probe-based counterpart to the theoretical curvature statistic $\gamma^j$ in Eq.~(\ref{eq:curvature_scalarization}), rather than an exact recovery of Hessian eigenvalues. Specifically, $\lambda_i^j$ and $\gamma^j$ are used only for theoretical spectral characterization, whereas $s_j^p$ is estimated from the observed curvature responses $q_r^j$.} We set $K=10$ by default. In practice, even the largest retained curvature response alone (i.e., $K=1$) provides a reliable indicator of operation-level prior significance. Further results are reported in \textit{\textbf{Section~\ref{sec:ablation} and Appendix~\ref{app:probing_efficiency}}}.



\subsubsection{\textbf{Stabilizing Adaptation via Transfer Significance}} In this context, we analyze the impact of sparse transfer gradients on model adaptation and propose rebalancing them using the gradient matrix.

\begin{figure}[t]
\centering
\centerline{\includegraphics[width=0.49\textwidth]{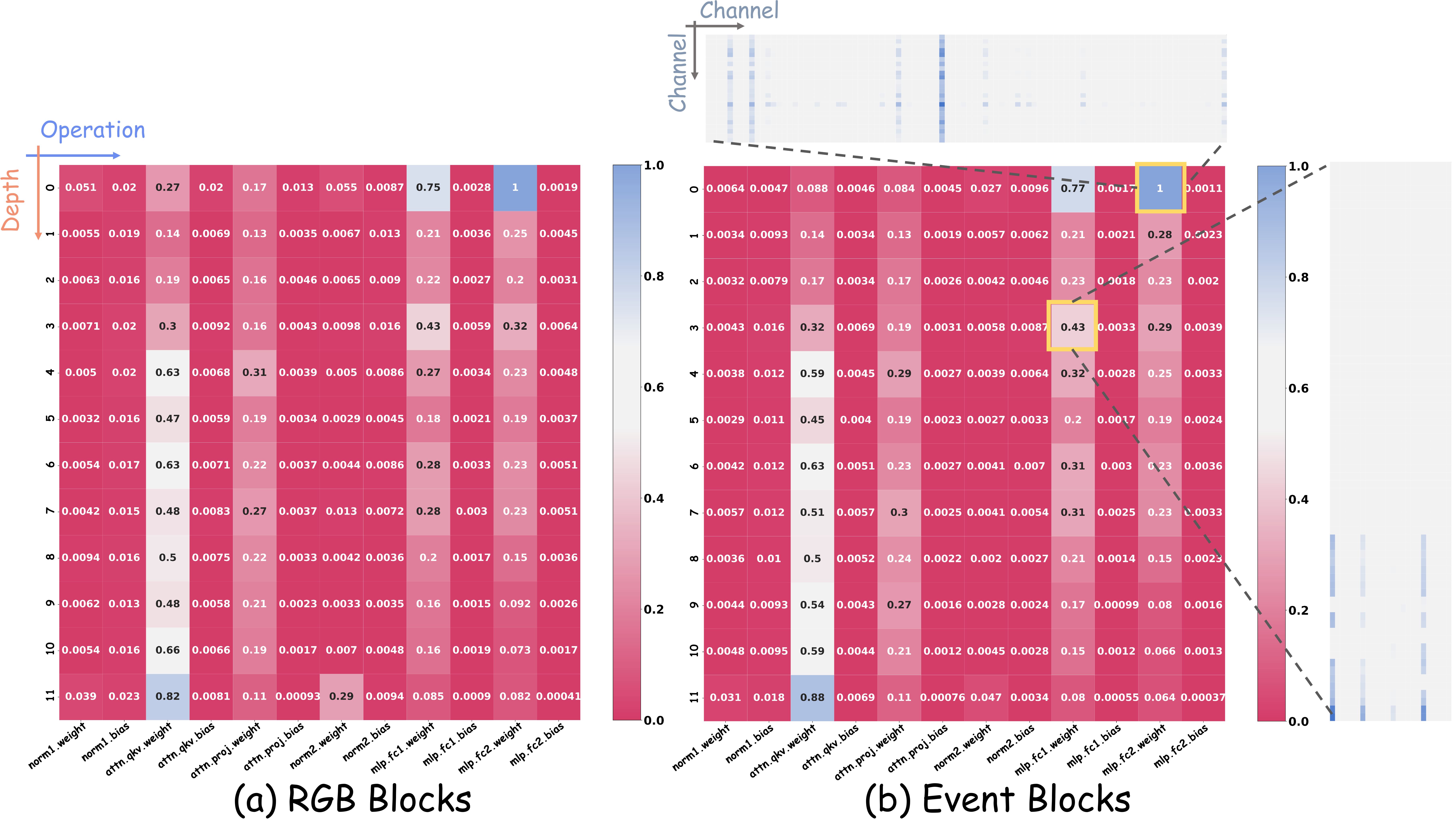}}
\vskip -0.1in
\caption{\textbf{Instantaneous transfer parameter significance} on the VisEvent dataset observed during fine-tuning. The high sparsity in the significance map indicates that only a few gradients (light areas) dominate the updates.}
\label{fig:transfer-significance}
\vskip -0.1in
\end{figure}

\noindent \textbf{\textit{Characterizing Transfer-Gradient Sparsity.}} While prior significance addresses forgetting of pre-trained domain, inadaptability in the downstream domain remains notably pronounced. Specifically, multi-modality tracking often involves diverse and heterogeneous samples exhibiting substantial gaps, such as auxiliary inputs or degraded images. These shifts manifest as highly sparse gradients during fine-tuning process on $D_t$, as illustrated in Fig.~\ref{fig:transfer-significance}. High sparsity indicates that only a few gradients dominate the updates, while the majority remain nearly static. This phenomenon sheds light on a key factor underlying the limited adaptability of multi-modality trackers and potentially results in temporal oscillations (see FFT test branch in Fig.~\ref{fig:training-trajectory}). Thus, we investigate the elevated adaptation risk induced by sparse gradients and propose a transfer significance to mitigate its impact.

\vspace{0.1cm}
Formally, we quantify the sparsity of the gradient by comparing its $L_1$ and $L_2$ norms. Let $\mathcal{G} \doteq \nabla_\theta\mathcal{L}\left(\theta \mid M\right)$ be the gradient on the current batch $M \subset D_t$. We define the sparsity measure: $\rho=\frac{\left\|\mathcal{G}\right\|_1}{\sqrt{|\theta|}\left\|\mathcal{G}\right\|_2}$. This definition leverages the fact that $L_1$ and $L_2$ norms characterize different aspects of the gradient vector: while $L_1$ reflects the total variation, $L_2$ emphasizes concentration~\cite{hurley2009comparing}. Thus, their ratio serves as an effective proxy for sparsity structure, where a smaller value means higher sparsity. Moreover, noting that single-step gradient magnitudes are typically bounded~\cite{nesterov2013introductory}, increasing sparsity (i.e., $\rho: 1 \rightarrow \frac{1}{\sqrt{|\theta|}}$) can amplify the $L_2$ norm of the gradient vector:    
\begin{equation}
\left\|\mathcal{G}\right\|_2= \frac{\left\|\mathcal{G}\right\|_1}{\rho\sqrt{|\theta|}}.
\label{eq:gradient_sparsity}
\end{equation}

\noindent \textbf{\textit{Sparsity-Induced Adaptation Risk.}} We characterize tuning volatility by the sensitivity of the target-domain
objective to small stochastic variations in parameter updates during
fine-tuning. Accordingly, we define the following response function:
\begin{equation}
\begin{aligned}
S\left(\theta, \delta \mid M\right)=\mathcal{L}\left(\theta+\delta \mid M\right)-\mathcal{L}\left(\theta \mid M\right),
\end{aligned}
\label{eq:gradient_significance}
\end{equation}

\noindent where $\delta$ a stochastic parameter displacement centered at the nominal gradient-descent step, i.e., $\delta \sim \mathcal{N}\left(-\alpha \mathcal{G}, \alpha^2\sigma^2 I\right)$, $\alpha$ is the learning rate and $\sigma$ controls the noise. This denotes that updates follow the descent direction with inherent noise (due to stochastic sampling, batch gap, etc.). 

\vspace{0.1cm}
Intuitively, a smooth and stable loss landscape should ensure that varying perturbations yield consistent responses. The expected magnitude of this difference defines the adaptation risk:
\begin{equation}
\begin{aligned}
\varepsilon_{ada}=\underset{\left(\delta, \delta^{\prime}\right)}{\mathbb{E}}\left[|S\left(\theta, \delta \mid M\right)-S\left(\theta, \delta^{\prime} \mid M\right)\right|].
\end{aligned}
\label{eq:adaptation_risk}
\end{equation}

\noindent By incorporating the linear approximation $S\left(\theta, \delta \mid M\right) \approx \mathcal{G}^T \delta$ into Eq.~(\ref{eq:adaptation_risk}), it can be shown that this adaptation risk scales with the $L_2$ norm of the gradient vector: \begin{equation}
\varepsilon_{ada} \le \alpha \sqrt{2 \sigma^2}\left\|\mathcal{G}\right\|_2.
\label{eq:adaptation_l2}
\end{equation}

\noindent Derivation details of Eqs.~(\ref{eq:gradient_sparsity})-(\ref{eq:adaptation_risk}) are provided in \textit{\textbf{Appendix~\ref{sec:a-2-2}}}. Thus, we can deduce that sparse gradients exacerbate tuning instability and adaptation risk. This highlights the need to avoid concentrating updates on specific parameters and emphasizes the importance of a balanced transfer gradient. 

\vspace{0.1cm}
\noindent \textbf{\textit{Transfer Significance Estimation.}} To formalize the estimation of transfer significance, we adapt the formulation of Eq.~(\ref{eq:gradient_significance}) to the standard fine-tuning process. In this case, we adjust the parameter $\theta$ via the gradient $\mathcal{G}$ rather than noise perturbations. Therefore, we can derive: $
S\left(\theta, \delta \mid M\right) \approx \mathcal{G}^T \mathcal{G}$. Here, the significance function is used to assess the overall adjustment of all parameters during transfer. To further investigate and regularize gradients, we extend the transfer significance to a parameter-wise formulation: 
\begin{equation} 
s_{n,i}^t=\left(\frac{\partial \mathcal{L}\left(\theta \mid M_i\right)}{\partial \theta_n}\right)^2.
\label{eq:transfer_significance}
\end{equation}

\noindent where $s_{n,i}^{t}$ denotes the transfer significance of the
$n^{\text{th}}$ parameter at iteration $i$. This formulation allows us to examine the granular impact of parameter changes during the transfer process and rebalance the gradient accordingly.

\begin{algorithm}[tb]
   \caption{Significance-Regularized Fine-Tuning}
   \label{alg:mst}
    \begin{algorithmic}
   \STATE {\bfseries Input:} source RGB tracking data $D_0$, downstream task dataset $D_t$, pre-trained weight $\theta_0$; curvature-response budget $K$,  number of model operations $N$, training steps $T$; initialized curvature set \textbf{$Q$}; \\
   \STATE {\bfseries Output:} Optimal multi-modality tracker parameters $\theta_t$; \\
   \STATE {\bfseries Step1:} \textbf{Prior Significance Estimation}: \\
   \FOR{batch $(x,y)$ in $D_0$}
   \FOR{$j \in\{1, \ldots, N\}$}
   \STATE Sample $v^j \sim \mathcal{N}\left(\mathbf{0}, I\right)$;
   \STATE Probe $q^j$ via Eq.~(\ref{eq:curvature_probe});
   \STATE Update curvature set \textbf{$Q$} for top-$K$ $q^j$;
   \ENDFOR
   \ENDFOR
   \STATE Calculate prior significance $s_j^p$ via Eq.~(\ref{eq:prior_significance}); \\
   \STATE {\bfseries Step2:} \textbf{Transfer Significance Estimation and Regularized Tuning}: \\
   \FOR{$i \in\{1, \ldots, T\}$} 
   \STATE Sample $i$-th batch data $M_i$ from $D_t$;
   \STATE Compute loss $\mathcal{L}(\theta \mid M_i)$ and raw gradients;\\
   \STATE Update transfer significance by $s_{n,i}^t=(\frac{\partial \mathcal{L}\left(\theta \mid M_i\right)}{\partial \theta_n})^2$; \\
   \STATE Normalize and combine significance $s_n$ by Eq.~(\ref{eq:combine_significance});\\
   \STATE Obtain the standard parameter displacement $\Delta\theta_{n,i}$ produced by the optimizer; \\
   \STATE Parameter update via Eq.~(\ref{eq:update}); \\
   \ENDFOR
\end{algorithmic}
\end{algorithm}
\vskip -0.1in

\subsection{Significance-Regularized Fine-Tuning} 
\label{sec:srtuning}
We regularize the learning process by previously discussed significance metrics to derive enhanced multi-modality trackers. To align with the unified regularization tuning, we rank both the operation-wise prior significance and parameter-wise transfer significance, and then normalize them within the continuous range of $[0,1]$. To harmonize these two significance components, we devise a dynamic linear schedule to adjust their weighted combination. \revise{At the beginning of training, the prior- and transfer-significance weights are
$\kappa$ and $1-\kappa$, respectively, where $\kappa\in[0,1]$.
During training, the two weights are linearly interpolated from
$(\kappa,1-\kappa)$ to $(1-\kappa,\kappa)$. This provides a smooth
balance between preserving pre-trained knowledge and stabilizing
target-domain adaptation throughout training.} Formally, at each training step $i$, the combined parameter significance is updated as follows:
\begin{equation} 
s_{n,i}=\left(\kappa+(1-2\kappa)\frac{i}{T}\right)s_j^p+\left(1-\kappa-(1-2\kappa)\frac{i}{T}\right)s_{n,i}^t,
\label{eq:combine_significance}
\end{equation}

\noindent where $\theta_n \in \theta^j$, $T$ is the total number of training steps. Finally, we normalize the joint significance into
$s_{n,i}\in[0,0.99]$.

\revise{We next incorporate the joint significance into parameter updates by regulating their update mobility. Let $\alpha$ denote the learning rate and $d_{n,i}$ denote the update direction produced by
the base optimizer for parameter $n$ at iteration $i$, with the
corresponding base update
$\Delta\theta_{n,i}=-\alpha d_{n,i}$.
We consider the following local mobility-regularized surrogate:
\begin{equation}
\Delta\theta_i^{*}
=
\arg\min_{\Delta\theta}
\sum_n
\left[
d_{n,i}^{\top}\Delta\theta_n
+
\frac{\|\Delta\theta_n\|_2^2}
{2\alpha(1-s_{n,i})}
\right].
\label{eq:local_surrogate}
\end{equation}
Its first-order optimality condition gives
$
\Delta\theta_{n,i}^{*}
=
(1-s_{n,i})\Delta\theta_{n,i}.
$
Thus, the significance factor $1-s_{n,i}$ can be interpreted as
controlling the local mobility of each parameter under the adopted
surrogate. Applying this modulation yields:}
\revise{
\begin{equation}
\theta_{n}^{(i+1)}
=
\theta_{n}^{(i)}
+
\left(1-s_{n,i}\right)
\Delta\theta_{n,i},
\qquad
\theta^{(0)}=\theta_0.
\label{eq:update}
\end{equation}}
\revise{A larger significance therefore induces a stronger local mobility
penalty and reduces the corresponding parameter update, suppressing
excessive drift while preserving the base optimizer's update.
We emphasize that the modulated update in Eq.~(\ref{eq:update}) follows from the exact minimizer of the adopted quadratic surrogate in Eq.~(\ref{eq:local_surrogate}); this
interpretation does not imply that the linear mapping $1-s_{n,i}$ is
unique or globally optimal. Further details are provided in
\textit{\textbf{Appendix~\ref{sec:a-2-3}}}.}

\vspace{0.1cm}
\noindent \textbf{\textit{Discussion.}} Unlike prior works that rely on static ``sensitivity'' for sparse tuning, SRFT introduces a hybrid, dynamic notion of "significance" combining source-domain curvature with target-domain optimization response. Instead of freezing insensitive parameters, SRFT penalizes highly sensitive ones to suppress excessive updates, without structural constraints or discontinuous optimization. This regularization-centric scheme forms an adaptive update path for cross-modal transfer, distinctly departing from existing sparsity- or retention-based methods. \revise{\textit{Additional comparative experiments can be found in \textbf{Appendix~\ref{sec:spt}, \ref{app:classical_regularization} and \ref{app:update_stabilization}}.}}

\begin{figure*}[t]
\centering
\includegraphics[clip, width=0.99\textwidth]{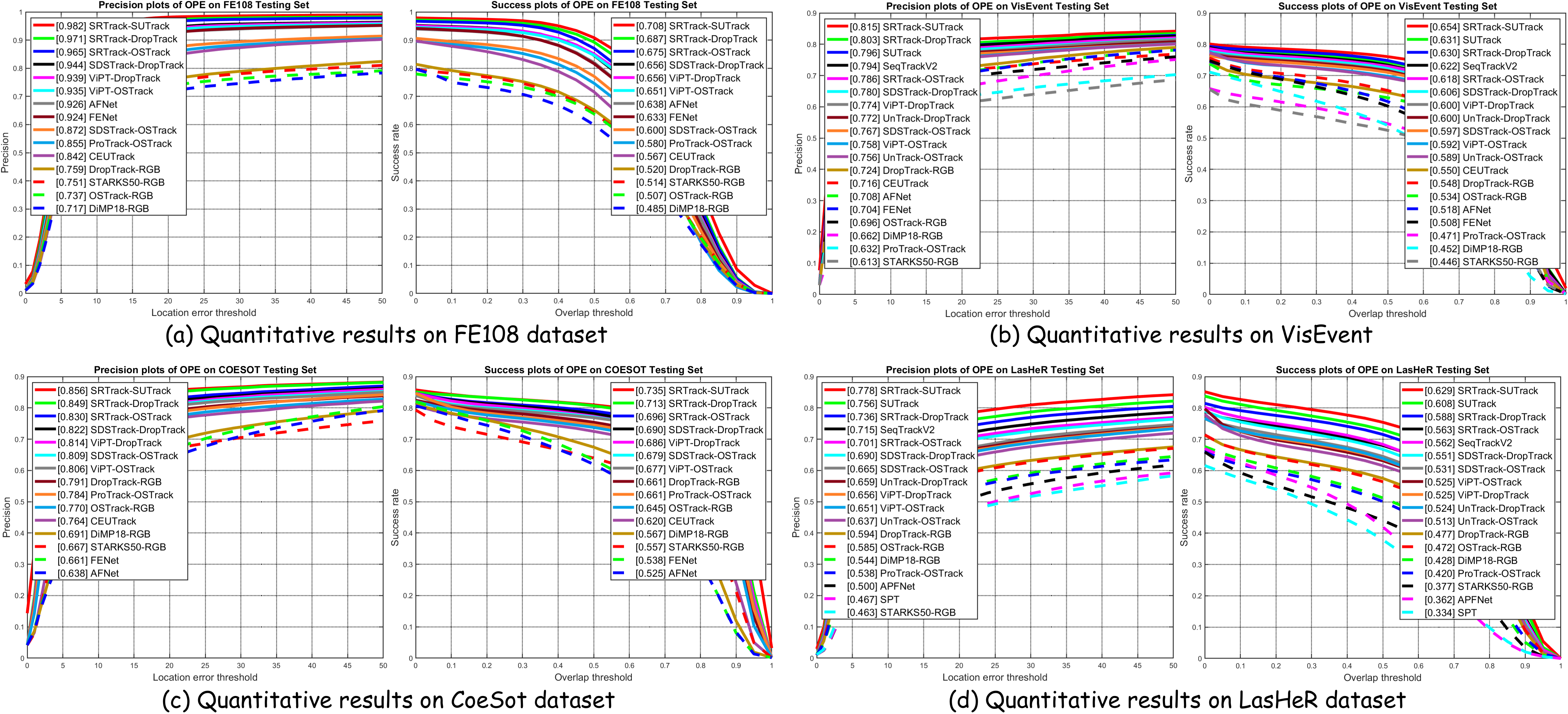}
\vskip -0.05in
\caption{\textbf{Precision and success plots} of the FE108, VisEvent, CoeSot and LasHeR datasets. Zoom in to see details.}
\label{fig:performance}
\vskip -0.10in
\end{figure*}

\begin{figure*}[thbp]
\begin{center}
\centerline{\includegraphics[width=0.99\textwidth]{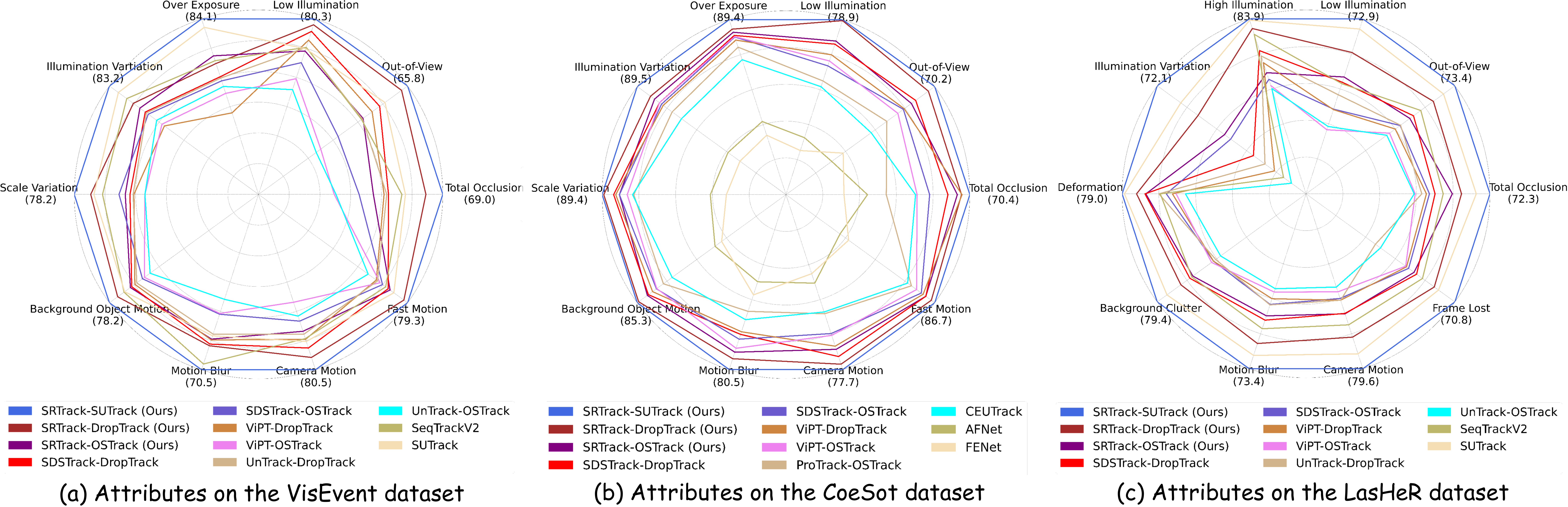}}
\vskip -0.05in
\caption{\textbf{Precision scores of different attributes} on the VisEvent, CoeSot and LasHeR test sets. Zoom in to see details.}
\label{fig:attributes}
\vskip -0.34in
\end{center}
\end{figure*}

\subsection{Learning Objectives}
The overall loss function of our method is the same as the foundation model without extra adjustments, shown as:
\begin{equation}
\mathcal{L}=\mathcal{L}_{\text {cls}}+\lambda_{\text {iou }} \mathcal{L}_{\text {iou }}+\lambda_{l_1} \mathcal{L}_1,
\label{eq:loss}
\end{equation}
where $\mathcal{L}_{cls}$ is the weighted focal loss for classification, $L_1$ loss $\mathcal{L}_1$ and GIoU loss $\mathcal{L}_{i o u}$ are employed for bounding box regression, $\lambda_{iou}=2$ and $\lambda_{l_1}=5$ are the loss weights, and all the corresponding settings are the same as~\citep{ye2022joint}.

\section{Experiments} \label{sec4}
\subsection{Experiment Settings}
\subsubsection{Benchmark Datasets}
\revise{\noindent \textbf{Source RGB tracking data.}
We denote by $D_0$ the source RGB tracking data used to train the
pre-trained source tracker, consisting of LaSOT~\citep{fan2019lasot},
COCO~\citep{lin2014microsoft}, TrackingNet~\citep{muller2018trackingnet},
and GOT-10k~\citep{huang2019got}. Note that $D_0$ refers to the
task-specific source tracking data rather than the generic backbone
pre-training corpus. Unless otherwise specified, prior significance is
estimated using the complete $D_0$. \textit{\textbf{Appendix~\ref{app:proxy_data}}}
further evaluates GOT-10k alone as a proxy when the complete $D_0$ is
unavailable; this is a proxy-data rather than a data-free setting.} \\
\noindent\textbf{Multi-modality tracking dataset.} To evaluate the effectiveness and generalization of the proposed method, we conduct comprehensive experiments on seven multi-modality benchmark datasets. For RGB-Event tracking, we use FE108~\citep{zhang2021object} (76 train / 32 test sequences) captured under degraded conditions, VisEvent~\citep{wang2023visevent}(500 train / 320 test sequences) covering dynamic outdoor scenes, and CoeSot~\citep{tang2022revisiting} with 578K image-event pairs (824 train / 528 test sequences). For RGB-Depth tracking, we adopt DepthTrack~\citep{yan2021depthtrack} (152 train / 50 test videos) and VOT-RGBD2022~\citep{kristan2022tenth} (127 test sequences). For RGB-Thermal tracking, we use LasHeR~\citep{li2020challenge} (979 train / 245 test sequences) and RGBT234~\citep{li2019rgb} (234 test videos). All training and evaluation strictly follow established protocols~\citep{zhu2023visual,wu2024single}. During fine-tuning, we use the training sets of FE108, VisEvent, and CoeSot for RGB-Event tasks; DepthTrack for RGB-Depth tasks; and LasHeR for RGB-Thermal tasks. The corresponding test sets, along with VOT-RGBD2022 and RGBT234, are used for evaluation.

\begin{table*}[t]
\captionsetup{skip=0pt}
\caption{Quantitative comparison on the RGB-Event datasets. \revise{Within each base-model group, the best result is highlighted in
\textbf{bold} and the second-best result is \underline{underlined}. methods built on DiMP/Stark-S are listed for reference only.}}
\centering
\renewcommand{\arraystretch}{1.0}
\setlength{\tabcolsep}{7.5pt}
\small
\begin{tabular}{c| c | c |c c | c c | c c }
    \midrule[1.1pt]
    \multirow{2}{*}{\textbf{Method}} &\multirow{2}{*}{\textbf{Reference}} &\multirow{2}{*}{\textbf{Base Model}}  &\multicolumn{2}{c|}{\textbf{FE108}} &\multicolumn{2}{c|}{\textbf{VisEvent}} &\multicolumn{2}{c}{\textbf{CoeSot}}  \\
    \cline{4-9}
    \multicolumn{1}{c|}{} & &  &SR &PR &SR  &PR &SR  &PR \\
    \midrule[1.1pt]
    \rowcolor{gray!10} \multicolumn{9}{c}{\textbf{Image-based Methods (Only-RGB)}}   \\
    DiMP \citep{bhat2019learning} &ICCV'19 &ResNet &48.5  &71.7  &45.2  &66.2  &56.7 &69.1   \\ 
    Stark-S \citep{yan2021learning} &ICCV'21 &ResNet &51.4  &75.1  &44.6   &61.3  &55.7 &66.7  \\
    OSTrack \citep{ye2022joint} &ECCV'22 &OSTrack-256   &50.7   &73.7  &53.4  &69.6  &64.3  &76.3   \\
    DropTrack \citep{wu2023dropmae} &CVPR'23 &DropTrack-384  &52.0    &75.9  &54.8  &72.4  &66.1  &79.1  \\
    \midrule[1.1pt]
    \rowcolor{blue!8} \multicolumn{9}{c}{\textbf{Cross-modal Transfer Learning}}   \\
    FENet \citep{zhang2021object} &ICCV'21 &DiMP &63.3  &92.4  &50.8  &70.4  &53.8   &66.1    \\
    AFNet \citep{zhang2023frame} &CVPR'23 &DiMP &63.8  &92.6  &51.8   &70.8  &52.5  &63.8   \\
    CEUTrack \citep{tang2022revisiting} &Arxiv'22 &OSTrack-B256 &56.7   &84.2  &55.0  &71.6  &62.0 &76.4 \\
    ProTrack \citep{yang2022prompting} &ACM MM'22 &OSTrack-B256 &58.0   &85.5  &47.1  &63.2  &66.1  &78.4  \\
    HRTrack \citep{zhu2023cross} &ICCV'23 &OSTrack-B256 &-   &-  &-  &-  &63.2  &71.9 \\
    ViPT \citep{zhu2023visual}  & CVPR'23 &OSTrack-B256 &\underline{65.1}   &\underline{93.5}  &59.2  &75.8  &67.7  &80.6 \\
    SDSTrack \citep{hou2024sdstrack}  &CVPR'24 &OSTrack-B256  &60.0   &87.2  &\underline{59.7} &\underline{76.7}  &\underline{67.9}  &\underline{80.9}  \\
    UnTrack~\citep{wu2024single} &CVPR'24 &OSTrack-B256 &-    &-  &58.9   &75.6  &-   &-  \\
    \midrule[1.1pt]
    \textbf{Ours}  &- &OSTrack-B256   &\textbf{67.5}   &\textbf{96.5}  &\textbf{61.8}  &\textbf{78.6}  &\textbf{69.6}  &\textbf{83.0}  \\
    \midrule[1.1pt]
    ViPT \citep{zhu2023visual}  &CVPR'23 &DropTrack-B384  &65.6   &93.9  &60.0  &77.4  &68.6  &81.4  \\
    SDSTrack \citep{hou2024sdstrack}  &CVPR'24 &DropTrack-B384  &\underline{65.6}  &\underline{94.4}  &\underline{60.6}  &\underline{78.0}  &\underline{69.0}  &\underline{82.2}  \\
    UnTrack~\citep{wu2024single} &CVPR'24 &DropTrack-B384 &-    &-  &60.0 &77.2  &-  &-  \\
    \midrule[1.1pt]
    \textbf{Ours}  &- &DropTrack-B384   &\textbf{68.7}   &\textbf{97.1}  &\textbf{63.0}  &\textbf{80.3}  &\textbf{71.3}  &\textbf{84.9}  \\
    \midrule[1.1pt]
    MamTrack~\citep{Sun_2025_CVPR} &CVPR'25 &HiViT-B384 &\underline{66.4}  &\underline{94.2}  &61.6 &79.2  &-  &- \\
    SUTrack~\citep{chen2025sutrack} &AAAI'25 &SUTrack-B384 &-  &-  &\underline{63.1} &\underline{79.6}  &-  &- \\
    \midrule[1.1pt]
    \textbf{Ours}  &- &SUTrack-B384   &\textbf{70.8}   &\textbf{98.2}  &\textbf{65.4}  &\textbf{81.5}  &\textbf{73.5}  &\textbf{85.6}  \\
    \bottomrule[1.1pt]
\end{tabular}
\label{tab:1}
\vskip -0.05in
\end{table*}

\subsubsection{Evaluation Metrics}
Adhering to recognized standards~\citep{hou2024sdstrack,wu2024single,hong2024onetracker,Sun_2025_CVPR}, we evaluated tracking performance with the following metrics. For all RGB-Event benchmarks (FE108, VisEvent and CoeSot), we utilize two widely used metrics, i.e., success rate (SR) and precision rate (PR). For DepthTrack, precision (Pr) and recall (Re) are used, with F-score ($F=\frac{2 Re \cdot Pr}{Re + Pr}$) as the primary measure. For VOT-RGBD2022, we assess accuracy (Acc), robustness (Rob), and expected average overlap (EAO). For LasHeR, success rate (SR) and precision rate (PR) are employed. For RGBT234, maximum success rate (MSR) and maximum precision rate (MPR) serve as evaluation metrics. Notably, our SR/PR precisely aligns with SR/PR in ViPT and OneTrack, Suc/Pre in SDSTrack, Success/Precision in UnTrack, or AUC/P in SUTrack. For all reported tracking metrics, the \textbf{larger}, the \textbf{better}.

\subsubsection{Pre-trained Models and Compared Methods}
In this work, we choose three prototypical one-stream RGB-based trackers as pre-trained baselines: OSTrack~\citep{ye2022joint}, the most widely adopted model; DropTrack~\citep{wu2023dropmae} and SUTrack~\citep{chen2025sutrack}, which offer improved generalization and performance. Notably, OSTrack and DropTrack are built upon ViT-B/16~\citep{dosovitskiy2020image}, while SUTrack adopts the HiViT-B/16~\citep{zhang2023hivit} as the backbone. Following their pre-trained configurations, these models differ in input resolution: OSTrack~(128$\times$128/256$\times$256), DropTrack~(192$\times$192/384$\times$384), and SUTrack~(192$\times$192/384$\times$384) for template and search regions, respectively. To comprehensively validate the effectiveness of our method, we conduct the following experiments in Table~\ref{tab:1} and Table~\ref{tab:2}. First, we construct strong RGB-based single-modal baselines under full fine-tuning. We then evaluate SRFT with full fine-tuning and further examine its compatibility with selected PEFT-style configurations. For fair comparison, all methods are grouped according to the pre-trained trackers. \textbf{Notably}, XTrack~\citep{tan2024xtrack}, OneTracker~\citep{hong2024onetracker} and MamTrack~\citep{Sun_2025_CVPR} adopt non-standard experimental settings and lack open-source implementations. XTrack~\citep{tan2024xtrack} and OneTracker use a 384$\times$384 pre-trained OSTrack, incompatible with standard benchmarks, so we compare it separately in Table~\ref{tab:3}. MamTrack, with unclear pretraining details but sharing the HiViT-B/16 backbone with SUTrack, is included in SUTrack-based comparisons. \revise{For the comparisons with state-of-the-art methods, we report published results or reproduce them using official implementations under reported setups, providing system-level comparisons across existing trackers.}

\begin{table*}[htbp]
\captionsetup{skip=3pt}
\caption{Quantitative comparison on the RGB-Depth and RGB-Thermal datasets. \revise{Within each base-model group, the best result is highlighted in
\textbf{bold} and the second-best result is \underline{underlined}. methods built on DiMP/Stark-S are listed for reference only.}}
\centering
\renewcommand{\arraystretch}{1.00}
\setlength{\tabcolsep}{4.0pt}
\small
\begin{tabular}{c |c | c  | c c c | c c c | c c | c c}
    \midrule[1.1pt]
    \multirow{2}{*}{\textbf{Method}} &\multirow{2}{*}{\textbf{Reference}}  &\multirow{2}{*}{\textbf{Base Model}} &\multicolumn{3}{c|}{\textbf{DepthTrack}}  &\multicolumn{3}{c|}{\textbf{VOT
RGBD2022}} &\multicolumn{2}{c|}{\textbf{LasHeR}}  &\multicolumn{2}{c}{\textbf{RGBT234}}  \\
    \cline{4-13}
    \multicolumn{1}{c|}{} &  & &Pr &Re &F-score  &Acc &Rob &EAO  &SR  &PR  &MSR  &MPR   \\
    \midrule[1.1pt]
    \rowcolor{gray!10} \multicolumn{13}{c}{\textbf{Image-based Methods (Only-RGB)}}   \\
    DiMP \citep{bhat2019learning} &ICCV'19 &ResNet &46.3	&42.8  &44.5  &70.3 &73.1 &54.3 &42.8  &54.4  &42.1 &62.5 \\ 
    Stark-S \citep{yan2021learning} &ICCV'21 &ResNet  &39.3	&37.6  &38.4  &65.4 &62.8  &48.2  &37.7 &46.3  &48.9 &66.5 \\
    OSTrack \citep{ye2022joint} &ECCV'22 &OSTrack-B256   &53.8	&53.0  &53.4  &80.3 &83.3 &67.6  &47.2  &58.5  &54.9 &72.9 \\
    DropTrack \citep{wu2023dropmae} &CVPR'23 &DropTrack-B384   &56.4  &55.8  &56.1 &81.5 &85.1  &69.2 &47.7   &59.4   &57.2 &75.8 \\
    \midrule[1.1pt]
      \rowcolor{blue!8} \multicolumn{13}{c}{\textbf{Cross-modal Transfer Learning}}   \\
    SPT \citep{zhu2023rgbd1k} &AAAI'23 &Stark-S  &52.7 &54.9  &53.8  &79.8 &85.1 &65.1 &33.4   &46.7 &55.5 &78.6 \\
    APFNet \citep{xiao2022attribute} &AAAI'22 &Stark-S  &51.6	&51.4  &51.5 &79.0 &83.7 &64.2 &36.2 &50.0 &57.9 &82.7 \\
    ProTrack \citep{yang2022prompting} & ACM MM'22 &OSTrack-B256  &58.3	&57.3  &57.8 &80.1 &80.2 &65.1 &42.0   &53.8 &59.9 &79.5\\
    ViPT \citep{zhu2023visual}  &CVPR'23 &OSTrack-B256   &59.2	&59.6  &59.4  &81.5 &87.1 &72.1  &52.5 &65.1  &61.7 &83.5 \\
    SDSTrack \citep{hou2024sdstrack}  &CVPR'24 &OSTrack-B256   &\underline{61.9}	&\underline{60.9}  &\underline{61.4}  &81.2 &\underline{88.3}  &\underline{72.8} &\underline{53.1}  &\underline{66.5} &\underline{62.5} &\underline{84.8} \\

    UnTrack~\citep{wu2024single} &CVPR'24 &OSTrack-B256 &61.1   &60.8	&61.0  &\underline{82.0}  &86.9 &72.1   &51.3	&63.7  &62.5  &84.2  \\
    \midrule[1.1pt]
    \textbf{Ours}  &- &OSTrack-B256   &\textbf{64.7} &\textbf{65.4}  &\textbf{65.1}  &\textbf{82.1}   &\textbf{88.8}	&\textbf{74.1} &\textbf{56.3}  &\textbf{70.1}   &\textbf{64.4}   &\textbf{87.2} \\
    \midrule[1.1pt]
    ViPT \citep{zhu2023visual}  &CVPR'23 &DropTrack-B384  &62.6	 &61.9  &62.2  &81.7   &87.3	&72.3 &52.5  &65.6  &63.4   &84.7\\
    SDSTrack \citep{hou2024sdstrack} &CVPR'24 &DropTrack-B384  &\underline{63.3} &\underline{62.2}  &\underline{62.7}  &81.4   &\underline{88.6}	&\underline{73.0} &\underline{55.1}  &\underline{69.0}   &\underline{65.0}   &\underline{87.1}\\
    UnTrack~\citep{wu2024single} &CVPR'24 &DropTrack-B384 &62.4   &61.7	&62.0  &\underline{82.1}   &87.1	&72.2   &52.4	 &65.9  &64.1  &85.2  \\
    \midrule[1.1pt]
    \textbf{Ours}  &- &DropTrack-B384  &\textbf{67.2}	&\textbf{67.1}  &\textbf{67.1} &\textbf{82.9}   &\textbf{89.2}	&\textbf{74.5}  &\textbf{58.8}   &\textbf{73.6} &\textbf{66.7}   &\textbf{89.1}\\
    \midrule[1.1pt]
    MamTrack~\citep{Sun_2025_CVPR} &CVPR'25 &HiViT-B384 &-  &-  &- &-  &-  &-  &54.2  &67.4  &62.4 &84.4 \\
    SUTrack~\citep{chen2025sutrack} &AAAI'25 &SUTrack-B384 &\underline{58.8}  &\underline{58.6}  &\underline{58.7} &\underline{83.0}  &\underline{90.6}  &\underline{75.4}  &\underline{60.8} &\underline{75.6}  &\underline{69.2} &\underline{92.1}\\
    \midrule[1.1pt]
    \textbf{Ours}  &- &SUTrack-B384  &\textbf{65.1}  &\textbf{65.2}  &\textbf{65.2} &\textbf{84.1}  &\textbf{92.6}  &\textbf{77.7}  &\textbf{62.9}  &\textbf{77.8}  &\textbf{70.3}  &\textbf{93.3}\\
    \bottomrule[1.1pt]
\end{tabular}
\label{tab:2}
\end{table*}

\subsubsection{Training Details}
We follow the data processing pipeline of SDSTrack~\citep{hou2024sdstrack} across all datasets, converting event data into color-polar event images, with no preprocessing for other auxiliary modalities. The models are trained on 8 NVIDIA 3090Ti GPUs with a batch size of 192 and 30 epochs. Each epoch involves sampling 80k samples. We utilize the AdamW optimizer with a learning rate set to $1 \times 10^{-4}$ and a weight decay set to $10^{-4}$. \revise{Controlled experiments isolate SRFT's impact by keeping all setups (initialization, pipeline, optimizer, schedule, protocol) constant while varying only the evaluated factor. Hyperparameters are tuned on validation sets (or a 20\% training holdout when official splits are absent) before test evaluation.}

\subsection{Comparison with State-of-the-Art Methods}
Extensive comparative analyses are presented in Table~\ref{tab:1}, Table~\ref{tab:2} and Table~\ref{tab:3}, where our method demonstrates excellent performance on all multi-modality tracking benchmarks after incorporating the proposed regularized tuning strategies. The corresponding precision and success plots are illustrated in Fig.~\ref{fig:performance}, further substantiating the quantitative results. Evidently, we can observe that both the RGB-only and the cross-modal trackers are becoming increasingly profitable with pre-trained models. In particular, cross-modal approaches exhibit substantial performance gains, highlighting the complementarity between RGB and auxiliary data under complex conditions. Overall, these system-level results demonstrate the strong performance
of the resulting multi-modality trackers across diverse benchmarks.

\begin{table*}[htbp]
\captionsetup{skip=3pt}
\caption{Quantitative comparison with OneTracker and XTrack.}
\centering
\renewcommand{\arraystretch}{1.0}
\setlength{\tabcolsep}{3.0pt}
\footnotesize
\begin{tabular}{c |c | c | c c |c c c | c c c | c c | c c}
    \midrule[1.1pt]
    \multirow{2}{*}{\textbf{Method}} &\multirow{2}{*}{\textbf{Reference}}  &\multirow{2}{*}{\textbf{Base Model}} &\multicolumn{2}{c|}{\textbf{VisEvent}} &\multicolumn{3}{c|}{\textbf{DepthTrack}}  &\multicolumn{3}{c|}{\textbf{VOT
RGBD2022}} &\multicolumn{2}{c|}{\textbf{LasHeR}}  &\multicolumn{2}{c}{\textbf{RGBT234}}  \\
    \cline{4-15}
    \multicolumn{1}{c|}{} &  & &SR  &PR  &Pr &Re &F-score  &Acc &Rob &EAO  &SR  &PR  &MSR  &MPR   \\
    \midrule[1.1pt]
    OneTracker~\citep{hong2024onetracker} &CVPR'24 &OSTrack-B384  &60.8  &76.7  &60.7   &60.4	&60.9  &81.9  &87.2 &72.7   &53.8	&67.2  &64.2  &85.7  \\
    XTrack~\citep{tan2024xtrack} &ICCV'25 &OSTrack-B384  &60.9  &77.5  &61.8   &62.0	&61.5  &82.1  &88.8 &74.0   &55.7	&69.1  &64.9  &87.4  \\
    \midrule[1.1pt]
    \textbf{Ours}  &- &OSTrack-B384   &\textbf{62.6}  &\textbf{79.4}  &\textbf{65.3} &\textbf{66.1}  &\textbf{65.6}  &\textbf{82.4}   &\textbf{88.9}	&\textbf{74.3} &\textbf{57.0}  &\textbf{71.2}   &\textbf{65.4}   &\textbf{87.9} \\
    \bottomrule[1.1pt]
\end{tabular}
\label{tab:3}
\vskip -0.1in
\end{table*}

\noindent \textbf{Results on RGB-Event Tracking.} As shown in Table~\ref{tab:1} and Table~\ref{tab:3}, our method achieves the best reported performance among the compared trackers across the evaluated RGB-Event benchmarks, achieving the highest precision scores of $98.2\%$, $81.5\%$ and $85.6\%$ on the FE108, VisEvent, and CoeSot datasets, respectively. In particular, on FE108, a dataset characterized by low-light conditions and heavy reliance on event information, our approach surpasses the previous best by a notable margin: $+3.0\%$ in PR and $+2.4\%$ in SR over OSTrack-B256. The full fine-tuning approaches (e.g., CEUTrack, MamTrack) yield limited improvements, while parameter-efficient fine-tuning paradigms (e.g., ViPT, SDSTrack) face performance bottlenecks. The compared full fine-tuning and parameter-efficient approaches show smaller gains under these settings, whereas SRFT achieves consistently stronger performance by explicitly regulating parameter updates during cross-modal adaptation.

\vspace{0.1em}
\noindent \textbf{Results on RGB-Depth Tracking.} As depicted in Table~\ref{tab:2} and Table~\ref{tab:3}, our method achieves the highest reported F-score among the compared
trackers on DepthTrack, obtaining the top performance $67.1\%$ in F-score. Under the OSTrack-B256 setting, our method yields substantial improvements: $+2.8\%$ in Pr, $+4.5\%$ in Re, and $+3.7\%$ in F-score. Similarly, when built on the DropTrack-B384 with richer pre-trained knowledge, our method demonstrates superior performance gains, $+3.9\%$ in Pr, $+4.9\%$ in Re, and $+4.4\%$ in F-score. Furthermore, despite no training on the VOT-RGBD2022 dataset, our method still delivers enhanced performance with a $+2.3\%$ gain in EAO over existing baselines. This result further supports the cross-dataset transferability of the proposed approach in RGB-Depth tracking scenarios.

\vspace{0.1em}
\noindent \textbf{Results on RGB-Thermal Tracking.} As listed in Table~\ref{tab:2} and Table~\ref{tab:3}, our method achieves $70.1\%$ precision and $56.3\%$ success when evaluated on the pre-trained OSTrack-B256. Furthermore, our method effectively unleashes the potential of the DropTrack, yielding substantial improvements of $+4.6\%$ in PR, $+3.7\%$ in SR. Remarkably, with SUTrack as our pre-trained baseline, our method achieves the best performance among the compared methods, reaching  $77.8\%$ in PR and $62.9\%$ in SR. In the cross-dataset evaluation on RGBT234, which is not used for adaptation, our method achieves $93.3\%$ MPR and $70.3\%$ MSR. These results demonstrate consistent improvements across both in-dataset and cross-dataset evaluations within the considered RGB-Thermal tracking setting.

\vspace{0.1em}
\noindent \textbf{Attribute Analysis.} To comprehensively analyze our method, we present a detailed per-attribute comparison in Fig.~\ref{fig:attributes}. Our approach achieves the best or competitive performance across most of the evaluated challenging attributes. Specifically, for motion-related sequences from the VisEvent and CoeSot datasets, our method delivers the best results, highlighting strong robustness against motion-induced degradation. Specifically, it yields precision gains of $5.6$ points under Motion
Blur and $3.0$ points under Camera Motion. On LasHeR, our regularization strategy yields notable improvements under extreme lighting conditions: $+13.4\%$ under Illumination Variation, $+6.2\%$ under Low Illumination, and $+5.3\%$ under Over Exposure. Moreover, our method also maintains superior performance across other challenging attributes such as Out-of-View and Frame Lost. Overall, these results indicate consistent gains across a diverse set of
challenging tracking conditions.

\subsection{Ablation Study and Analysis} \label{sec:ablation}
We conducted ablative analyses to verify the effectiveness and characteristics of significance-aware regularization tuning. In the ablation section, “full fine-tuning” refers to training the entire backbone, excluding the box head (as in Section~\ref{sec:preliminaries}). All methods use the pre-trained OSTrack-B256 weights unless specified otherwise.

\vspace{0.1em}
\noindent \textbf{Effectiveness of Prior and Transfer Significance.} We conduct a series of comprehensive experiments to uncover the interplay and effectiveness of the two proposed significance components, as summarized in Table~\ref{tab:ablation}. Comparing \textbf{(a)} and \textbf{(b)} demonstrates that full fine-tuning substantially improves target-domain tracking performance over zero-shot evaluation. Further, the contrast between \textbf{(b)} and \textbf{(c)} (or \textbf{(d)} and \textbf{(e)}) shows that the prior significance regularization significantly improves SR by $+2.4\%$ and PR by $+3.2\%$ on LasHeR, highlighting the benefit of source-sensitivity-aware update regularization. Similarly, comparing \textbf{(b)} and \textbf{(d)} (or \textbf{(c)} and \textbf{(e)}) reveals that transfer-significance-based parameter-update modulation consistently improves target-domain tracking performance. Notably, the simultaneous application of both prior and transfer significance yields greater improvements than either method alone, indicating that the two significance components provide complementary benefits. While the transfer significance penalty alone yields only modest improvement on FE108, it still contributes to the overall leading performance of our method.
\begin{table}[t]
\captionsetup{skip=2pt}
\caption{Ablative study results of the proposed key components. \textbf{“FFT”} refers to  full
fine-tuning of the backbone; \textbf{“PS”} represents prior significance; \textbf{“TS”} is transfer significance. \textbf{(a)} is the zero-shot performance; \textbf{(b)} serves as the full-fine-tuning baseline; \textbf{(e)} is the complete SRFT configuration.}
\centering
\renewcommand{\arraystretch}{1.0}
\setlength{\tabcolsep}{3.0pt}
\small
\begin{tabular}{c |c c c | c c| c c c |c c}
   \toprule
    \multirow{2}{*}{\textbf{Exp.}} &\multirow{2}{*}{\textbf{FFT}} &\multirow{2}{*}{\textbf{PS}} &\multirow{2}{*}{\textbf{TS}} &\multicolumn{2}{c|}{\textbf{FE108}} &\multicolumn{3}{c|}{\textbf{DepthTrack}} &\multicolumn{2}{c}{\textbf{LasHeR}}  \\
    \cline{5-11}
    \multicolumn{1}{c|}{} &  & & &SR &PR &Pr  &Re &F-score  &SR &PR \\
    \hline
    \textbf{(a)}  &  & &  &48.8 &75.2  &38.2 &36.0  &37.1  &36.7  &43.5    \\
    \textbf{(b)}  &$\checkmark$ &  &  &65.2 &93.5  &61.6 &61.4  &61.5  &53.2   &65.8 \\
    \textbf{(c)}  &$\checkmark$ &$\checkmark$  &  &66.8  &95.4   &64.1 &64.5  &64.3  &55.6  &69.0   \\
    \textbf{(d)}  &$\checkmark$ &  &$\checkmark$  &66.6  &94.9  &63.9 &64.2  &64.1  &55.1  &68.4   \\
    \textbf{(e)}  &$\checkmark$ &$\checkmark$  &$\checkmark$ &67.4  &96.5  &64.7 &65.4  &65.1  &56.3  &70.1 \\
    \bottomrule
\end{tabular}
\label{tab:ablation}
\end{table}

\begin{table}[t]
\captionsetup{skip=2pt}
\caption{Ablation analysis of the retained curvature-response count $K$ for prior-significance estimation.}
\centering
\renewcommand{\arraystretch}{1.0}
\setlength{\tabcolsep}{6.0pt}
\small
\begin{tabular}{c | c c|c c c|c c}
    \toprule
    \multirow{2}{*}{\textbf{Exp.}} &\multicolumn{2}{c|}{\textbf{FE108}} &\multicolumn{3}{c|}{\textbf{DepthTrack}} &\multicolumn{2}{c}{\textbf{LasHeR}}  \\  \cline{2-8}
    \multicolumn{1}{c|}{} &SR &PR &Pr  &Re &F-score  &SR &PR \\
    \hline
    $K$=1   &66.5  &95.0   &63.9 &64.3  &64.1  &55.2  &68.7   \\
    $K$=2  &66.6  &95.2   &64.0 &64.4  &64.2  &55.4  &68.8     \\
    $K$=5  &66.8 &95.3  &64.1 &64.4  &64.2  &55.5   &69.0  \\
    $K$=10   &66.8  &95.4   &64.1 &64.5  &64.3  &55.6  &69.0 \\
    $K$=20   &66.4  &95.2   &64.1 &64.4  &64.2  &55.4  &68.9 \\
   \bottomrule
\end{tabular}
\label{tab:eigenvalue_count}
\end{table}

\begin{figure}[htbp]
\centering
\includegraphics[clip, width=0.45\textwidth]{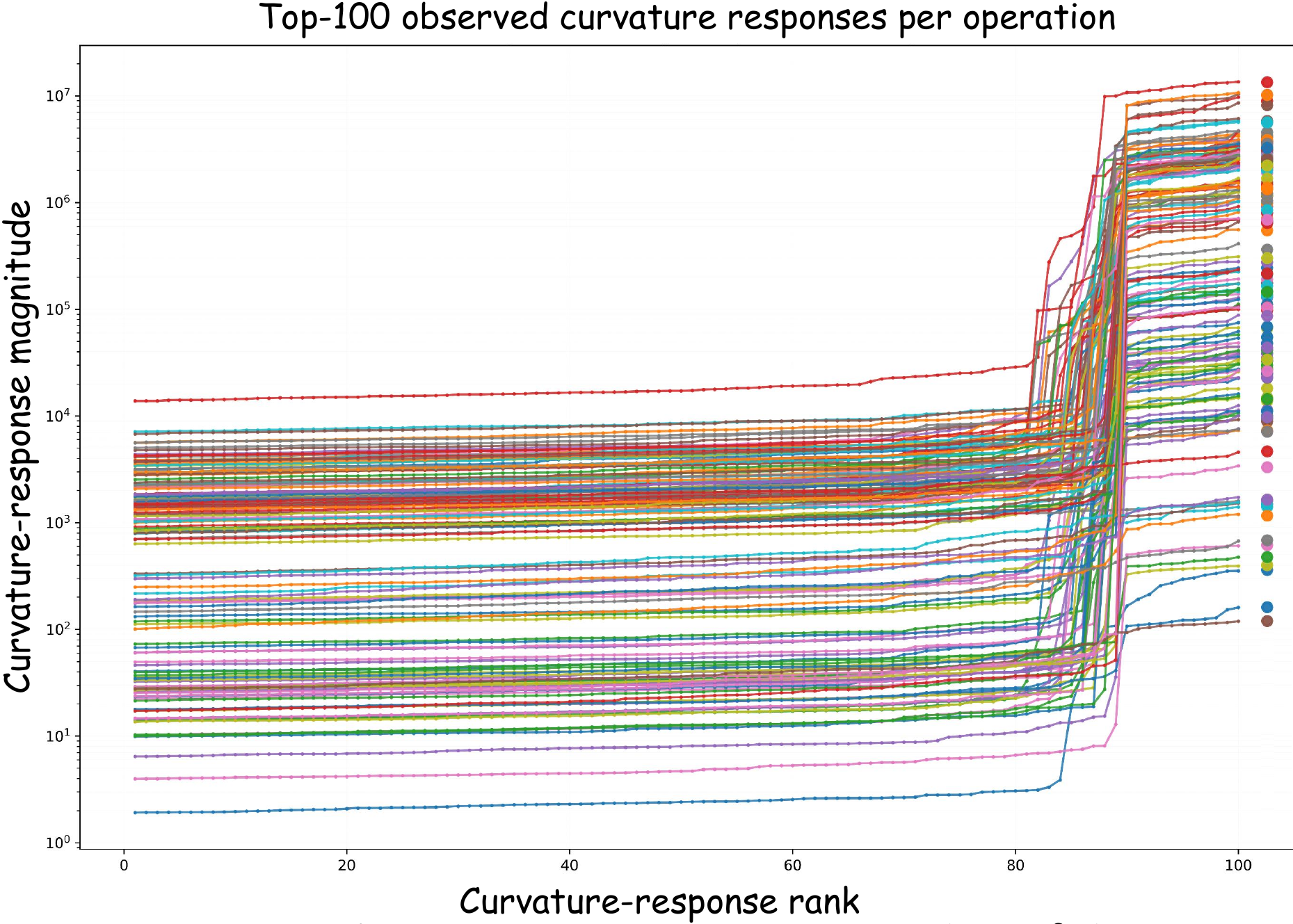}
\vskip -0.05in
\caption{\revise{\textbf{Operation-wise observed curvature responses} for prior-significance estimation. The probe-based directional curvature responses of each operation in OSTrack-B256 are sorted and plotted on a logarithmic scale. The dominant observed responses exhibit a relatively stable high-response region, suggesting that the operation-wise sensitivity estimate is not strongly affected by retaining additional lower-ranked responses.}}
\label{fig:eigenvalue}
\vskip -0.1in
\end{figure}

\vspace{0.2em}
\noindent \revise{\textbf{Top-$K$ Curvature Responses for Prior Significance.} \revise{In this section, we examine the effect of the number of retained curvature responses $K$ in Eq.~(\ref{eq:prior_significance}).} For each operation, we record the directional curvature responses obtained from Rayleigh-quotient probing and compute prior significance using the top-$K$ observed responses. As illustrated in Fig.~\ref{fig:eigenvalue}, the operation-wise
curvature-response profiles exhibit similar dominant-response patterns
across operations. Consistent with this observation, Table~\ref{tab:eigenvalue_count} shows that varying $K$ from 1 to 20 has
only a minor impact: $K=5$ performs on par with the default $K=10$,
and even $K=1$ still yields a clear improvement over the baseline. These results indicate that prior-significance estimation is robust
to the retained observed-response budget $K$ once the dominant curvature
responses are captured. Here, $K$ controls how many observed responses
are retained to form the operation-level significance statistic; it does
not represent a progressive reconstruction of the Hessian spectrum.}

\vspace{0.3em}
\noindent \textbf{Significance Harmony Coefficient $\kappa$.} In this section, we evaluate the impact of different significance harmony coefficients, $\kappa$ as defined in Eq.~(\ref{eq:combine_significance}), with the detailed results presented in Table~\ref{tab:coefficient}. By systematically increasing $\kappa$ from 0.1 to 0.9, we observe that a moderately large coefficient (e.g., 0.6) yields the best performance, outperforming both smaller (e.g., 0.3) and larger values (e.g., 0.9). The results exhibit a broad
optimum around $\kappa=0.5$--$0.7$, with $\kappa=0.6$ providing the best
overall performance. This indicates that neither prior significance nor
transfer significance should dominate the combined significance, and that
a moderate balance between source-knowledge preservation and target-domain
adaptation is beneficial for fine-tuning. \textit{Additional significance fusion scheduling strategies are provided in \textbf{Appendix~\ref{sec:sign_fusion}}.}

\begin{table}[t]
\captionsetup{skip=2pt}
\caption{Ablation analysis of significance harmony coefficient on the FE108 dataset.}
\centering
\renewcommand{\arraystretch}{1.0}
\setlength{\tabcolsep}{4.5pt}
\small
\begin{tabular}{c | c c c c c c c c c}
    \toprule
    \textbf{$\kappa$}  &0.1 &0.2  &0.3   &0.4  &0.5  &0.6  &0.7   &0.8  &0.9 \\
    \hline
    SR &66.5 &66.7 &66.9   &66.9  &67.2  &67.4  &67.1   &66.8 &66.9 \\
    PR &94.7 &94.8  &95.1   &95.3  &96.0  &96.5  &96.2   &95.8  &95.5 \\
    \bottomrule
\end{tabular}
\label{tab:coefficient}
\end{table}

\vspace{0.3em}
\noindent\textbf{Impact of Box Head Tuning.} In our default setting, “full fine-tuning” updates the backbone while keeping the box head frozen. To assess the impact of tuning the box head, we also examine a more aggressive variant, complete fine-tuning, where the box head is updated as well. As shown in Table~\ref{tab:box-head}, unfreezing the box head further exacerbates overfitting. While our regularization method remains effective, the results reveal an important caveat: unfreezing the box head during cross-modal adaptation can be risky, suggesting that updating the pre-trained box head can adversely affect the source-domain tracking behavior retained by this component.

\begin{table}[t]
\captionsetup{skip=2pt}
\caption{Ablation results of box head tuning. \textbf{“Box”} denotes tuning the box head, \textbf{“S-Reg”} represents the significance-aware regularization tuning method.}
\centering
\renewcommand{\arraystretch}{1.0}
\setlength{\tabcolsep}{3.5pt}
\small
\begin{tabular}{c | c| c c|c c c|c c}
    \toprule
    \multirow{2}{*}{\textbf{Exp.}} &\multirow{2}{*}{\textbf{Box}} &\multicolumn{2}{c|}{\textbf{FE108}} &\multicolumn{3}{c|}{\textbf{DepthTrack}} &\multicolumn{2}{c}{\textbf{LasHeR}}  \\  \cline{3-9}
    \multicolumn{1}{c|}{} & &SR &PR &Pr  &Re &F-score  &SR &PR \\
    \hline
    FFT   &w/ &63.8 &91.0  &59.4 &58.7  &59.0  &51.5   &64.4   \\
    FFT  &w/o &65.2 &93.5  &61.6 &61.4  &61.5  &53.2   &65.8     \\
    \hline
    FFT+\textbf{S-Reg}  &w/ &66.2 &95.3  &62.6 &63.4  &63.0  &54.4   &67.7  \\
    FFT+\textbf{S-Reg}   &w/o &67.4  &96.5  &64.7 &65.4  &65.1  &56.3  &70.1 \\
   \bottomrule
\end{tabular}
\label{tab:box-head}
\end{table}

\vspace{0.2em}
\noindent \textbf{Compatibility with PEFT Methods.} The evaluated PEFT methods~\cite{zhu2023visual,hou2024sdstrack,wu2024single} freeze most pre-trained parameters and optimize a small set of additional parameters. We investigate whether significance-aware regularization remains beneficial when their pre-trained backbones are additionally unfrozen for adaptation. To evaluate the compatibility of our proposed regularization techniques with these methods, we unfreeze their backbones and retrain them with our regularization applied. As shown in Table~\ref{tab:compat-peft}, our approach significantly enhances ViPT's performance on VisEvent, yielding gains of $+2.1\%$ in SR and $+2.0\%$ in PR. Notably, even for the unified tracker UnTrack, our approach achieves superior results on LasHeR: $+2.7\%$ in SR and $+3.7\%$ in PR. This demonstrates that overly rigid constraints on pre-trained models limit their transfer potential. For SDSTrack, SRFT partially recovers the performance degradation caused
by unfreezing the backbone, but does not surpass the original frozen-backbone
configuration. This suggests that the benefit of SRFT depends on the
underlying adaptation architecture, particularly for methods whose
modal-specific adapters are designed to operate with a largely frozen
pre-trained backbone.

\begin{table}[t]
\captionsetup{skip=2pt}
\caption{Compatibility study results of the proposed regularization tuning method on existing PEFT methods. \textbf{“FFT”} denotes full fine-tuning of the backbone, while \textbf{“S-Reg”} represents the significance-aware regularization tuning scheme.}
\centering
\renewcommand{\arraystretch}{1.0}
\setlength{\tabcolsep}{2.2pt}
\small
\begin{tabular}{c | c c|c c c|c c}
    \toprule
    \multirow{2}{*}{\textbf{Exp.}} &\multicolumn{2}{c|}{\textbf{VisEvent}} &\multicolumn{3}{c|}{\textbf{DepthTrack}} &\multicolumn{2}{c}{\textbf{LasHeR}}  \\  \cline{2-8}
    \multicolumn{1}{c|}{} &SR &PR &Pr  &Re &F-score  &SR &PR \\
    \hline
    ViPT &59.2  &75.8  &59.2  &59.6 &59.4  &52.5  &65.1    \\
    ViPT+\textbf{FFT} &57.5  &74.0  &58.2  &57.4 &57.8  &50.9  &63.8 \\
    ViPT+\textbf{FFT}+\textbf{S-Reg} &61.3  &77.8  &61.9  &61.4 &61.6  &54.9  &68.4   \\
    \hline
        UnTrack &58.9  &75.6  &61.1 &60.8 &61.0  &51.3  &63.7 \\
    UnTrack+\textbf{FFT} &56.6  &73.1  &58.2  &56.8 &57.5  &48.1  &60.2 \\
    UnTrack+\textbf{FFT}+\textbf{S-Reg} &60.2  &76.9  &61.9  &62.5 &62.2  &54.0  &67.4 \\
    \hline
    SDSTrack &59.7  &76.7  &61.9  &60.9 &61.4  &53.1  &66.5 \\
    SDSTrack+\textbf{FFT} &56.0  &72.0  &57.8  &56.4 &57.1  &50.7  &63.8 \\
    SDSTrack+\textbf{FFT}+\textbf{S-Reg} &57.8  &74.8  &59.5  &58.3 &58.9  &52.5  &65.8 \\
    \hline
    Ours    &61.8  &78.6  &64.7  &65.4 &65.1  &56.3  &70.1  \\
    \bottomrule
\end{tabular}
\label{tab:compat-peft}
\vskip -0.1in
\end{table}

\vspace{0.2em}
\noindent \textbf{Compatibility with Single-modality Inputs.} This work aims to mitigate the misfitting issue when adapting pre-trained RGB trackers to downstream tasks. A central question is how the proposed regularization methods perform on single-modality data. To investigate this, we conduct ablation studies highlighting their impact on different modalities. As shown in Table~\ref{tab:compat-single}, both RGB and auxiliary modalities benefit significantly from our regularization techniques. For example, the RGB and depth modalities achieve F-score gains of $+5.3\%$ and $+4.0\%$, respectively, on the DepthTrack dataset. Despite substantial distribution differences, our method consistently improves tracking performance for both RGB-only and auxiliary-modality inputs across the evaluated datasets. These findings further support the benefit of significance-aware update regularization when adapting pre-trained trackers to downstream domains.

\begin{table}[t]
\captionsetup{skip=2pt}
\caption{Compatibility study results of the proposed regularization tuning method on single-modality data. ``RGB'' and ``Auxiliary'' denote training and evaluation using
only the corresponding modality. \textbf{“SRFT”} represents the significance-aware regularization tuning method.}
\centering
\renewcommand{\arraystretch}{1.0}
\setlength{\tabcolsep}{2.5pt}
\small
\begin{tabular}{c | c c|c c c|c c}
    \toprule
    \multirow{2}{*}{\textbf{Exp.}} &\multicolumn{2}{c|}{\textbf{CoeSot}} &\multicolumn{3}{c|}{\textbf{DepthTrack}} &\multicolumn{2}{c}{\textbf{LasHeR}}  \\  \cline{2-8}
    \multicolumn{1}{c|}{} &SR &PR &Pr  &Re &F-score  &SR &PR \\
    \hline
    RGB    &64.3 &76.3  &53.9  &53.0 &53.4   &47.2  &58.6   \\
    RGB+\textbf{SRFT}    &68.0  &80.0  &58.8  &58.6 &58.7   &50.3  &62.6   \\
    \hline
    Auxiliary  &57.5  &69.8  &49.0  &47.3 &48.1   &42.7  &53.7 \\
    Auxiliary+\textbf{SRFT}  &60.5  &73.7  &52.9  &51.3 &52.1  &45.9  &57.8 \\
   \bottomrule
\end{tabular}
\label{tab:compat-single}
\vskip -0.1in
\end{table}

\begin{figure*}[htbp]
\centering
\includegraphics[clip, width=0.98\textwidth]{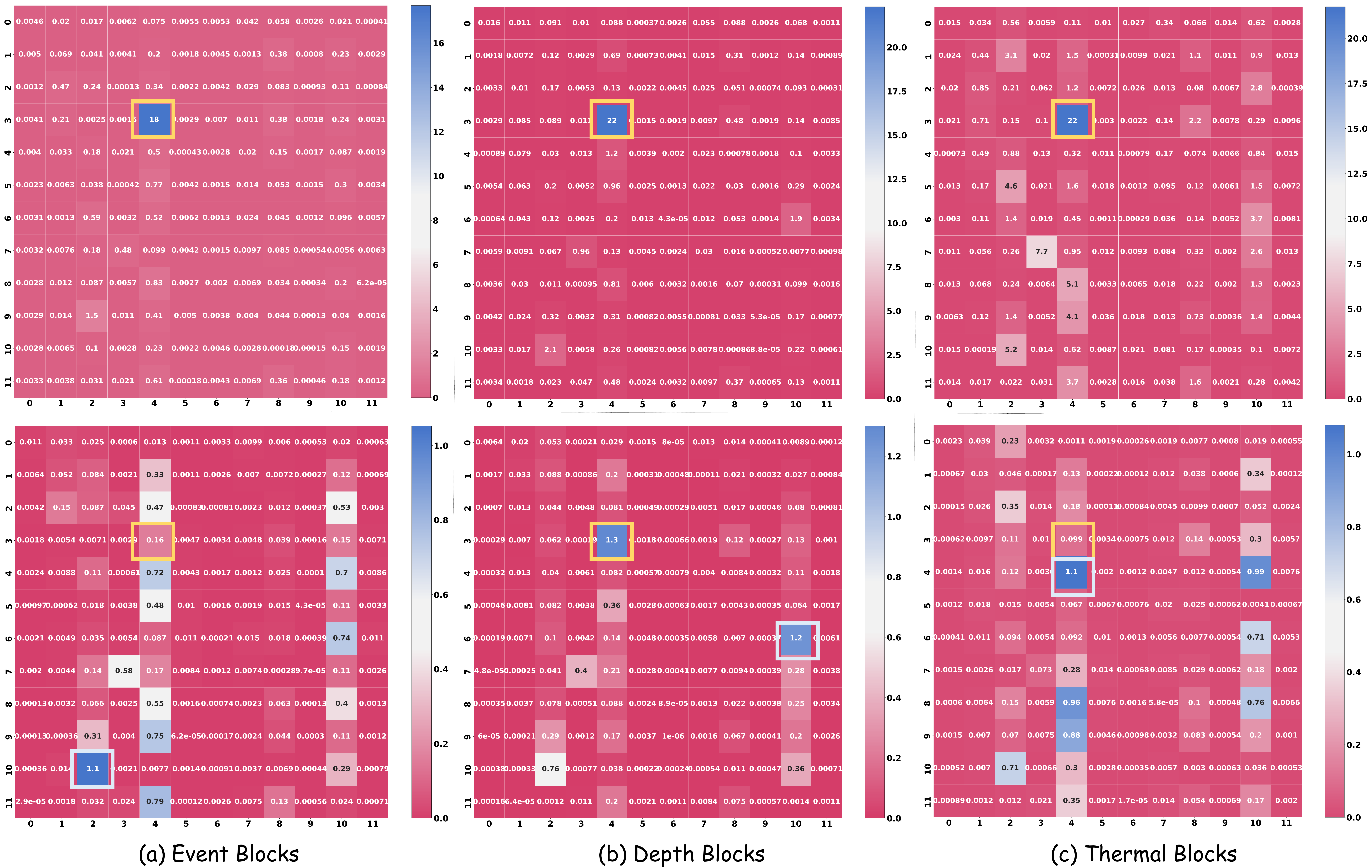}
\vskip -0.1in
\caption{\textbf{Operation-wise weight distances $\frac{\left\| \theta_t- \theta_0\right\|_2}{\left\|\theta_0\right\|_2}$ between the tuned model and the pre-trained model after training}, including results from vanilla full fine-tuning (\textbf{upper}) and our regularized tuning (\textbf{bottom}). The auxiliary branches are derived from the VisEvent, DepthTrack, and LasHeR datasets, respectively.}
\label{fig:weight_diff_operation}
\vskip -0.1in
\end{figure*}

\vspace{0.2em}
\noindent \textbf{Impact of Input Event Representations.} In this work, we focus on constructing suitable frame-like event representations tailored to cross-modal transfer. In this experiment, we keep the same training setup, varying only the input event representations. As shown in Table~\ref{tab:representation}, surface-based representations (TSLTD, Time Surface) yield lower
tracking accuracy under our transfer setting. In contrast, count-based representations (Event Count, Event Frame) offer more robust performance. The time-interpolation-based Event Volume further improves results by utilizing time-weighted and multi-channel methods to preserve spatio-temporal information. Notably, we employ simple color-polarity event images (with ViPT) to align with the RGB pre-trained model, leading to superior performance.

\begin{table}[t]
\captionsetup{skip=2pt}
\caption{Ablation results of different event representations.}
\centering
\renewcommand{\arraystretch}{1.1}
\setlength{\tabcolsep}{2.0pt}
\small
\begin{tabular}{c | c c| c c|c c}
    \toprule
    \multirow{2}{*}{\textbf{Exp.}} &\multicolumn{2}{c|}{\textbf{FE108}} &\multicolumn{2}{c|}{\textbf{VisEvent}} &\multicolumn{2}{c}{\textbf{CoeSot}}  \\  \cline{2-7}
    \multicolumn{1}{c|}{} &SR &PR &SR &PR  &SR &PR \\
    \hline
    Event Frame~\cite{rebecq2017real}   &66.8  &95.5  &61.2  &77.6  &68.8   &82.0   \\
    Event Count~\cite{maqueda2018event} &66.3 &95.1  &59.8  &76.9  &67.9   &81.2   \\
    Time Surface~\cite{sironi2018hats}  &66.1  &94.9  &60.3 &77.1  &67.4  &80.6 \\
    TSLTD~\cite{chen2020end} &66.5 &95.2  &61.0  &77.4  &67.9   &81.3 \\
    Event Volume~\cite{zhu2019unsupervised} &67.3 &96.1  &61.4  &77.9  &69.1   &82.4 \\
    \hline
    Color-Polar Event Image (Ours)   &67.5  &96.5  &61.8 &78.6  &69.6  &83.0 \\
   \bottomrule
\end{tabular}
\label{tab:representation}
\vskip -0.1in
\end{table}

\begin{table}[htbp]
\captionsetup{skip=2pt}
\caption{Ablation analysis of smaller learning rates $\alpha$ on the LasHeR dataset.}
\centering
\renewcommand{\arraystretch}{1.0}
\setlength{\tabcolsep}{4.0pt}
\small       
\begin{tabular}{c |c c c c}
   \toprule
    \textbf{Exp.} &$\alpha=10^{-4}$ &$\alpha=10^{-5}$ & $\alpha=10^{-6}$ & Ours \\
    \hline 
     SR &53.2  &52.9 &52.5  &56.3    \\ 
     PR &65.8  &66.2 &65.4  &70.1     \\ 
    \bottomrule
\end{tabular}
\label{tab:small}
\vskip -0.1in
\end{table}

\begin{figure}[htbp]
\centering
\includegraphics[clip, width=0.48\textwidth]{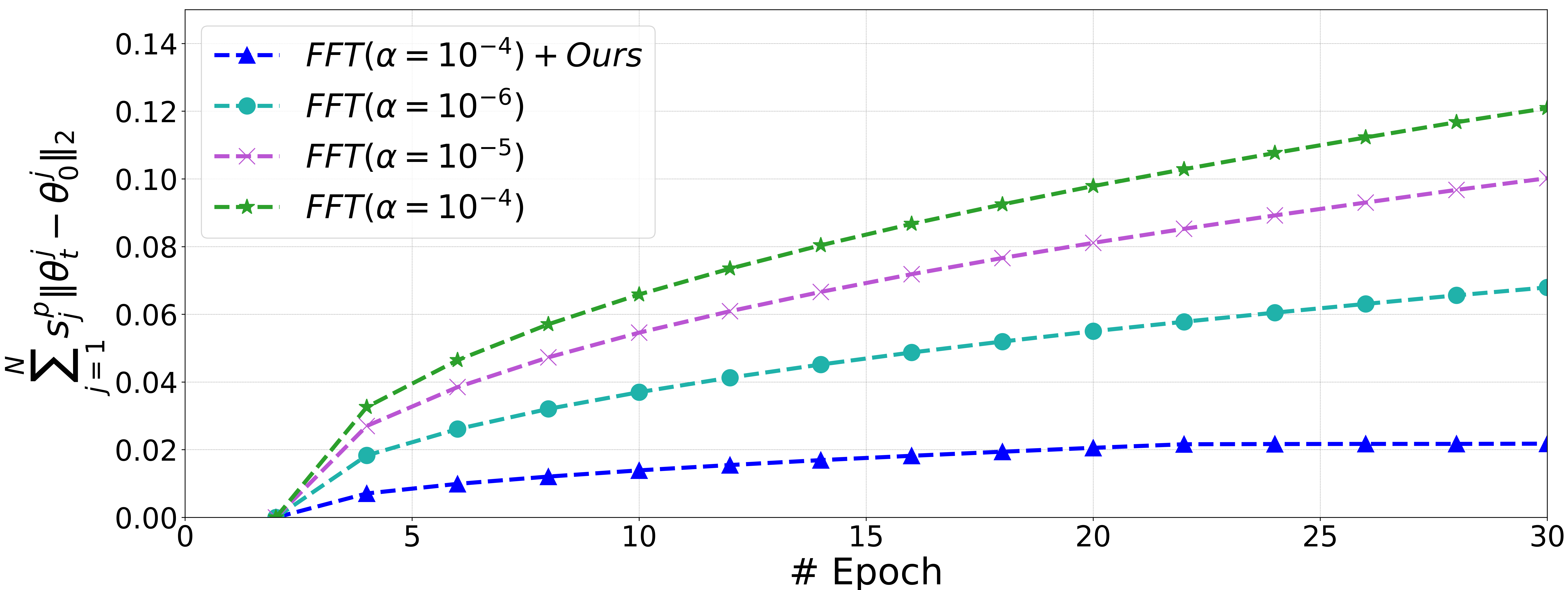}
\vskip -0.1in
\caption{\textbf{Significance-weighted weight distances} between our regularized tuning method and the full fine-tuning (with different learning rates) on the DepthTrack.}
\label{fig:weight_diff_time}
\vskip -0.2in
\end{figure}

\noindent \textbf{Effectiveness of Smaller Learning Rates.} In this paper, SRFT modulates parameter updates according to prior and transfer significance. A natural alternative is to uniformly reduce the learning rate. To investigate this, we evaluate the effect of reduced learning rates. As shown in Table~\ref{tab:small}, a smaller learning rate (i.e., $\alpha=10^{-5}$) yields negligible improvement in PR, but at the cost of a decline in SR. Further reduction (i.e., $\alpha=10^{-6}$) leads to overall performance degradation. These findings suggest that merely reducing the learning rate is insufficient to address over-fitting in cross-modal adaptation, while simultaneously limiting transfer potential, as it uniformly suppresses parameter updates without prioritizing those that are sensitive or high-risk.

\vspace{0.2em}
\noindent \textbf{Observations on Weight Variations.} To further analyze the effect of significance-aware regularization on parameter updates, we visualize parameter dynamics during training from both spatial and temporal perspectives, as depicted in Fig.~\ref{fig:weight_diff_operation} and Fig.~\ref{fig:weight_diff_time}. Overall, our regularization encourages smaller and more broadly distributed parameter deviations than vanilla FFT, consistent with its intended role of suppressing excessive updates to highly significant parameters. As shown in Fig.~\ref{fig:weight_diff_operation}, for parameters with high prior significance, such as \texttt{blocks.3.attn.proj.weight} (highlighted in the orange box), our method effectively suppresses excessive updates. As further demonstrated in Fig.~\ref{fig:weight_diff_time}, our method reaches a stable parameter-deviation level earlier and achieves a substantial reduction in significance-weighted weight deviations (e.g., up to $80\%$), indicating stronger suppression of deviations in parameters assigned high prior significance while maintaining target-task adaptation. Under vanilla fine-tuning, parameter deviations are more concentrated in a limited set of regions, whereas significance-aware regularization yields a more distributed update pattern. After applying the significance-aware penalties, the parameter updates become more balanced and distributed, effectively decentralizing the adaptation burden from a few parameters to a broader set. This redistribution is consistent with the intended effect of avoiding excessive reliance on a small subset of parameter updates.

\begin{table}[t]
\captionsetup{skip=4pt}
\caption{Computational complexity and speed analysis on the LasHeR dataset. The base trackers are denoted by abbreviations (e.g., OS/Drop/SU). Model size is reported in millions of parameters (M). \revise{``Ours'' denotes the
complete architecture in Fig.~\ref{fig:architecture}, which employs
symmetric backbones.
``+SRFT'' denotes unfreezing the corresponding backbone
parameters and applying significance-regularized fine-tuning.}}
\centering
\renewcommand{\arraystretch}{1.1}
\setlength{\tabcolsep}{1.3pt}
\footnotesize 
\begin{tabular}{c | c | c c c| c c}
    \toprule
    Method &Model  &Params &GFLOPs  &FPS  &SR  &PR \\
    \hline
    OSTrack~\cite{ye2022joint}   &OS  &92.1  &58.1  &98.4  &47.2  &58.5    \\
    ProTrack~\cite{yang2022prompting}   &OS  &92.7  &58.4  &92.3  &42.0  &53.8   \\
    ViPT~\cite{zhu2023visual}    &OS  &92.9  &59.9  &88.6  &52.5  &65.1 \\
    SDSTrack~\cite{hou2024sdstrack}   &OS  &102.1  &108.7  &44.6  &53.1  &66.5 \\
    UnTrack~\cite{wu2024single}   &OS  &98.7  &62.6  &75.6  &51.3  &63.7 \\
    ViPT+\revise{SRFT}  &OS  &92.9  &59.9  &88.6  &54.9  &68.4\\
    UnTrack+\revise{SRFT}  &OS  &98.7  &62.6  &75.6  &54.0  &67.4 \\
    Ours  &OS  &202.0  &149.0  &49.0  &56.3  &70.1 \\
    \hline
    DropTrack~\cite{wu2023dropmae}   &Drop  &92.1  &130.6  &57.6  &47.7  &59.4   \\
    ViPT~\cite{zhu2023visual}  &Drop  &92.9  &131.9  &48.8  &52.5  &65.6 \\
    SDSTrack~\cite{hou2024sdstrack}  &Drop  &102.1  &244.2  &26.8  &55.1  &69.0 \\
    UnTrack~\cite{wu2024single}  &Drop  &98.7  &138.2  &42.3  &52.4  &65.9 \\
    ViPT+\revise{SRFT}  &Drop   &92.9  &131.9  &48.8  &57.1 &70.5 \\
    UnTrack+\revise{SRFT}  &Drop  &98.7  &138.2  &42.3  &56.3  &69.4 \\
    Ours  &Drop  &202.0  &335.3  &29.2  &58.8  &73.6 \\
    \hline
    SUTrack~\cite{chen2025sutrack}  &SU  &65.29  &106.9  &37.6  &60.8  &75.6   \\
    Ours  &SU &141.1  &243.7  &24.8 &62.9  &77.8 \\
    \bottomrule
\end{tabular}
\label{tab:efficiency}
\vskip -0.1in
\end{table}

\subsection{Computational Efficiency}
In addition to tracking accuracy, we evaluate the computational efficiency of our significance-aware regularization tuning by analyzing both training and inference overhead.

\vspace{0.2em}
\noindent \textbf{Computational Overhead of Prior Significance.} 
Our method requires an offline estimation of prior significance through operation-wise source-loss curvature probing, which introduces a non-trivial preprocessing cost compared to purely plug-and-play PEFT baselines. This overhead can become a practical barrier when scaling to larger backbones or when frequent architectural changes demand repeated recomputation. Nonetheless, this cost is paid once and does not materially affect the per-iteration training and inference runtime after the significance is computed (\revise{\textit{details in the \textbf{Appendix~\ref{app:prior_efficiency}, \ref{app:training_efficiency}, \ref{app:probing_efficiency}}})}.

\vspace{0.2em}
\noindent \textbf{Efficiency of SRFT.} We compare the training efficiency of our SRFT against vanilla FFT. Since prior significance is computed once as a preprocessing step, we exclude it from per-iteration training time. SRFT introduces online transfer-significance estimation with marginal computational overhead, as this significance is computed directly from the gradients already available during optimization: 39.0 s/iter vs. 37.5 s/iter for vanilla FFT (+$4.0\%$).

\vspace{0.2em}
\noindent \textbf{Inference Cost and Speed.} We compare the computational complexity and runtime efficiency of representative baselines, focusing on trackers instantiated with three widely used pre-trained backbones: OSTrack, DropTrack and SUTrack. For a fair comparison, we report both theoretical cost (e.g., parameter count and FLOPs) and practical throughput (FPS) under a unified evaluation protocol. All experiments are conducted on the same hardware platform, equipped with an Intel(R) Xeon(R) 6456C CPU, 256 GB RAM, and a single NVIDIA RTX 3090Ti GPU. As reported in Table~\ref{tab:efficiency}, our method operates at 24.8 FPS while delivering top-tier accuracy under the SUTrack setting. When integrated into ViPT (“ViPT+\revise{SRFT}”), it yields improved accuracy while preserving the same inference speed and complexity. Under the DropTrack base model, the resulting tracker still operates at 48.8 FPS, maintaining a respectable speed given its focus on accuracy. \revise{
Table~\ref{tab:efficiency} reports system-level
accuracy--efficiency rather than serving as a controlled attribution
of the gains to SRFT. The larger parameter count and FLOPs of ``Ours''
arise from the complete dual-stream architecture, while SRFT itself
introduces no additional inference modules, parameters, or FLOPs.}

\subsection{Limitations and Discussion} 
From a computational-efficiency perspective, a practical consideration is that prior-significance estimation through operation-wise source-loss curvature probing remains an additional offline procedure compared with plug-and-play PEFT. \revise{Although reduced probing substantially decreases this cost (details in \textit{\textbf{Appendix \ref{app:probing_efficiency}}}), the additional preprocessing cost may remain non-negligible for larger backbones, frequent architecture changes, or rapid ablation cycles. Moreover, SRFT requires access to source-domain tracking data $D_0$ for
prior-significance estimation. Although the complete source RGB tracking set may not always be available, our experiments in
\textit{\textbf{Appendix~\ref{app:proxy_data}}} show that a subset of the source-domain tracking data can serve as a practical proxy and retain
most of the performance benefit. Nevertheless, SRFT in its current form is not data-free, and developing
fully data-free source-sensitivity estimation remains an important direction
for future work.}

\section{Conclusion} \label{sec5}
This study revisits the critical misfitting issues encountered when adapting
pre-trained RGB trackers to multi-modality tracking tasks. By introducing two
complementary forms of parameter significance to characterize source-knowledge
sensitivity and target-domain adaptability, respectively, SRFT adaptively
modulates parameter updates to balance knowledge preservation and downstream
adaptation. Extensive experiments across various multi-modality tracking
benchmarks demonstrate the effectiveness of SRFT, with controlled ablations
further validating the contribution of significance regularization. More broadly, a key insight from our study highlights the importance of significance-aware fine-tuning for
balancing generalization and adaptability in cross-domain transfer. We believe these insights will pave the way for further advancements in multi-modality transfer learning, particularly in the context of scene perception tasks. \revise{Our empirical validation is restricted to RGB-to-RGB-X
tracking, and we do not claim empirical generalization beyond this setting.
Nevertheless, SRFT is not intrinsically tied to a specific modality, and
extending it to unsupervised-pretraining-to-tracking transfer and broader
downstream fine-tuning scenarios remains an important direction for future
work.}

\section*{Data Availability}
This work does not propose a new dataset. All the datasets we used are publicly available and well cited in the paper. FE108: \url{https://zhangjiqing.com/dataset/}; VisEvent: \url{https://sites.google.com/view/viseventtrack/}; CoeSot: \url{https://github.com/Event-AHU/COESOT}; DepthTrack: \url{https://github.com/xiaozai/det}; VOT RGBD2022: \url{https://www.votchallenge.net/vot2022/dataset.html}; LasHeR: \url{https://github.com/BUGPLEASEOUT/LasHeR}; RGBT234: \url{https://github.com/xuboyue1999/RGBT-Tracking/tree/main}.

\section*{Conflict of Interest}
The authors affirm that there are no commercial or associative relationships that could be perceived as a conflict of interest related to the submitted work.

\bibliographystyle{spbasic}
\bibliography{reference}

@inproceedings{yan2021learning,
  title={Learning spatio-temporal transformer for visual tracking},
  author={Yan, Bin and Peng, Houwen and Fu, Jianlong and Wang, Dong and Lu, Huchuan},
  booktitle={Proceedings of the IEEE/CVF international conference on computer vision},
  pages={10448--10457},
  year={2021}
}

@inproceedings{ye2022joint,
  title={Joint feature learning and relation modeling for tracking: A one-stream framework},
  author={Ye, Botao and Chang, Hong and Ma, Bingpeng and Shan, Shiguang and Chen, Xilin},
  booktitle={European Conference on Computer Vision},
  pages={341--357},
  year={2022},
  organization={Springer}
}

@inproceedings{wu2023dropmae,
  title={Dropmae: Masked autoencoders with spatial-attention dropout for tracking tasks},
  author={Wu, Qiangqiang and Yang, Tianyu and Liu, Ziquan and Wu, Baoyuan and Shan, Ying and Chan, Antoni B},
  booktitle={Proceedings of the IEEE/CVF Conference on Computer Vision and Pattern Recognition},
  pages={14561--14571},
  year={2023}
}

@article{tang2022revisiting,
  title={Revisiting color-event based tracking: A unified network, dataset, and metric},
  author={Tang, Chuanming and Wang, Xiao and Huang, Ju and Jiang, Bo and Zhu, Lin and Zhang, Jianlin and Wang, Yaowei and Tian, Yonghong},
  journal={arXiv preprint arXiv:2211.11010},
  year={2022}
}

@article{wang2023visevent,
  title={Visevent: Reliable object tracking via collaboration of frame and event flows},
  author={Wang, Xiao and Li, Jianing and Zhu, Lin and Zhang, Zhipeng and Chen, Zhe and Li, Xin and Wang, Yaowei and Tian, Yonghong and Wu, Feng},
  journal={IEEE Transactions on Cybernetics},
  year={2023},
  publisher={IEEE}
}

@inproceedings{li2020challenge,
  title={Challenge-aware RGBT tracking},
  author={Li, Chenglong and Liu, Lei and Lu, Andong and Ji, Qing and Tang, Jin},
  booktitle={European conference on computer vision},
  pages={222--237},
  year={2020},
  organization={Springer}
}

@inproceedings{yan2021depthtrack,
  title={Depthtrack: Unveiling the power of rgbd tracking},
  author={Yan, Song and Yang, Jinyu and K{\"a}pyl{\"a}, Jani and Zheng, Feng and Leonardis, Ale{\v{s}} and K{\"a}m{\"a}r{\"a}inen, Joni-Kristian},
  booktitle={Proceedings of the IEEE/CVF International Conference on Computer Vision},
  pages={10725--10733},
  year={2021}
}

@inproceedings{zhu2023cross,
  title={Cross-modal orthogonal high-rank augmentation for rgb-event transformer-trackers},
  author={Zhu, Zhiyu and Hou, Junhui and Wu, Dapeng Oliver},
  booktitle={Proceedings of the IEEE/CVF International Conference on Computer Vision},
  pages={22045--22055},
  year={2023}
}

@inproceedings{jia2022visual,
  title={Visual prompt tuning},
  author={Jia, Menglin and Tang, Luming and Chen, Bor-Chun and Cardie, Claire and Belongie, Serge and Hariharan, Bharath and Lim, Ser-Nam},
  booktitle={European Conference on Computer Vision},
  pages={709--727},
  year={2022},
  organization={Springer}
}

@article{chen2022adaptformer,
  title={Adaptformer: Adapting vision transformers for scalable visual recognition},
  author={Chen, Shoufa and Ge, Chongjian and Tong, Zhan and Wang, Jiangliu and Song, Yibing and Wang, Jue and Luo, Ping},
  journal={Advances in Neural Information Processing Systems},
  volume={35},
  pages={16664--16678},
  year={2022}
}

@article{hu2021lora,
  title={Lora: Low-rank adaptation of large language models},
  author={Hu, Edward J and Shen, Yelong and Wallis, Phillip and Allen-Zhu, Zeyuan and Li, Yuanzhi and Wang, Shean and Wang, Lu and Chen, Weizhu},
  journal={arXiv preprint arXiv:2106.09685},
  year={2021}
}

@inproceedings{yang2022prompting,
  title={Prompting for multi-modal tracking},
  author={Yang, Jinyu and Li, Zhe and Zheng, Feng and Leonardis, Ales and Song, Jingkuan},
  booktitle={Proceedings of the 30th ACM international conference on multimedia},
  pages={3492--3500},
  year={2022}
}

@inproceedings{zhu2023visual,
  title={Visual prompt multi-modal tracking},
  author={Zhu, Jiawen and Lai, Simiao and Chen, Xin and Wang, Dong and Lu, Huchuan},
  booktitle={Proceedings of the IEEE/CVF conference on computer vision and pattern recognition},
  pages={9516--9526},
  year={2023}
}

@inproceedings{hou2024sdstrack,
  title={Sdstrack: Self-distillation symmetric adapter learning for multi-modal visual object tracking},
  author={Hou, Xiaojun and Xing, Jiazheng and Qian, Yijie and Guo, Yaowei and Xin, Shuo and Chen, Junhao and Tang, Kai and Wang, Mengmeng and Jiang, Zhengkai and Liu, Liang and others},
  booktitle={Proceedings of the IEEE/CVF Conference on Computer Vision and Pattern Recognition},
  pages={26551--26561},
  year={2024}
}

@inproceedings{zhang2021object,
  title={Object tracking by jointly exploiting frame and event domain},
  author={Zhang, Jiqing and Yang, Xin and Fu, Yingkai and Wei, Xiaopeng and Yin, Baocai and Dong, Bo},
  booktitle={Proceedings of the IEEE/CVF International Conference on Computer Vision},
  pages={13043--13052},
  year={2021}
}

@inproceedings{zhang2023frame,
  title={Frame-event alignment and fusion network for high frame rate tracking},
  author={Zhang, Jiqing and Wang, Yuanchen and Liu, Wenxi and Li, Meng and Bai, Jinpeng and Yin, Baocai and Yang, Xin},
  booktitle={Proceedings of the IEEE/CVF Conference on Computer Vision and Pattern Recognition},
  pages={9781--9790},
  year={2023}
}

@article{zhang2021jointly,
  title={Jointly modeling motion and appearance cues for robust RGB-T tracking},
  author={Zhang, Pengyu and Zhao, Jie and Bo, Chunjuan and Wang, Dong and Lu, Huchuan and Yang, Xiaoyun},
  journal={IEEE Transactions on Image Processing},
  volume={30},
  pages={3335--3347},
  year={2021},
  publisher={IEEE}
}

@article{zhang2024universal,
  title={A Universal Event-Based Plug-In Module for Visual Object Tracking in Degraded Conditions},
  author={Zhang, Jiqing and Dong, Bo and Fu, Yingkai and Wang, Yuanchen and Wei, Xiaopeng and Yin, Baocai and Yang, Xin},
  journal={International Journal of Computer Vision},
  volume={132},
  number={5},
  pages={1857--1879},
  year={2024},
  publisher={Springer}
}

@inproceedings{bhat2019learning,
  title={Learning discriminative model prediction for tracking},
  author={Bhat, Goutam and Danelljan, Martin and Gool, Luc Van and Timofte, Radu},
  booktitle={Proceedings of the IEEE/CVF international conference on computer vision},
  pages={6182--6191},
  year={2019}
}

@article{dosovitskiy2020image,
  title={An image is worth 16x16 words: Transformers for image recognition at scale},
  author={Dosovitskiy, Alexey and Beyer, Lucas and Kolesnikov, Alexander and Weissenborn, Dirk and Zhai, Xiaohua and Unterthiner, Thomas and Dehghani, Mostafa and Minderer, Matthias and Heigold, Georg and Gelly, Sylvain and others},
  journal={arXiv preprint arXiv:2010.11929},
  year={2020}
}

@inproceedings{zhu2023rgbd1k,
  title={RGBD1K: A large-scale dataset and benchmark for RGB-D object tracking},
  author={Zhu, Xue-Feng and Xu, Tianyang and Tang, Zhangyong and Wu, Zucheng and Liu, Haodong and Yang, Xiao and Wu, Xiao-Jun and Kittler, Josef},
  booktitle={Proceedings of the AAAI Conference on Artificial Intelligence},
  volume={37},
  pages={3870--3878},
  year={2023}
}

@inproceedings{xiao2022attribute,
  title={Attribute-based progressive fusion network for rgbt tracking},
  author={Xiao, Yun and Yang, Mengmeng and Li, Chenglong and Liu, Lei and Tang, Jin},
  booktitle={Proceedings of the AAAI Conference on Artificial Intelligence},
  volume={36},
  pages={2831--2838},
  year={2022}
}

@inproceedings{chen2023seqtrack,
  title={Seqtrack: Sequence to sequence learning for visual object tracking},
  author={Chen, Xin and Peng, Houwen and Wang, Dong and Lu, Huchuan and Hu, Han},
  booktitle={Proceedings of the IEEE/CVF conference on computer vision and pattern recognition},
  pages={14572--14581},
  year={2023}
}

@article{lin2022swintrack,
  title={Swintrack: A simple and strong baseline for transformer tracking},
  author={Lin, Liting and Fan, Heng and Zhang, Zhipeng and Xu, Yong and Ling, Haibin},
  journal={Advances in Neural Information Processing Systems},
  volume={35},
  pages={16743--16754},
  year={2022}
}

@inproceedings{cai2024hiptrack,
  title={HIPTrack: Visual Tracking with Historical Prompts},
  author={Cai, Wenrui and Liu, Qingjie and Wang, Yunhong},
  booktitle={Proceedings of the IEEE/CVF Conference on Computer Vision and Pattern Recognition},
  pages={19258--19267},
  year={2024}
}

@inproceedings{zheng2024nettrack,
  title={NetTrack: Tracking Highly Dynamic Objects with a Net},
  author={Zheng, Guangze and Lin, Shijie and Zuo, Haobo and Fu, Changhong and Pan, Jia},
  booktitle={Proceedings of the IEEE/CVF Conference on Computer Vision and Pattern Recognition},
  pages={19145--19155},
  year={2024}
}

@inproceedings{xie2024diffusiontrack,
  title={DiffusionTrack: Point Set Diffusion Model for Visual Object Tracking},
  author={Xie, Fei and Wang, Zhongdao and Ma, Chao},
  booktitle={Proceedings of the IEEE/CVF Conference on Computer Vision and Pattern Recognition},
  pages={19113--19124},
  year={2024}
}

@inproceedings{hong2024onetracker,
  title={Onetracker: Unifying visual object tracking with foundation models and efficient tuning},
  author={Hong, Lingyi and Yan, Shilin and Zhang, Renrui and Li, Wanyun and Zhou, Xinyu and Guo, Pinxue and Jiang, Kaixun and Chen, Yiting and Li, Jinglun and Chen, Zhaoyu and others},
  booktitle={Proceedings of the IEEE/CVF Conference on Computer Vision and Pattern Recognition},
  pages={19079--19091},
  year={2024}
}

@inproceedings{qian2021dal,
  title={DAL: A deep depth-aware long-term tracker},
  author={Qian, Yanlin and Yan, Song and Luke{\v{z}}i{\v{c}}, Alan and Kristan, Matej and K{\"a}m{\"a}r{\"a}inen, Joni-Kristian and Matas, Ji{\v{r}}{\'\i}},
  booktitle={2020 25th International conference on pattern recognition (ICPR)},
  pages={7825--7832},
  year={2021},
  organization={IEEE}
}

@inproceedings{cao2024bi,
  title={Bi-directional adapter for multimodal tracking},
  author={Cao, Bing and Guo, Junliang and Zhu, Pengfei and Hu, Qinghua},
  booktitle={Proceedings of the AAAI Conference on Artificial Intelligence},
  volume={38},
  pages={927--935},
  year={2024}
}

@inproceedings{hui2023bridging,
  title={Bridging search region interaction with template for rgb-t tracking},
  author={Hui, Tianrui and Xun, Zizheng and Peng, Fengguang and Huang, Junshi and Wei, Xiaoming and Wei, Xiaolin and Dai, Jiao and Han, Jizhong and Liu, Si},
  booktitle={Proceedings of the IEEE/CVF Conference on Computer Vision and Pattern Recognition},
  pages={13630--13639},
  year={2023}
}

@inproceedings{wu2024single,
  title={Single-model and any-modality for video object tracking},
  author={Wu, Zongwei and Zheng, Jilai and Ren, Xiangxuan and Vasluianu, Florin-Alexandru and Ma, Chao and Paudel, Danda Pani and Van Gool, Luc and Timofte, Radu},
  booktitle={Proceedings of the IEEE/CVF Conference on Computer Vision and Pattern Recognition},
  pages={19156--19166},
  year={2024}
}

@article{chen2024crossei,
  title={Crossei: Boosting motion-oriented object tracking with an event camera},
  author={Chen, Zhiwen and Wu, Jinjian and Dong, Weisheng and Li, Leida and Shi, Guangming},
  journal={IEEE Transactions on Image Processing},
  year={2024},
  publisher={IEEE}
}

@inproceedings{kristan2022tenth,
  title={The tenth visual object tracking vot2022 challenge results},
  author={Kristan, Matej and Leonardis, Ale{\v{s}} and Matas, Ji{\v{r}}{\'\i} and Felsberg, Michael and Pflugfelder, Roman and K{\"a}m{\"a}r{\"a}inen, Joni-Kristian and Chang, Hyung Jin and Danelljan, Martin and Zajc, Luka {\v{C}}ehovin and Luke{\v{z}}i{\v{c}}, Alan and others},
  booktitle={European Conference on Computer Vision},
  pages={431--460},
  year={2022},
  organization={Springer}
}

@article{li2019rgb,
  title={RGB-T object tracking: Benchmark and baseline},
  author={Li, Chenglong and Liang, Xinyan and Lu, Yijuan and Zhao, Nan and Tang, Jin},
  journal={Pattern Recognition},
  volume={96},
  pages={106977},
  year={2019},
  publisher={Elsevier}
}

@inproceedings{chen2025sutrack,
  title={Sutrack: Towards simple and unified single object tracking},
  author={Chen, Xin and Kang, Ben and Geng, Wanting and Zhu, Jiawen and Liu, Yi and Wang, Dong and Lu, Huchuan},
  booktitle={Proceedings of the AAAI Conference on Artificial Intelligence},
  volume={39},
  pages={2239--2247},
  year={2025}
}

@InProceedings{Sun_2025_CVPR,
    author    = {Sun, Chuanyu and Zhang, Jiqing and Wang, Yang and Ge, Huilin and Xia, Qianchen and Yin, Baocai and Yang, Xin},
    title     = {Exploring Historical Information for RGBE Visual Tracking with Mamba},
    booktitle = {Proceedings of the Computer Vision and Pattern Recognition Conference (CVPR)},
    month     = {June},
    year      = {2025},
    pages     = {6500-6509}
}

@InProceedings{Wang_2024_CVPR,
    author    = {Wang, Xiao and Wang, Shiao and Tang, Chuanming and Zhu, Lin and Jiang, Bo and Tian, Yonghong and Tang, Jin},
    title     = {Event Stream-based Visual Object Tracking: A High-Resolution Benchmark Dataset and A Novel Baseline},
    booktitle = {Proceedings of the IEEE/CVF Conference on Computer Vision and Pattern Recognition (CVPR)},
    month     = {June},
    year      = {2024},
    pages     = {19248-19257}
}

@misc{GDSTrack_2025_IJCAI,
      title={Modality-Guided Dynamic Graph Fusion and Temporal Diffusion for Self-Supervised RGB-T Tracking}, 
      author={Shenglan Li and Rui Yao and Yong Zhou and Hancheng Zhu and Kunyang Sun and Bing Liu and Zhiwen Shao and Jiaqi Zhao},
      year={2025},
      eprint={2505.03507},
      archivePrefix={arXiv},
      primaryClass={cs.CV},
      url={https://arxiv.org/abs/2505.03507}, 
}

@ARTICLE{10819455,
  author={Liu, Zelin and Wang, Xinggang and Wang, Cheng and Liu, Wenyu and Bai, Xiang},
  journal={IEEE Transactions on Circuits and Systems for Video Technology}, 
  title={SparseTrack: Multi-Object Tracking by Performing Scene Decomposition Based on Pseudo-Depth}, 
  year={2025},
  volume={35},
  number={5},
  pages={4870-4882},}

@InProceedings{Xiang_2025_CVPR,
    author    = {Xiang, Xinyu and Yan, Qinglong and Zhang, Hao and Ma, Jiayi},
    title     = {ACAttack: Adaptive Cross Attacking RGB-T Tracker via Multi-Modal Response Decoupling},
    booktitle = {Proceedings of the Computer Vision and Pattern Recognition Conference (CVPR)},
    month     = {June},
    year      = {2025},
    pages     = {22099-22108}
}

@inproceedings{tan2024xtrack,
  title={XTrack: Multimodal Training Boosts RGB-X Video Object Trackers},
  author={Tan, Yuedong and Wu, Zongwei and Fu, Yuqian and Zhou, Zhuyun and Sun, Guolei and Ma, Chao and Paudel, Danda Pani and Van Gool, Luc and Timofte, Radu},
  booktitle = {Proceedings of the IEEE/CVF International Conference on Computer Vision (ICCV)},
  year={2025}
}

@inproceedings{fengcstrack,
  title={CSTrack: Enhancing RGB-X Tracking via Compact Spatiotemporal Features},
  author={Feng, Xiaokun and Zhang, Dailing and Hu, Shiyu and Li, Xuchen and Wu, Meiqi and Zhang, Jing and Chen, Xiaotang and Huang, Kaiqi},
  booktitle={Forty-second International Conference on Machine Learning},
  year={2025}
}

@article{tan2025you,
  title={What You Have is What You Track: Adaptive and Robust Multimodal Tracking},
  author={Tan, Yuedong and Shao, Jiawei and Zamfir, Eduard and Li, Ruanjun and An, Zhaochong and Ma, Chao and Paudel, Danda and Van Gool, Luc and Timofte, Radu and Wu, Zongwei},
  journal={arXiv preprint arXiv:2507.05899},
  year={2025}
}

@inproceedings{bai2024artrackv2,
  title={Artrackv2: Prompting autoregressive tracker where to look and how to describe},
  author={Bai, Yifan and Zhao, Zeyang and Gong, Yihong and Wei, Xing},
  booktitle={Proceedings of the IEEE/CVF conference on computer vision and pattern recognition},
  pages={19048--19057},
  year={2024}
}

@article{zhang2024revisiting,
  title={Revisiting motion information for RGB-event tracking with MOT philosophy},
  author={Zhang, Tianlu and Debattista, Kurt and Zhang, Qiang and Han, Jungong and others},
  journal={Advances in Neural Information Processing Systems},
  volume={37},
  pages={89346--89372},
  year={2024}
}

@inproceedings{wei2023autoregressive,
  title={Autoregressive visual tracking},
  author={Wei, Xing and Bai, Yifan and Zheng, Yongchao and Shi, Dahu and Gong, Yihong},
  booktitle={Proceedings of the IEEE/CVF conference on computer vision and pattern recognition},
  pages={9697--9706},
  year={2023}
}

@inproceedings{lukezic2019cdtb,
  title={Cdtb: A color and depth visual object tracking dataset and benchmark},
  author={Lukezic, Alan and Kart, Ugur and Kapyla, Jani and Durmush, Ahmed and Kamarainen, Joni-Kristian and Matas, Jiri and Kristan, Matej},
  booktitle={Proceedings of the IEEE/CVF International Conference on Computer Vision},
  pages={10013--10022},
  year={2019}
}

@article{foret2020sharpness,
  title={Sharpness-aware minimization for efficiently improving generalization},
  author={Foret, Pierre and Kleiner, Ariel and Mobahi, Hossein and Neyshabur, Behnam},
  journal={arXiv preprint arXiv:2010.01412},
  year={2020}
}

@inproceedings{amari2019fisher,
  title={Fisher information and natural gradient learning in random deep networks},
  author={Amari, Shun-ichi and Karakida, Ryo and Oizumi, Masafumi},
  booktitle={The 22nd International Conference on Artificial Intelligence and Statistics},
  pages={694--702},
  year={2019},
  organization={PMLR}
}

@book{nesterov2013introductory,
  title={Introductory lectures on convex optimization: A basic course},
  author={Nesterov, Yurii},
  volume={87},
  year={2013},
  publisher={Springer Science \& Business Media}
}

@article{leveque1998finite,
  title={Finite difference methods for differential equations},
  author={LeVeque, Randall J},
  journal={Draft version for use in AMath},
  volume={585},
  number={6},
  pages={112},
  year={1998}
}

@inproceedings{lin2024tracking,
  title={Tracking meets lora: Faster training, larger model, stronger performance},
  author={Lin, Liting and Fan, Heng and Zhang, Zhipeng and Wang, Yaowei and Xu, Yong and Ling, Haibin},
  booktitle={European Conference on Computer Vision},
  pages={300--318},
  year={2024},
  organization={Springer}
}

@inproceedings{zheng2024odtrack,
  title={Odtrack: Online dense temporal token learning for visual tracking},
  author={Zheng, Yaozong and Zhong, Bineng and Liang, Qihua and Mo, Zhiyi and Zhang, Shengping and Li, Xianxian},
  booktitle={Proceedings of the AAAI conference on artificial intelligence},
  volume={38},
  pages={7588--7596},
  year={2024}
}

@article{hurley2009comparing,
  title={Comparing measures of sparsity},
  author={Hurley, Niall and Rickard, Scott},
  journal={IEEE Transactions on Information Theory},
  volume={55},
  number={10},
  pages={4723--4741},
  year={2009},
  publisher={IEEE}
}

@inproceedings{zhang2023hivit,
  title={Hivit: A simpler and more efficient design of hierarchical vision transformer},
  author={Zhang, Xiaosong and Tian, Yunjie and Xie, Lingxi and Huang, Wei and Dai, Qi and Ye, Qixiang and Tian, Qi},
  booktitle={The eleventh international conference on learning representations},
  year={2023}
}

@misc{chen2024unifiedsequencetosequencelearningsingle,
      title={Unified Sequence-to-Sequence Learning for Single- and Multi-Modal Visual Object Tracking}, 
      author={Xin Chen and Ben Kang and Jiawen Zhu and Dong Wang and Houwen Peng and Huchuan Lu},
      year={2024},
      eprint={2304.14394},
      archivePrefix={arXiv},
      primaryClass={cs.CV},
      url={https://arxiv.org/abs/2304.14394}, 
}

@inproceedings{rebecq2017real,
  author={Rebecq, Henri and Horstschaefer, Timo and Scaramuzza, Davide},
  title        = {Real-time Visual-Inertial Odometry for Event Cameras using Keyframe-based
                  Nonlinear Optimization},
  booktitle    = {British Machine Vision Conference},
  year         = {2017},
}

@inproceedings{maqueda2018event,
  title={Event-based vision meets deep learning on steering prediction for self-driving cars},
  author={Maqueda, Ana I and Loquercio, Antonio and Gallego, Guillermo and Garc{\'\i}a, Narciso and Scaramuzza, Davide},
  booktitle={Proceedings of the IEEE conference on computer vision and pattern recognition},
  pages={5419--5427},
  year={2018}
}

@inproceedings{sironi2018hats,
  title={HATS: Histograms of averaged time surfaces for robust event-based object classification},
  author={Sironi, Amos and Brambilla, Manuele and Bourdis, Nicolas and Lagorce, Xavier and Benosman, Ryad},
  booktitle={Proceedings of the IEEE conference on computer vision and pattern recognition},
  pages={1731--1740},
  year={2018}
}

@inproceedings{chen2020end,
  title={End-to-end learning of object motion estimation from retinal events for event-based object tracking},
  author={Chen, Haosheng and Suter, David and Wu, Qiangqiang and Wang, Hanzi},
  booktitle={Proceedings of the AAAI Conference on Artificial Intelligence},
  volume={34},
  pages={10534--10541},
  year={2020}
}

@inproceedings{zhu2019unsupervised,
  title={Unsupervised event-based learning of optical flow, depth, and egomotion},
  author={Zhu, Alex Zihao and Yuan, Liangzhe and Chaney, Kenneth and Daniilidis, Kostas},
  booktitle={Proceedings of the IEEE/CVF Conference on Computer Vision and Pattern Recognition},
  pages={989--997},
  year={2019}
}

@article{zhang2023glenet,
  title={Glenet: Boosting 3d object detectors with generative label uncertainty estimation},
  author={Zhang, Yifan and Zhang, Qijian and Zhu, Zhiyu and Hou, Junhui and Yuan, Yixuan},
  journal={International Journal of Computer Vision},
  volume={131},
  number={12},
  pages={3332--3352},
  year={2023},
  publisher={Springer}
}

@article{zhu2022learning,
  title={Learning graph-embedded key-event back-tracing for object tracking in event clouds},
  author={Zhu, Zhiyu and Hou, Junhui and Lyu, Xianqiang},
  journal={Advances in Neural Information Processing Systems},
  volume={35},
  pages={7462--7476},
  year={2022}
}

@article{li2015rayleigh,
  title={Rayleigh quotient based optimization methods for eigenvalue problems},
  author={Li, Ren-Cang},
  journal={Matrix Functions and Matrix Equations},
  volume={19},
  pages={76--108},
  year={2015},
  publisher={World Scientific}
}

@article{wang2025mamba,
  title={Mamba-FETrack V2: Revisiting State Space Model for Frame-Event based Visual Object Tracking},
  author={Wang, Shiao and Huang, Ju and Ma, Qingchuan and Gao, Jinfeng and Xu, Chunyi and Wang, Xiao and Chen, Lan and Jiang, Bo},
  journal={arXiv preprint arXiv:2506.23783},
  year={2025}
}

@article{wang2025towards,
  title={Towards Low-Latency Event Stream-based Visual Object Tracking: A Slow-Fast Approach},
  author={Wang, Shiao and Wang, Xiao and Jin, Liye and Jiang, Bo and Zhu, Lin and Chen, Lan and Tian, Yonghong and Luo, Bin},
  journal={arXiv preprint arXiv:2505.12903},
  year={2025}
}

@inproceedings{ghorbani2019investigation,
  title={An investigation into neural net optimization via hessian eigenvalue density},
  author={Ghorbani, Behrooz and Krishnan, Shankar and Xiao, Ying},
  booktitle={International Conference on Machine Learning},
  pages={2232--2241},
  year={2019},
  organization={PMLR}
}

@inproceedings{rame2022fishr,
  title={Fishr: Invariant gradient variances for out-of-distribution generalization},
  author={Rame, Alexandre and Dancette, Corentin and Cord, Matthieu},
  booktitle={International Conference on Machine Learning},
  pages={18347--18377},
  year={2022},
  organization={PMLR}
}

@inproceedings{he2023sensitivity,
  title={Sensitivity-aware visual parameter-efficient fine-tuning},
  author={He, Haoyu and Cai, Jianfei and Zhang, Jing and Tao, Dacheng and Zhuang, Bohan},
  booktitle={Proceedings of the IEEE/CVF International Conference on Computer Vision},
  pages={11825--11835},
  year={2023}
}

@article{wang2024model,
  title={Model sensitivity aware continual learning},
  author={Wang, Zhenyi and Huang, Heng},
  journal={Advances in Neural Information Processing Systems},
  volume={37},
  pages={132583--132613},
  year={2024}
}

@inproceedings{chen2025sensitivity,
  title={Sensitivity-Aware Efficient Fine-Tuning via Compact Dynamic-Rank Adaptation},
  author={Chen, Tianran and Chen, Jiarui and Zhang, Baoquan and Yu, Zhehao and Chen, Shidong and Ye, Rui and Li, Xutao and Ye, Yunming},
  booktitle={Proceedings of the Computer Vision and Pattern Recognition Conference},
  pages={9655--9664},
  year={2025}
}

@article{fefferman2016testing,
  title={Testing the manifold hypothesis},
  author={Fefferman, Charles and Mitter, Sanjoy and Narayanan, Hariharan},
  journal={Journal of the American Mathematical Society},
  volume={29},
  number={4},
  pages={983--1049},
  year={2016}
}

@article{meilua2024manifold,
  title={Manifold learning: What, how, and why},
  author={Meil{\u{a}}, Marina and Zhang, Hanyu},
  journal={Annual Review of Statistics and Its Application},
  volume={11},
  number={1},
  pages={393--417},
  year={2024},
  publisher={Annual Reviews}
}

@article{zheng2025towards,
  title={Towards lifelong learning of large language models: A survey},
  author={Zheng, Junhao and Qiu, Shengjie and Shi, Chengming and Ma, Qianli},
  journal={ACM Computing Surveys},
  volume={57},
  number={8},
  pages={1--35},
  year={2025},
  publisher={ACM New York, NY}
}

@article{zhou2025revisiting,
  title={Revisiting class-incremental learning with pre-trained models: Generalizability and adaptivity are all you need},
  author={Zhou, Da-Wei and Cai, Zi-Wen and Ye, Han-Jia and Zhan, De-Chuan and Liu, Ziwei},
  journal={International Journal of Computer Vision},
  volume={133},
  number={3},
  pages={1012--1032},
  year={2025},
  publisher={Springer}
}

@article{bengio2013representation,
  title={Representation learning: A review and new perspectives},
  author={Bengio, Yoshua and Courville, Aaron and Vincent, Pascal},
  journal={IEEE transactions on pattern analysis and machine intelligence},
  volume={35},
  number={8},
  pages={1798--1828},
  year={2013},
  publisher={IEEE}
}

@article{kunstner2019limitations,
  title={Limitations of the empirical fisher approximation for natural gradient descent},
  author={Kunstner, Frederik and Hennig, Philipp and Balles, Lukas},
  journal={Advances in neural information processing systems},
  volume={32},
  year={2019}
}

@inproceedings{fan2019lasot,
  title={Lasot: A high-quality benchmark for large-scale single object tracking},
  author={Fan, Heng and Lin, Liting and Yang, Fan and Chu, Peng and Deng, Ge and Yu, Sijia and Bai, Hexin and Xu, Yong and Liao, Chunyuan and Ling, Haibin},
  booktitle={2019 IEEE/CVF Conference on Computer Vision and Pattern Recognition (CVPR)},
  pages={5369--5378},
  year={2019},
  organization={IEEE}
}

@inproceedings{lin2014microsoft,
  title={Microsoft coco: Common objects in context},
  author={Lin, Tsung-Yi and Maire, Michael and Belongie, Serge and Hays, James and Perona, Pietro and Ramanan, Deva and Doll{\'a}r, Piotr and Zitnick, C Lawrence},
  booktitle={European conference on computer vision},
  pages={740--755},
  year={2014},
  organization={Springer}
}

@inproceedings{muller2018trackingnet,
  title={Trackingnet: A large-scale dataset and benchmark for object tracking in the wild},
  author={Muller, Matthias and Bibi, Adel and Giancola, Silvio and Alsubaihi, Salman and Ghanem, Bernard},
  booktitle={Proceedings of the European conference on computer vision (ECCV)},
  pages={300--317},
  year={2018}
}

@article{huang2019got,
  title={Got-10k: A large high-diversity benchmark for generic object tracking in the wild},
  author={Huang, Lianghua and Zhao, Xin and Huang, Kaiqi},
  journal={IEEE transactions on pattern analysis and machine intelligence},
  volume={43},
  number={5},
  pages={1562--1577},
  year={2019},
  publisher={IEEE}
}

@inproceedings{xuhong2018explicit,
  title={Explicit inductive bias for transfer learning with convolutional networks},
  author={Xuhong, LI and Grandvalet, Yves and Davoine, Franck},
  booktitle={International conference on machine learning},
  pages={2825--2834},
  year={2018},
  organization={PMLR}
}

@article{kirkpatrick2017overcoming,
  title={Overcoming catastrophic forgetting in neural networks},
  author={Kirkpatrick, James and Pascanu, Razvan and Rabinowitz, Neil and Veness, Joel and Desjardins, Guillaume and Rusu, Andrei A and Milan, Kieran and Quan, John and Ramalho, Tiago and Grabska-Barwinska, Agnieszka and others},
  journal={Proceedings of the national academy of sciences},
  volume={114},
  number={13},
  pages={3521--3526},
  year={2017},
  publisher={National Academy of Sciences}
}

@inproceedings{chaudhry2018riemannian,
  title={Riemannian walk for incremental learning: Understanding forgetting and intransigence},
  author={Chaudhry, Arslan and Dokania, Puneet K and Ajanthan, Thalaiyasingam and Torr, Philip HS},
  booktitle={European conference on computer vision},
  pages={556--572},
  year={2018},
  organization={Springer}
}

@inproceedings{martens2015optimizing,
  title={Optimizing neural networks with kronecker-factored approximate curvature},
  author={Martens, James and Grosse, Roger},
  booktitle={International conference on machine learning},
  pages={2408--2417},
  year={2015},
  organization={PMLR}
}


\clearpage

\appendix
\renewcommand{\thesection}{\Alph{section}}

\section*{Appendix} 
\revise{This appendix contains the following contents. In Section A, we provide
implementation details for multi-modality tracking, including a more
detailed description of the network architecture. In Section B, we present
the complete derivations, proofs, and optimization interpretation of SRFT.
In Section C, we provide additional comparisons and analyses, including
classical regularization and optimization baselines, computational and
training efficiency, reduced curvature probing and large-backbone costs,
robustness to incomplete source data, alternative prior-significance
estimators, functional sensitivity, update-stabilization strategies,
transfer-significance robustness, and significance fusion schedules.
Finally, Section D provides qualitative tracking visualizations.}
\begin{itemize}
    \item Section A: Implementation details.
    \item Section B: Theoretical derivations and proofs.
    \item Section C: Additional experiments and analyses.
    \item Section D: Visualization of tracking results.
\end{itemize}

\section{Implementation details} \label{sec:a-1}
\begin{figure}[htbp]
\begin{center}
\centerline{\includegraphics[width=0.35\textwidth]{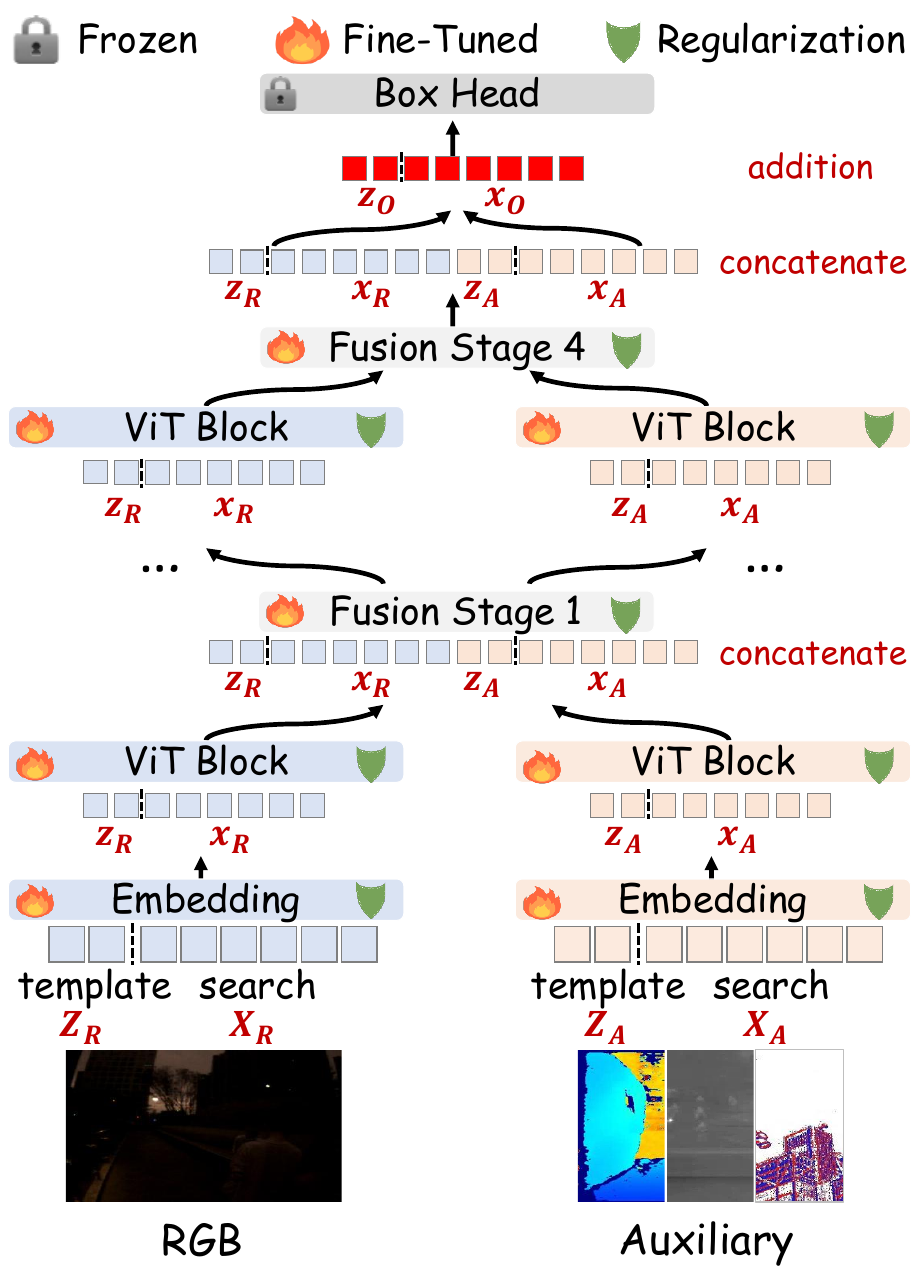}}
\vskip -0.15in
\caption{More detailed network architecture of our multi-modality trackers.}
\label{fig:appendix-architecture}
\end{center}
\vskip -0.4in
\end{figure}

The detailed structure of the multi-modality tracker is shown in Fig.~\ref{fig:appendix-architecture}. The input of our proposed tracking model consists of a pair of template frames and a pair of search frames, i.e., one RGB template frame $\mathrm{Z}_{\mathrm{R}} \in \mathbb{R}^{H_z \times W_z \times 3}$, one RGB search frame $\mathrm{X}_{\mathrm{R}} \in \mathbb{R}^{H_x \times W_x \times 3}$, one auxiliary-modal template frame $\mathrm{Z}_{\mathrm{A}} \in \mathbb{R}^{H_z \times W_z \times 3}$, and one auxiliary-modality search frame $\mathrm{X}_{\mathrm{A}} \in \mathbb{R}^{H_x \times W_x \times 3}$. Notably, to make event data compatible with the RGB domain, we aggregate the event set between the image and its next one into a three-channel color-polar event image. These data are first split and flattened into sequences of patches $\mathrm{z}_{\mathrm{R}}, \mathrm{z}_{\mathrm{A}} \in \mathbb{R}^{N_z \times \left(3 P^2\right)}$ and $\mathrm{x}_{\mathrm{R}}, \mathrm{x}_{\mathrm{A}} \in \mathbb{R}^{N_x \times \left(3 P^2\right)}$, where $P \times P$ is the resolution of each patch ($P=16$), and $N_z=\frac{H_z W_z}{P^2}$, $N_x=\frac{H_x W_x}{P^2}$. Next, two modal-aware patch embedding layers are used to project $\mathrm{z}_{\mathrm{R}}, \mathrm{x}_{\mathrm{R}}$ and $\mathrm{z}_{\mathrm{A}}, \mathrm{x}_{\mathrm{A}}$ into the D-dimensional latent space, $\mathrm{z}_{\mathrm{R}}, \mathrm{z}_{\mathrm{A}} \in \mathbb{R}^{N_z \times D}$ and $\mathrm{x}_{\mathrm{R}}, \mathrm{x}_{\mathrm{A}} \in \mathbb{R}^{N_x \times D}$. The patch embeddings $\mathrm{z}_{\mathrm{R}}$ and $\mathrm{x}_{\mathrm{R}}$ are concatenated as $\mathbf{H}_{\mathrm{R}}^{(0)} = \left[\mathrm{z}_{\mathrm{R}} ; \mathrm{x}_{\mathrm{R}}\right] \in \mathbb{R}^{\left(N_z+N_x\right) \times D}$, and $\mathrm{z}_{\mathrm{A}}$ and $\mathrm{x}_{\mathrm{A}}$ are concatenated as $\mathbf{H}_{\mathrm{A}}^{(0)} = \left[\mathrm{z}_{\mathrm{A}} ; \mathrm{x}_{\mathrm{A}}\right] \in \mathbb{R}^{\left(N_z+N_x\right) \times D}$. 
The computation of modality-aware ViT block can be formulated as:
$$
\begin{gathered}
\mathbf{H}_{\mathrm{X}}^{\prime(l)} = \mathbf{H}_{\mathrm{X}}^{(l-1)} + \operatorname{MSA}\left(\mathrm{LN}\left(\mathbf{H}_{\mathrm{X}}^{(l-1)}\right)\right) \\
\mathbf{H}_{\mathrm{X}}^{(l)} = \mathbf{H}_{\mathrm{X}}^{\prime(l)} + \operatorname{MLP}\left(\mathrm{LN}\left(\mathbf{H}_{\mathrm{X}}^{\prime(l)}\right)\right)
\end{gathered}
$$
where $X \in \{R,A\}$, $\mathbf{X}_{\mathrm{X}}^{(l-1)}$ and $\mathbf{H}_{\mathrm{X}}^{(l)}$ represent the outputs of the $(l-1)$-th and $l$-th ViT blocks, respectively. For the cross-modal block, we concatenate $\mathbf{H}_{\mathrm{F}} = \left[\mathrm{z}_{\mathrm{R}};\mathrm{x}_{\mathrm{R}};\mathrm{z}_{\mathrm{A}} ; \mathrm{x}_{\mathrm{A}}\right] \in \mathbb{R}^{\left(N_z+N_x+N_z+N_x\right) \times D}$ as input, and use the same attention block as above for cross-modal feature interaction. After the multi-modality interaction/fusion at one stage, the RGB and auxiliary features/tokens are separated and processed independently using modality-specific ViT blocks, preparing them for the next multi-modality interaction. After the final multi-modality interaction stage, we combine the modality-specific search and template tokens to produce the fused tokens that are input to the box head: $\mathrm{z}_{\mathrm{O}} = (\mathrm{z}_{\mathrm{R}} + \mathrm{z}_{\mathrm{A}})/2$ and $\mathrm{x}_{\mathrm{O}} = (\mathrm{x}_{\mathrm{R}} + \mathrm{x}_{\mathrm{A}})/2$. In our setup, the specific ViT blocks used for multi-modality interaction in OSTrack and DropTrack are layers 2, 5, 8, and 11, while in SUTrack, they are layers 13, 19, 25, and 31. 

\section{Proof of Derivations} \label{sec:a-2}

\subsection{Proof of Proposition 1}
\label{sec:a-2-1}
We provide the proof of Proposition~1 by separately analyzing the
matrix discrepancy introduced by the group-isotropic scalarization and
its effect on the local quadratic source-loss response.

\noindent\textbf{1. Exact Scalarization Error.}
For the $j$-th operation, let the corresponding source-loss Hessian
block admit the eigendecomposition
\begin{equation}
\mathcal{H}^{(\theta_0^j)}
=
V^j \Lambda^j (V^j)^{T},
\end{equation}
where
\begin{equation}
\Lambda^j
=
\operatorname{diag}
\left(
\lambda_1^j,
\lambda_2^j,
\ldots,
\lambda_{|\theta^j|}^j
\right),
\qquad
\lambda_1^j
\ge
\lambda_2^j
\ge
\cdots
\ge
\lambda_{|\theta^j|}^j,
\end{equation}
and $V^j$ is an orthogonal matrix.

The scalar curvature statistic used for the group-isotropic
scalarization is defined as
\begin{equation}
\gamma^j
=
\frac{1}{K}
\sum_{i=1}^{K}
\lambda_i^j,
\end{equation}
and the corresponding scalarization is
\begin{equation}
\widetilde{\mathcal{H}}^{(\theta_0^j)}
=
\gamma^j I_{|\theta^j|}.
\end{equation}

Since the Frobenius norm is invariant under orthogonal
transformations, we have
\begin{align}
&
\left\|
\mathcal{H}^{(\theta_0^j)}
-
\widetilde{\mathcal{H}}^{(\theta_0^j)}
\right\|_F^2
\nonumber\\
&=
\left\|
V^j
\Lambda^j
(V^j)^T
-
\gamma^j I_{|\theta^j|}
\right\|_F^2
\nonumber\\
&=
\left\|
V^j
\left(
\Lambda^j-\gamma^j I_{|\theta^j|}
\right)
(V^j)^T
\right\|_F^2
\nonumber\\
&=
\left\|
\Lambda^j-\gamma^j I_{|\theta^j|}
\right\|_F^2
\nonumber\\
&=
\sum_{i=1}^{|\theta^j|}
\left(
\lambda_i^j-\gamma^j
\right)^2.
\label{eq:app_scalar_error_1}
\end{align}

Separating the top-$K$ eigenvalues from the remaining spectrum gives
\begin{align}
&
\left\|
\mathcal{H}^{(\theta_0^j)}
-
\widetilde{\mathcal{H}}^{(\theta_0^j)}
\right\|_F^2
\nonumber\\
&=
\sum_{i=1}^{K}
\left(
\lambda_i^j-\gamma^j
\right)^2
+
\sum_{i=K+1}^{|\theta^j|}
\left(
\lambda_i^j-\gamma^j
\right)^2.
\label{eq:app_scalar_error_2}
\end{align}

By the definition of $\gamma^j$,
\begin{equation}
\frac{1}{K}
\sum_{i=1}^{K}
\lambda_i^j
=
\gamma^j.
\end{equation}
We define
\begin{equation}
\operatorname{Var}(\lambda_{1:K}^j)
=
\frac{1}{K}
\sum_{i=1}^{K}
\left(
\lambda_i^j-\gamma^j
\right)^2.
\label{eq:app_topk_variance}
\end{equation}
Therefore,
\begin{equation}
\sum_{i=1}^{K}
\left(
\lambda_i^j-\gamma^j
\right)^2
=
K\operatorname{Var}(\lambda_{1:K}^j).
\end{equation}

Substituting this expression into
Eq.~(\ref{eq:app_scalar_error_2}) yields
\begin{align}
&
\left\|
\mathcal{H}^{(\theta_0^j)}
-
\widetilde{\mathcal{H}}^{(\theta_0^j)}
\right\|_F^2
\nonumber\\
&=
K\operatorname{Var}(\lambda_{1:K}^j)
+
\sum_{i=K+1}^{|\theta^j|}
\left(
\lambda_i^j-\gamma^j
\right)^2.
\label{eq:app_exact_scalar_error}
\end{align}

Hence, the scalarization error contains two distinct components.
The first term measures the dispersion among the top-$K$ eigenvalues
and therefore quantifies the information lost when they are compressed
into a single scalar. The second term measures the mismatch between
their mean and the remaining spectrum. In particular, the first term
vanishes only when the top-$K$ eigenvalues are identical. Therefore,
the scalarization error cannot in general be characterized solely by
the discarded spectral tail.

\medskip
\noindent\textbf{2. Bounded Source-Loss Response Error.}
We next analyze how the group-isotropic scalarization affects the
local quadratic source-loss response.

Let the operation-wise block-diagonal approximation of the source-loss
Hessian be
\begin{equation}
\mathcal{H}_{\mathrm{bd}}^{(\theta_0)}
=
\operatorname{blkdiag}
\left(
\mathcal{H}^{(\theta_0^1)},
\ldots,
\mathcal{H}^{(\theta_0^N)}
\right),
\end{equation}
and let its group-isotropic scalarization be
\begin{equation}
\widetilde{\mathcal{H}}^{(\theta_0)}
=
\operatorname{blkdiag}
\left(
\widetilde{\mathcal{H}}^{(\theta_0^1)},
\ldots,
\widetilde{\mathcal{H}}^{(\theta_0^N)}
\right).
\end{equation}

For a parameter displacement
$\Delta\theta\in\mathbb{R}^{|\theta|}$,
define the quadratic source-loss response associated with a symmetric
matrix $\mathcal{A}$ as
\begin{equation}
\varepsilon_{\mathrm{gen}}(\mathcal{A})
=
\frac{1}{2}
\Delta\theta^T
\mathcal{A}
\Delta\theta.
\label{eq:app_quadratic_response}
\end{equation}

The discrepancy between the responses induced by
$\mathcal{H}_{\mathrm{bd}}^{(\theta_0)}$ and
$\widetilde{\mathcal{H}}^{(\theta_0)}$ is
\begin{align}
&
\left|
\varepsilon_{\mathrm{gen}}
\left(
\mathcal{H}_{\mathrm{bd}}^{(\theta_0)}
\right)
-
\varepsilon_{\mathrm{gen}}
\left(
\widetilde{\mathcal{H}}^{(\theta_0)}
\right)
\right|
\nonumber\\
&=
\frac{1}{2}
\left|
\Delta\theta^T
\left(
\mathcal{H}_{\mathrm{bd}}^{(\theta_0)}
-
\widetilde{\mathcal{H}}^{(\theta_0)}
\right)
\Delta\theta
\right|.
\label{eq:app_response_error_1}
\end{align}

For any symmetric matrix $\mathcal{A}$ and vector $x$, the
Rayleigh-quotient bound gives
\begin{equation}
\left|
x^T\mathcal{A}x
\right|
\le
\|\mathcal{A}\|_2
\|x\|_2^2.
\end{equation}
Applying this inequality to
Eq.~(\ref{eq:app_response_error_1}) yields
\begin{align}
&
\left|
\varepsilon_{\mathrm{gen}}
\left(
\mathcal{H}_{\mathrm{bd}}^{(\theta_0)}
\right)
-
\varepsilon_{\mathrm{gen}}
\left(
\widetilde{\mathcal{H}}^{(\theta_0)}
\right)
\right|
\nonumber\\
&\le
\frac{1}{2}
\|\Delta\theta\|_2^2
\left\|
\mathcal{H}_{\mathrm{bd}}^{(\theta_0)}
-
\widetilde{\mathcal{H}}^{(\theta_0)}
\right\|_2.
\label{eq:app_response_error_2}
\end{align}

Because
$\mathcal{H}_{\mathrm{bd}}^{(\theta_0)}
-
\widetilde{\mathcal{H}}^{(\theta_0)}$
is block diagonal, its spectral norm is the largest spectral norm
among its constituent blocks:
\begin{align}
&
\left\|
\mathcal{H}_{\mathrm{bd}}^{(\theta_0)}
-
\widetilde{\mathcal{H}}^{(\theta_0)}
\right\|_2
\nonumber\\
&=
\max_{1\le j\le N}
\left\|
\mathcal{H}^{(\theta_0^j)}
-
\widetilde{\mathcal{H}}^{(\theta_0^j)}
\right\|_2.
\label{eq:app_block_norm}
\end{align}

For operation $j$,
\begin{align}
\mathcal{H}^{(\theta_0^j)}
-
\widetilde{\mathcal{H}}^{(\theta_0^j)}
&=
V^j
\left(
\Lambda^j-\gamma^j I_{|\theta^j|}
\right)
(V^j)^T.
\end{align}
Hence, its eigenvalues are
$
\lambda_i^j-\gamma^j
$,
and therefore
\begin{align}
&
\left\|
\mathcal{H}^{(\theta_0^j)}
-
\widetilde{\mathcal{H}}^{(\theta_0^j)}
\right\|_2
\nonumber\\
&=
\max_{1\le i\le|\theta^j|}
\left|
\lambda_i^j-\gamma^j
\right|.
\label{eq:app_group_spectral}
\end{align}

Since
$
\lambda_1^j
\ge
\lambda_i^j
\ge
\lambda_{|\theta^j|}^j
$
and $\gamma^j$ is the mean of the top-$K$ eigenvalues, we have
\begin{equation}
\lambda_{|\theta^j|}^j
\le
\gamma^j
\le
\lambda_1^j.
\end{equation}
Consequently,
\begin{equation}
\max_{1\le i\le|\theta^j|}
\left|
\lambda_i^j-\gamma^j
\right|
=
\max
\left\{
\lambda_1^j-\gamma^j,\,
\gamma^j-\lambda_{|\theta^j|}^j
\right\}.
\label{eq:app_endpoint_bound}
\end{equation}

Combining Eqs.~(\ref{eq:app_response_error_2}),
(\ref{eq:app_block_norm}), and
(\ref{eq:app_endpoint_bound}), we finally obtain
\begin{align}
&
\left|
\varepsilon_{\mathrm{gen}}
\left(
\mathcal{H}_{\mathrm{bd}}^{(\theta_0)}
\right)
-
\varepsilon_{\mathrm{gen}}
\left(
\widetilde{\mathcal{H}}^{(\theta_0)}
\right)
\right|
\nonumber\\
&\le
\frac{1}{2}
\|\Delta\theta\|_2^2
\max_{1\le j\le N}
\left\{
\lambda_1^j-\gamma^j,\,
\gamma^j-\lambda_{|\theta^j|}^j
\right\}.
\label{eq:app_final_response_bound}
\end{align}

This bound characterizes the additional discrepancy introduced by
compressing each operation-wise Hessian block into a group-isotropic
scalar representation. Importantly,
$\mathcal{H}_{\mathrm{bd}}^{(\theta_0)}$ itself is an operation-wise
block-diagonal approximation of the complete source-loss Hessian:
the cross-operation curvature blocks are omitted rather than assumed
to vanish. Therefore,
Eq.~(\ref{eq:app_final_response_bound}) quantifies the error induced
by the subsequent group-isotropic scalarization, rather than the total
discrepancy between
$\widetilde{\mathcal{H}}^{(\theta_0)}$
and the complete source-loss Hessian.

Moreover, the scalarization analyzed above is not intended to preserve
the within-operation eigenspace or directional curvature geometry.
Proposition~1 characterizes the information loss of the corresponding
exact spectral scalarization in terms of the Hessian eigenvalues
$\lambda_i^j$. In practical SRFT, these exact eigenvalues are not
explicitly computed; instead, Rayleigh-quotient probing provides
observed directional curvature responses, whose retained top-$K$ mean
is used as the operation-wise prior-significance statistic for
parameter-update regulation.

\subsection{Derivation of Transfer Significance}
\label{sec:a-2-2}

We state the local assumptions under which the analysis in
Eqs.~(9)--(12) holds and provide a more explicit derivation.

\paragraph{Assumptions.}
\textbf{Lipschitz-continuous gradients.}
The loss $\mathcal{L}(\theta\mid M)$ is differentiable and
$\beta$-smooth in a neighborhood of the current $\theta$, i.e.,
\[
\|\nabla_\theta \mathcal{L}(\theta+\Delta\mid M)
-\nabla_\theta \mathcal{L}(\theta\mid M)\|_2
\le \beta\|\Delta\|_2.
\]
This implies a first-order Taylor expansion with a quadratic remainder:
\[
\mathcal{L}(\theta+\delta\mid M)
=
\mathcal{L}(\theta\mid M)
+
\mathcal{G}^{\top}\delta
+
r(\delta),
\]
where
$\mathcal{G}=\nabla_\theta\mathcal{L}(\theta\mid M)$ and
$|r(\delta)|\le\frac{\beta}{2}\|\delta\|_2^2$.
\\

\noindent
\textbf{Bounded single-step gradient magnitude.}
The per-step gradient magnitude is bounded, as commonly assumed in
stochastic optimization. This assumption is used when relating gradient
sparsity to the amplification of $\|\mathcal{G}\|_2$ under a fixed
$\|\mathcal{G}\|_1$.
\\

\noindent
\textbf{Explicit stochastic-update model.}
$\delta$ and $\delta'$ are independent stochastic parameter
displacements centered at the same nominal gradient-descent step,
as described in Eq.~(10).

\paragraph{Setup.}
Let
$\mathcal{G}=\nabla_\theta\mathcal{L}(\theta\mid M)$ and recall:
\begin{align}
\epsilon_{\text{ada}}
&=
\mathbb{E}\left[
\left|
S(\theta,\delta\mid M)
-
S(\theta,\delta'\mid M)
\right|
\right],\\
S(\theta,\delta\mid M)
&=
\mathcal{L}(\theta+\delta\mid M)
-
\mathcal{L}(\theta\mid M).
\end{align}
We model two stochastic parameter displacements around the same
nominal gradient-descent step as
\begin{equation}
\delta
=
-\alpha\mathcal{G}+\alpha\xi,
\qquad
\delta'
=
-\alpha\mathcal{G}+\alpha\xi',
\label{eq:app_noise_model}
\end{equation}
where
$\xi,\xi'\overset{\text{i.i.d.}}{\sim}
\mathcal{N}(0,\sigma^2I)$.
Equivalently,
$\delta,\delta'\sim
\mathcal{N}(-\alpha\mathcal{G},\alpha^2\sigma^2I)$.
Thus, $\delta$ and $\delta'$ share the same nominal descent step
$-\alpha\mathcal{G}$ but differ in their stochastic components.

\paragraph{Linearization (Lipschitz continuity).}
Since $\mathcal{L}(\theta\mid M)$ is locally $\beta$-smooth around
$\theta$, a first-order Taylor expansion gives
\begin{equation}
\mathcal{L}(\theta+\delta\mid M)
=
\mathcal{L}(\theta\mid M)
+
\mathcal{G}^{\top}\delta
+
r(\delta),
\qquad
|r(\delta)|
\le
\frac{\beta}{2}\|\delta\|_2^2.
\label{eq:app_taylor}
\end{equation}
Hence,
\begin{equation}
S(\theta,\delta\mid M)
=
\mathcal{G}^{\top}\delta+r(\delta).
\end{equation}
Therefore,
\begin{equation}
\begin{aligned}
&S(\theta,\delta\mid M)
-
S(\theta,\delta'\mid M)
\\
&=
\mathcal{G}^{\top}(\delta-\delta')
+
r(\delta)-r(\delta').
\end{aligned}
\label{eq:app_Sdiff}
\end{equation}
When the stochastic displacements are sufficiently small such that
the Taylor-remainder term is dominated, the leading term becomes
\begin{equation}
\begin{aligned}
S(\theta,\delta\mid M)
-
S(\theta,\delta'\mid M)
&\approx
\mathcal{G}^{\top}(\delta-\delta')
\\
&=
\alpha\,\mathcal{G}^{\top}(\xi-\xi').
\end{aligned}
\label{eq:app_linear_main}
\end{equation}

\paragraph{Upper bound for the linearized response.}
Let
$X:=\alpha\,\mathcal{G}^{\top}(\xi-\xi')$.
Since
$\xi-\xi'\sim\mathcal{N}(0,2\sigma^2I)$, we have
\[
X
\sim
\mathcal{N}\!\left(
0,\,
2\alpha^2\sigma^2\|\mathcal{G}\|_2^2
\right).
\]
Thus,
\begin{equation}
\mathbb{E}|X|
\le
\sqrt{\mathbb{E}[X^2]}
=
\sqrt{\mathrm{Var}(X)}
=
\alpha\sqrt{2\sigma^2}\,
\|\mathcal{G}\|_2.
\label{eq:app_bound}
\end{equation}
Combining Eqs.~\eqref{eq:app_linear_main}
and~\eqref{eq:app_bound}, the linearized adaptation risk satisfies
\begin{equation}
\epsilon_{\text{ada}}
\approx
\mathbb{E}|X|
\le
\alpha\sqrt{2\sigma^2}\,
\|\mathcal{G}\|_2.
\end{equation}
Therefore, to first order, the adaptation risk scales at most linearly
with $\|\mathcal{G}\|_2$.

\paragraph{Tighter closed form.}
Moreover, for a zero-mean Gaussian
$X\sim\mathcal{N}(0,s^2)$,
$\mathbb{E}|X|=s\sqrt{2/\pi}$.
Hence,
\begin{equation}
\begin{aligned}
\mathbb{E}|X|
&=
\sqrt{\frac{2}{\pi}}
\sqrt{2\alpha^2\sigma^2\|\mathcal{G}\|_2^2}
\\
&=
\frac{2\alpha\sigma}{\sqrt{\pi}}
\|\mathcal{G}\|_2
\\
&\le
\alpha\sqrt{2\sigma^2}\,
\|\mathcal{G}\|_2,
\end{aligned}
\end{equation}
where the last inequality follows from
$2/\sqrt{\pi}\le\sqrt{2}$.

\paragraph{Deviation from linearization.}
From Eqs.~\eqref{eq:app_Sdiff} and~\eqref{eq:app_taylor},
the approximation error is controlled by the Taylor remainders:
\begin{equation}
\begin{aligned}
\left|
\epsilon_{\text{ada}}
-
\mathbb{E}|X|
\right|
&\le
\mathbb{E}
\left|
r(\delta)-r(\delta')
\right|
\\
&\le
\frac{\beta}{2}
\mathbb{E}
\left(
\|\delta\|_2^2+
\|\delta'\|_2^2
\right)
\\
&=
\beta\,\mathbb{E}\|\delta\|_2^2.
\end{aligned}
\end{equation}
Thus, the first-order linear scaling is most accurate in locally smooth
regions and when the stochastic parameter displacements are sufficiently
small.

\paragraph{Connection to parameter-wise transfer significance.}
The above analysis shows that the leading-order adaptation risk is
controlled by the gradient magnitude $\|\mathcal{G}\|_2$. Since
\begin{equation}
\|\mathcal{G}\|_2^2
=
\sum_{n=1}^{|\theta|}
\left(
\frac{\partial\mathcal{L}(\theta\mid M)}
{\partial\theta_n}
\right)^2,
\end{equation}
the squared gradient of parameter $\theta_n$ gives its additive
contribution to the squared gradient magnitude. This motivates the
instantaneous parameter-wise transfer significance in Eq.~(13),
\begin{equation}
s^t_{n,i}
=
\left(
\frac{\partial
\mathcal{L}(\theta^{(i)}\mid M_i)}
{\partial\theta_n}
\right)^2.
\end{equation}
Accordingly, $s^t_{n,i}$ is interpreted as a local,
target-conditioned parameter-sensitivity signal rather than as an
independent adaptation-risk value for each parameter.

\subsection{Optimization Interpretation of Significance-Regularized Updates}
\label{sec:a-2-3}

The significance analysis determines the update mobility of individual
parameters, where the operation-wise prior significance is shared by
parameters belonging to the same operation and the transfer significance
is estimated parameter-wise. Here, we provide further details on how the
estimated significance is translated into parameter-update mobility.

Let $\alpha$ denote the learning rate and $d_{n,i}$ denote the update
direction produced by the base optimizer for parameter $n$ at iteration
$i$, before significance modulation, with the corresponding base update
\begin{equation}
\Delta\theta_{n,i}=-\alpha d_{n,i}.
\label{eq:base_update_app}
\end{equation}
Given the normalized combined significance
$s_{n,i}\in[0,0.99]$, we consider the local mobility-regularized
surrogate introduced in the main text:
\begin{equation}
\Delta\theta_i^{*}
=
\arg\min_{\Delta\theta}
\sum_n
\left[
d_{n,i}^{\top}\Delta\theta_n
+
\frac{\|\Delta\theta_n\|_2^2}
{2\alpha(1-s_{n,i})}
\right].
\label{eq:local_surrogate_app}
\end{equation}

The linear term preserves the update preference specified by the base
optimizer, whereas the quadratic term defines a significance-dependent
movement cost. In particular, as $s_{n,i}$ increases, the same parameter
displacement incurs a larger cost, thereby assigning lower update
mobility to more significant parameters.

Since Eq.~(\ref{eq:local_surrogate_app}) is separable across parameters,
its first-order optimality condition for parameter $n$ is
\begin{equation}
d_{n,i}
+
\frac{\Delta\theta_n}
{\alpha(1-s_{n,i})}
=
0.
\label{eq:local_optimality}
\end{equation}
Therefore, the exact minimizer of the adopted local surrogate is
\begin{equation}
\Delta\theta_{n,i}^{*}
=
-\alpha(1-s_{n,i})d_{n,i}
=
(1-s_{n,i})\Delta\theta_{n,i}.
\label{eq:local_solution}
\end{equation}
The corresponding parameter update is
\begin{equation}
\theta_n^{(i+1)}
=
\theta_n^{(i)}
+
\Delta\theta_{n,i}^{*}
=
\theta_n^{(i)}
+
(1-s_{n,i})\Delta\theta_{n,i},
\label{eq:local_srft_update}
\end{equation}
which is identical to the SRFT update in
Eq.~(\ref{eq:update}) of the main text.

For the AdamW optimizer used in our experiments,
$\Delta\theta_{n,i}$ denotes the complete update produced by the base
optimizer before significance modulation. Specifically, the raw
gradient is first used by AdamW to update its first- and second-moment
statistics and construct the optimizer update, including the decoupled
weight-decay contribution. SRFT does not scale the raw gradient before
the moment statistics are computed; instead, the significance factor
$(1-s_{n,i})$ is applied to the resulting complete AdamW update.
Thus, the base optimizer determines the update, whereas SRFT controls
its allowed mobility according to the estimated parameter significance.

We emphasize that the modulated update follows from the exact minimizer
of the adopted quadratic surrogate in
Eq.~(\ref{eq:local_surrogate_app}). This local optimization
interpretation does not imply that the linear mapping $1-s_{n,i}$ is
unique or globally optimal; other monotonic mappings from significance
to update mobility are in principle possible.

\section{Additional Ablation Studies} \label{sec:a-3}

\subsection{Comparison with Sensitivity-aware Sparse Tuning.} 
\label{sec:spt}
In our method, parameter significance is used to regularize the update of all parameters, rather than to select a subset of ``sensitive'' parameters for sparse fine-tuning as in SPT~\citep{he2023sensitivity}. This distinction is particularly important for cross-modal adaptation. To examine this, we compare our approach with SPT under identical training settings, varying only the trainable parameter ratio $\tau$ for SPT (top-$\tau$ sensitive parameters).

As reported in Tab.~\ref{tab:spt}, SPT exhibits a clear performance--sparsity trade-off: when $\tau$ decreases, tracking accuracy drops sharply (e.g., F-score $61.0 \rightarrow 60.2 \rightarrow 54.5$ for $\tau=50\%, 20\%, 10\%$), indicating that cross-modal adaptation requires distributed parameter updates rather than a highly sparse subset. Even with a relatively large budget ($\tau=50\%$), SPT remains essentially on par with full fine-tuning (F-score $61.0$ vs.\ $61.5$), suggesting that merely keeping the ``most sensitive'' parameters does not yield a meaningful adaptation gain beyond standard tuning. This observation is consistent with the nature of multi-modality tracking: the adaptation signal is typically heterogeneous across layers and operators, and restricting updates to a subset can easily break the coordinated adjustments needed for modality fusion and temporal consistency. In contrast, our method achieves a substantially higher F-score ($65.1$) while keeping $\tau=100\%$ trainable parameters, demonstrating that the key factor is not sparsifying the update but steering it. By reweighting parameter changes according to their significance, our regularization encourages the model to preserve parameters that are critical to general tracking behavior while allowing more flexible updates on parameters that are more transferable for cross-modal adaptation. As a result, our method improves both precision and recall (Pr $64.7$, Re $65.4$), indicating more accurate target localization and more complete target recovery under depth-induced appearance variations.

\vspace{-2pt}
\begin{table}[htbp]
\caption{Comparison of the tracking performance between SPT and our regularized fine-tuning method based on the DepthTrack dataset, using the pre-trained OSTrack-B256 as the base model. We have set up a series of trainable parameter ratios $\tau$ of SPT (top-$\tau$ sensitive parameters) to fully explore its adaptation effect. }
\centering
\renewcommand{\arraystretch}{1.1}
\setlength{\tabcolsep}{2.0pt}
\small
\begin{tabular}{c |c  c c }
   \toprule
    \textbf{Exp.} &Pr  &Re &F-score \\
    \hline
    w/o fine-tuning ($\tau=0\%$)  &38.2 &36.0  &37.1      \\ 
    full fine-tuning ($\tau=100\%$)  &61.6 &61.4  &61.5       \\ 
    SPT ($\tau=50\%$)  &61.1 &60.9  &61.0       \\ 
    SPT ($\tau=20\%$)  &60.7  &59.7 &60.2       \\ 
    SPT ($\tau=10\%$)  &56.2  &52.8 &54.5      \\ 
    \hline
    Ours ($\tau=100\%$)  &64.7  &65.4 &65.1      \\ 
    \bottomrule
\end{tabular}
\label{tab:spt}
\vskip -0.3in
\end{table}

\begin{table*}[htbp]
\centering
\caption{Computational cost of prior-significance curvature probing on OSTrack-B256 (128$\times$128/256$\times$256 input resolution for template and search regions). }
\renewcommand{\arraystretch}{1.0}
\setlength{\tabcolsep}{2.0pt}
\small
\begin{tabular}{l|c|ccc}
\toprule
\multirow{2}{*}{\textbf{Metric}} &\multirow{2}{*}{\textbf{Granularity}} & \multicolumn{3}{c}{\textbf{Computational Cost}} \\ 
 & & \textbf{Per-Group (Operation-wise)} & \textbf{Per-Iteration (Model-wise)} & \textbf{Total Offline Process} \\ 
\midrule
\textbf{FLOPs} & Theoretical Ops &104.8 GFLOPs &15.5 TFLOPs & $\sim$ 6.2 PFLOPs$^{\dagger}$ \\
\midrule
\textbf{Latency} & Wall-clock Time & $294.5$ ms & $43.0$ s &47.8 h ($\sim$2 day)$^{\dagger}$ \\
\midrule
\textbf{Memory} & Peak Usage (GPU) &$164.2$ GB (Batch=640, 8$\times$GPUs) & - &-  \\
\bottomrule
\multicolumn{4}{l}{\footnotesize $^{\dagger}$ Estimated over the full pre-trained tracking datasets. The preprocessing is performed only once.} \\
\end{tabular}%
\label{tab:fim_cost}
\vskip -0.1in
\end{table*}

\subsection{Comparison with Classical Parameter-Importance and Curvature-Aware Methods}
\label{app:classical_regularization}
We further compare SRFT with representative parameter-importance
regularization and curvature-aware optimization methods, including
L2-SP~\citep{xuhong2018explicit}, EWC~\citep{kirkpatrick2017overcoming},
RWalk~\citep{chaudhry2018riemannian}, SAM~\citep{foret2020sharpness},
and a diagonal empirical-Fisher approximation to natural-gradient
optimization (Diag-NG)~\citep{martens2015optimizing}. These baselines cover parameter anchoring,
source-importance regularization, optimization-path-aware importance,
sharpness-aware optimization, and curvature-based gradient
preconditioning.

\noindent\textbf{Experimental setting.}
All methods start from the same pre-trained OSTrack-B256 checkpoint and
use the same adaptation data, batch size, number of epochs, data
augmentation, base learning rate, weight decay, learning-rate schedule,
and evaluation protocol as FFT; all other settings remain unchanged.
L2-SP, EWC, RWalk, SAM, and SRFT use AdamW as the base optimizer,
whereas Diag-NG replaces the standard gradient direction with a
diagonally preconditioned direction.

We use fixed method-specific hyperparameters across FE108, DepthTrack,
and LasHeR: $\lambda_{\mathrm{SP}}=0.01$ for L2-SP,
$\lambda_{\mathrm{EWC}}=400$ for EWC,
$\lambda_{\mathrm{RW}}=1000$ and $\epsilon_{\mathrm{RW}}=0.1$ for
RWalk, $\rho=0.05$ for SAM, and $\delta=10^{-3}$ for Diag-NG.
No test sequence is used for hyperparameter selection. Thus, the
compared variants mainly differ in how parameter updates are regularized
or modulated.

\noindent\textbf{Compared methods.}

\noindent\textbf{\textit{L2-SP.}}
L2-SP constrains the adapted parameters toward their pre-trained
initialization $\theta_0$ through
\[
\mathcal{L}_{\mathrm{L2\text{-}SP}}
=
\mathcal{L}_{\mathrm{trk}}
+
\lambda_{\mathrm{SP}}
\|\theta-\theta_0\|_2^2.
\]
It therefore provides an isotropic parameter-anchoring baseline without
estimating parameter-specific source importance.

\noindent\textbf{\textit{EWC.}}
EWC introduces non-uniform source preservation according to
parameter importance. We estimate a diagonal empirical Fisher from the
source RGB tracking data as
\[
\widehat{F}_j^{\mathrm{src}}
=
\frac{1}{N}
\sum_{m=1}^{N}
\left(
\frac{\partial \ell_m}{\partial\theta_j}
\right)^2,
\]
and optimize
\[
\mathcal{L}_{\mathrm{EWC}}
=
\mathcal{L}_{\mathrm{trk}}
+
\lambda_{\mathrm{EWC}}
\sum_j
\widehat{F}_j^{\mathrm{src}}
\left(
\theta_j-\theta_j^0
\right)^2.
\]
The same source tracking data are used for estimating source importance
in EWC and prior significance in SRFT. Here, the empirical Fisher is
used only as the classical source-importance estimator adopted by EWC
and should not be confused with the source-loss Hessian used to formulate
prior significance in SRFT.

\noindent\textbf{\textit{RWalk.}}
RWalk combines source-domain parameter importance with
optimization-path information accumulated during adaptation. Let
$F_n^{\mathrm{src}}$ denote the source empirical-Fisher importance of
parameter $n$. The path-based importance is computed as
\[
S_n
=
\max
\left(
0,\,
\sum_t
\frac{
-g_{n,t}\Delta\theta_{n,t}
}{
\frac{1}{2}F_{n,t}(\Delta\theta_{n,t})^2
+
\epsilon_{\mathrm{RW}}
}
\right),
\qquad
\Delta\theta_{n,t}
=
\theta_{n,t+1}-\theta_{n,t},
\]
where $g_{n,t}$ is the current gradient and $F_{n,t}$ is the online
diagonal empirical-Fisher statistic. The combined importance is
\[
\Omega_n^{\mathrm{RW}}
=
F_n^{\mathrm{src}}
+
S_n,
\]
and RWalk optimizes
\[
\mathcal{L}_{\mathrm{RWalk}}
=
\mathcal{L}_{\mathrm{trk}}
+
\lambda_{\mathrm{RW}}
\sum_n
\Omega_n^{\mathrm{RW}}
\left(
\theta_n-\theta_{0,n}
\right)^2.
\]
Thus, RWalk extends fixed source-domain importance with
optimization-path information accumulated during adaptation.

\noindent\textbf{\textit{SAM.}}
For each mini-batch, SAM first computes
\[
g_t
=
\nabla_{\theta}
\mathcal{L}_{\mathrm{trk}}(\theta_t)
\]
and constructs the parameter perturbation
\[
\delta_t^{\mathrm{SAM}}
=
\rho
\frac{
g_t
}{
\|g_t\|_2+\epsilon_{\mathrm{SAM}}
},
\qquad
\rho=0.05,
\qquad
\epsilon_{\mathrm{SAM}}=0.001.
\]
The loss is then evaluated at
$\theta_t+\delta_t^{\mathrm{SAM}}$, and the resulting gradient is
passed to AdamW to update the original parameters $\theta_t$.
SAM therefore provides a representative sharpness-aware baseline by
optimizing the loss under a local parameter perturbation.

\noindent\textbf{\textit{Diag-NG.}}
For natural-gradient optimization, constructing and inverting a
full-matrix Fisher is computationally prohibitive for OSTrack-B256.
We therefore use a diagonal empirical-Fisher approximation,
\begin{equation}
\widehat{F}_n
=
\frac{1}{B}
\sum_{m=1}^{B}
\left(
\frac{\partial
\ell(\theta;x_m,y_m)}
{\partial\theta_n}
\right)^2,
\label{eq:diag_fisher}
\end{equation}
and precondition the gradient as
\begin{equation}
\widetilde{g}_n
=
\frac{g_n}
{\widehat{F}_n+\delta},
\qquad
\delta=10^{-3}.
\label{eq:diag_ng}
\end{equation}
Diag-NG directly updates the parameters using this diagonally
preconditioned direction rather than applying AdamW moment
normalization. We refer to this practical baseline as Diag-NG.
Here, the empirical Fisher is used specifically as the approximation
adopted by this baseline and is distinct from the source-loss Hessian
formulation used for SRFT.

\begin{table*}[t]
\centering
\caption{
Comparison with representative parameter-importance regularization,
sharpness-aware optimization, and curvature-aware optimization methods.
Diag-NG denotes a diagonal empirical-Fisher approximation to
natural-gradient optimization. All variants use the same pre-trained
tracker and adaptation protocol; all other settings remain unchanged.
Best results are shown in bold.
}
\vskip -0.1in
\small
\setlength{\tabcolsep}{4.0pt}
\begin{tabular}{l|l|cc|ccc|cc}
\toprule
\multirow{2}{*}{\textbf{Method}}
&
\multirow{2}{*}{\textbf{Regularization / Update Criterion}}
&
\multicolumn{2}{c|}{FE108}
&
\multicolumn{3}{c|}{DepthTrack}
&
\multicolumn{2}{c}{LasHeR}
\\
\cline{3-9}
&
&
SR & PR
&
Pr & Re & F-score
&
SR & PR
\\
\midrule
FFT
& None
& 65.2 & 93.5
& 61.6 & 61.4 & 61.5
& 53.2 & 65.8
\\

L2-SP
& Distance to initialization
& 65.6 & 94.0
& 62.4 & 62.5 & 62.4
& 53.8 & 66.6
\\

EWC
& Source parameter importance
& 65.9 & 94.3
& 62.9 & 63.0 & 62.9
& 54.2 & 67.2
\\

RWalk
& Importance + optimization path
& 66.1 & 94.5
& 63.2 & 63.1 & 63.1
& 54.5 & 67.6
\\

SAM
& Local sharpness
& 65.8 & 94.2
& 62.7 & 62.9 & 62.8
& 54.0 & 67.0
\\

Diag-NG
& Diagonal empirical-Fisher preconditioning
& 66.0 & 94.4
& 63.0 & 63.1 & 63.0
& 54.3 & 67.4
\\

SRFT (Ours)
& Prior + transfer significance
& \textbf{67.4} & \textbf{96.5}
& \textbf{64.7} & \textbf{65.4} & \textbf{65.1}
& \textbf{56.3} & \textbf{70.1}
\\
\bottomrule
\end{tabular}
\label{tab:classical_regularization}
\vskip -0.1in
\end{table*}

\noindent\textbf{Results.}
As shown in Table~\ref{tab:classical_regularization}, all classical
regularization and optimization strategies improve over unconstrained
FFT, confirming the benefit of controlling parameter updates during
cross-modal adaptation. Among these alternatives, RWalk achieves the
strongest overall performance, while SRFT consistently obtains the best
results across FE108, DepthTrack, and LasHeR.

The compared methods regulate adaptation from different perspectives.
L2-SP uses uniform proximity to the pre-trained parameters; EWC
introduces fixed source-domain parameter importance; RWalk additionally
incorporates optimization-path information; SAM regularizes local
sharpness; and Diag-NG modifies the gradient through curvature-based
preconditioning. In contrast, SRFT combines prior significance, which
characterizes source-domain sensitivity, with transfer significance,
which captures the evolving target-domain optimization response. The
consistent improvements indicate that this joint source- and target-aware
significance modulation provides additional benefits beyond the
representative parameter-anchoring, importance-based, sharpness-aware,
and curvature-preconditioning baselines considered here.

These approaches are conceptually related rather than mutually
exclusive. The comparison therefore does not imply that SRFT replaces
the underlying regularization or optimization principles; instead, it
shows the effectiveness of significance-guided update modulation under
the same cross-modal adaptation setting.

\subsection{Computational Overhead of Prior Significance}
\label{app:prior_efficiency}
Estimating prior significance introduces an additional offline preprocessing stage. SRFT does not explicitly construct or eigendecompose the full source-loss Hessian, whose storage alone would scale as $\mathcal{O}(P^2)$ for a tracker with $P$ parameters and is infeasible for models such as OSTrack-B256 ($\sim$86M parameters). Instead, SRFT partitions the backbone into operation groups and estimates dominant directional curvature responses through Rayleigh-quotient probing with symmetric finite differences. Thus, the practical cost is mainly determined by the number of operation groups, probing directions, and source-loss evaluations, rather than explicit Hessian construction. Reducing the probing ratio directly decreases this cost, as analyzed in Appendix~\ref{app:probing_efficiency}.

As reported in Table~\ref{tab:fim_cost}, the 100\%-probing configuration on OSTrack-B256 requires about 104.8 GFLOPs and 294.5 ms per operation group, 15.5 TFLOPs and 43.0 s per model-wise probing iteration, and 47.8 h in total on 8 NVIDIA RTX 3090 Ti GPUs. The peak memory of 164.2 GB mainly arises from probing-batch activations rather than Hessian storage and can be reduced with a smaller probing batch. Although this offline cost is non-negligible, prior significance is a reusable checkpoint-level statistic and introduces no additional model parameters or inference FLOPs. Moreover, Appendix~\ref{app:probing_efficiency} shows that sparse probing can substantially reduce preprocessing cost while retaining most of the downstream benefit.

\begin{figure}[t]
\centering
\includegraphics[clip, width=0.48\textwidth]{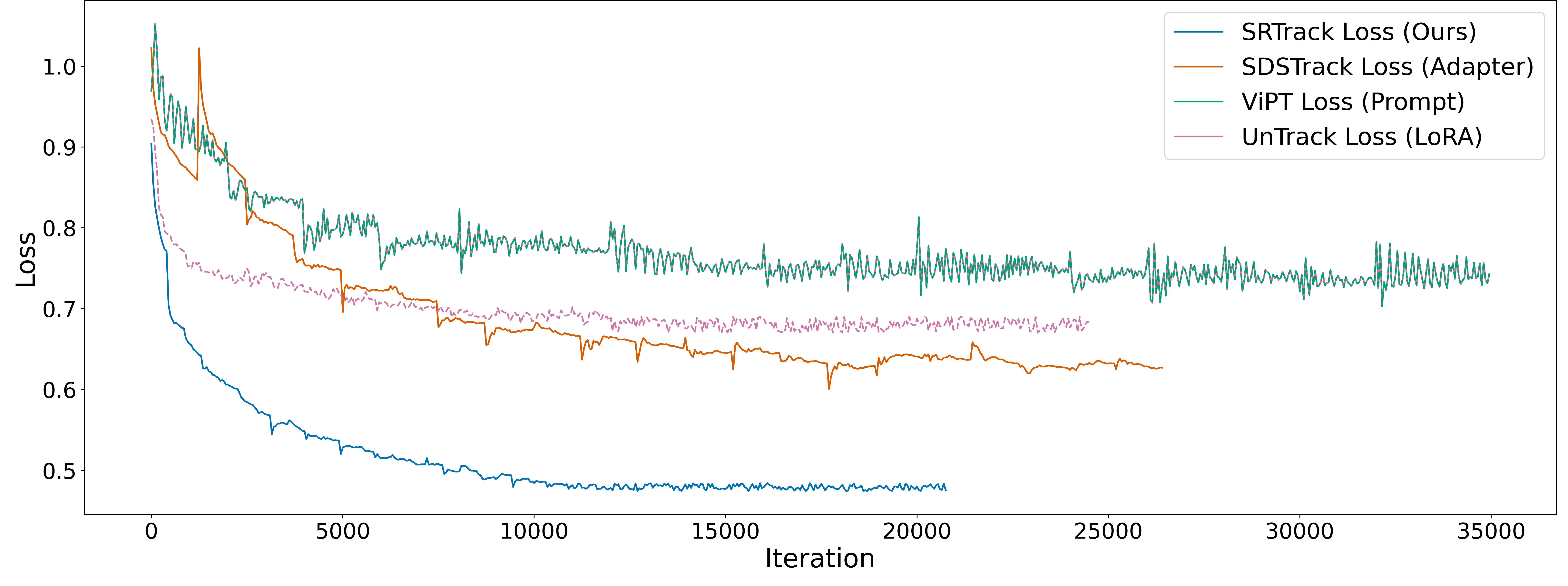}
\vskip -0.15in
\caption{Comparison of training loss for different multi-modality tracking methods on DepthTrack dataset. Note that all methods adopt pre-trained OSTrack initialization. Evidently, our method converges in fewer training iterations and reaches a lower loss value. Moreover, the training dynamics are noticeably more stable, exhibiting reduced oscillation throughout fine-tuning.}
\label{fig:loss}
\vskip -0.2in
\end{figure}

\subsection{Training Efficiency of SRFT}
\label{app:training_efficiency}
We explicitly acknowledge that unlike plug-and-play PEFT methods (e.g., Prompt Tuning, Adapters, LoRA), which incur zero preprocessing cost via random initialization, our SRFT involves an offline prior-significance curvature estimation phase. However, we argue that this offline computation represents a strategic \textbf{``pay once, benefit long-term"} investment. This upfront cost is justified by the following characteristics over PEFT:

\noindent \textbf{1. Accelerated Convergence:} The prior-significance estimation is a one-shot process performed on the source RGB tracking data. The resulting significance map is then frozen and reused for subsequent multi-modality tracking tasks based on the same pre-trained checkpoint. Rather than relying on a small set of newly introduced task-specific parameters as in many PEFT methods, SRFT directly modulates the updates of the pre-trained backbone using significance-aware regularization. As shown in Tab.~\ref{tab:train time} and Fig.~\ref{fig:loss}, SRFT reaches its optimal performance with fewer training epochs in the evaluated configurations. Depending on the base method, this reduction can partially or, in some cases, fully offset the offline preprocessing cost; however, the total time-to-result is not uniformly lower because prior-significance estimation introduces a non-negligible one-time cost.

\noindent \textbf{2. Zero Inference Latency:} This is of paramount importance for real-time applications. As detailed in the original manuscript, the proposed regularization techniques are applied exclusively during training. SRFT itself introduces no additional modules, parameters, or inference FLOPs once the significance statistics are obtained.

\noindent \textbf{3. Superior Performance:} Despite the additional offline cost, SRFT consistently improves tracking accuracy in the evaluated settings, demonstrating a favorable accuracy–computation trade-off.

\begin{table*}[thbp]
\centering
\vskip -0.1in
\caption{Comparison of preprocess time, trainable parameter and training time across different multi-modality tracking methods. All experiments are conducted based on the pre-trained OSTrack-B256. ``Preprocessing” refers to prior significance estimation. The training epochs (listed sequentially from left to right) correspond to the FE108, VisEvent, CoeSot, LasHeR, and DepthTrack datasets, respectively. Note that UnTrack employs a joint training strategy on the VisEvent, LasHeR, and DepthTrack datasets.}
\renewcommand{\arraystretch}{1.0}
\setlength{\tabcolsep}{4.0pt}
\footnotesize
\begin{tabular}{c|c|c|c|c|c}
\toprule
\textbf{Method} & \textbf{Trainable Parameter (M)} & \textbf{Preprocess Time (h)} & \textbf{Total Epochs} & \textbf{Training Time (h)} & \textbf{Total Time (h)} \\
\midrule
ViPT (Prompt) &0.8 & - &300 \scriptsize{$(60 \times 5)$} & $15.3 min \times 300 \approx 76.5 \text{ h}$ & 76.5 \\
\midrule
ViPT + SRFT &86.4 & 47.8 h &100 \scriptsize{$(20 \times 5)$} & $15.9 min \times 100 \approx 26.5 \text{ h}$ & 74.3 \\
\midrule
SDSTrack (Adapter) &14.8 & - &205 \scriptsize{$(50\times3+40+15)$} & $21.9 min \times 205 \approx 74.8 \text{ h}$ &74.8 \\
\midrule
SDSTrack + SRFT &100.6 & 47.8 h &85 \scriptsize{$(20\times3+15+10)$} & $22.8 min \times 85 \approx 32.3 \text{ h}$ & 80.1 \\
\midrule
UnTrack (LoRA) &6.6 & - &80 &$18.5 min \times 80 \approx 19.3 \text{h}$ &19.3 \\
\midrule
UnTrack + SRFT &92.4 & 47.8 h & $40 $ &$19.4 min \times 40 \approx 10.3 \text{ h}$ & 58.1 \\
\bottomrule
\end{tabular}%
\label{tab:train time}
\end{table*}

\subsection{Reduced Rayleigh-Quotient Probing}
\label{app:probing_efficiency}

The reported 47.8-hour preprocessing time in Table~\ref{tab:fim_cost} corresponds to the
100\% Rayleigh-quotient probing configuration used in our current
experiments. Unless otherwise specified, the main experimental results are reported using the 100\%-probing, $K=10$ reference configuration. The 20\%-probing, top-1 setting is evaluated separately as an efficiency-oriented alternative to characterize the accuracy--preprocessing-cost trade-off. This setting serves only as a comprehensive experimental reference (and as the denominator for reporting relative cost); it is not a required configuration for applying SRFT. We therefore systematically investigate reduced probing ratios and report the resulting accuracy--efficiency trade-off. 

We first clarify the respective roles of the probing ratio and the
number of retained curvature responses. In our implementation, Rayleigh-quotient
probing is performed before selecting the top-$K$ observed curvature responses. Consequently, the preprocessing time is determined primarily by the number of probes. Therefore, the 100\%-probing top-10 and 100\%-probing top-1 configurations require the same preprocessing time. In contrast, the probing batch size mainly
affects peak GPU memory through the activations required during source-loss
evaluation, without changing the definition of the retained-response
statistic. The purpose of the top-1 experiment ($K=1$) is not to reduce the
probing cost, but to examine whether retaining only the dominant
observed curvature response is sufficient to construct an informative
operation-level prior-significance score.

\begin{figure}[t]
\centering
\includegraphics[clip, width=0.48\textwidth]{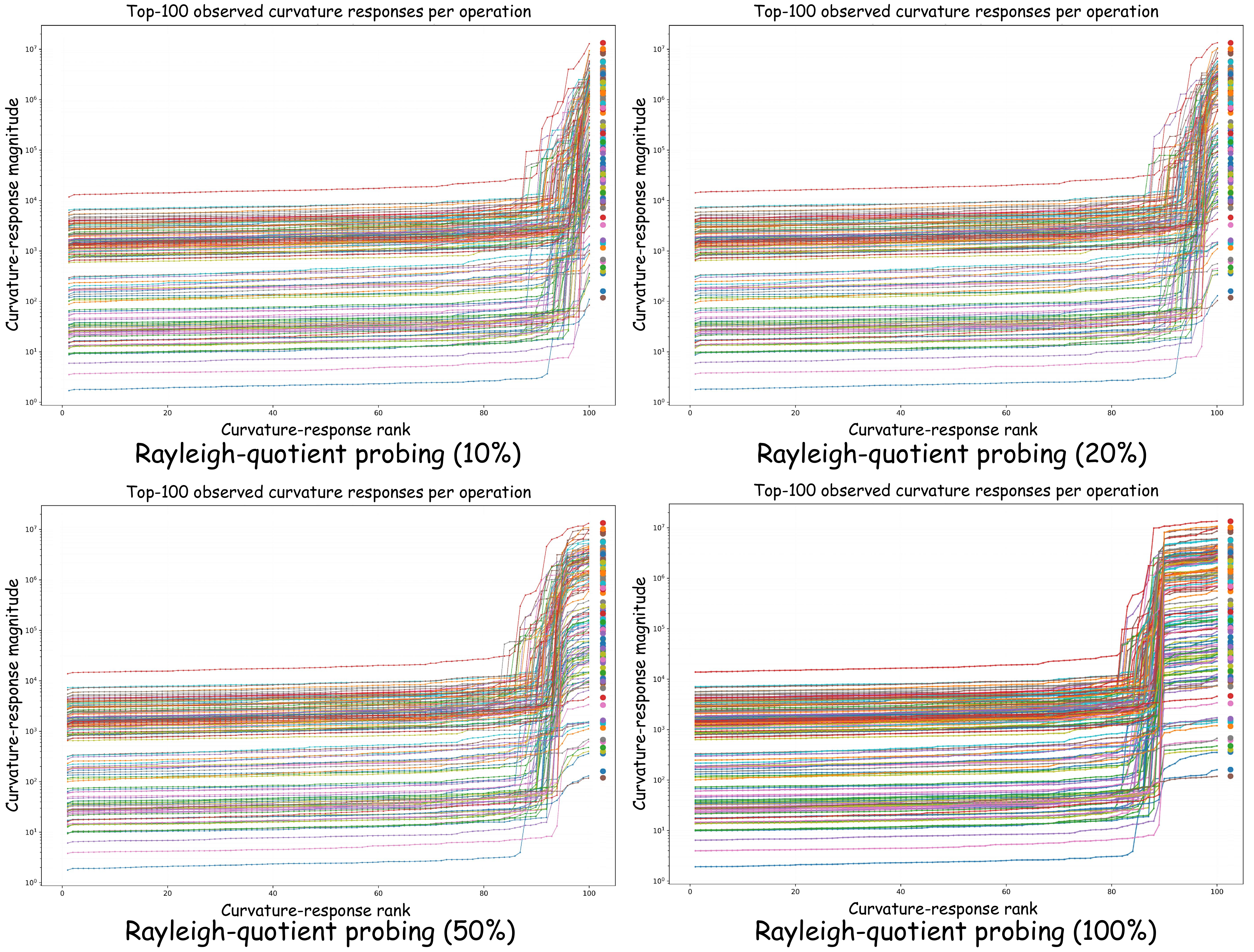}
\vskip -0.1in
\caption{
Operation-wise sampled curvature responses using different
Rayleigh-quotient probing ratios. Each panel shows the top-100
observed curvature responses obtained with 10\%, 20\%, 50\%, and 100\% probing,
respectively. Reduced probing preserves the overall curvature-response distribution,
dominant operation ordering, and transition between the leading
curvature responses and the low-magnitude response tail.
}
\label{fig:r1}
\vskip -0.15in
\end{figure}

As shown in Table~\ref{tab:r1}, replacing top-10 with top-1 under
100\% probing changes every reported metric by at most 0.4 points across
FE108, DepthTrack, and LasHeR. This confirms the observation in the
existing $K$-ablation that the dominant curvature response
already captures most of the relative operation-level sensitivity
required by prior-significance regularization. We further evaluate top-1 estimation using 50\%, 20\%, and 10\%
Rayleigh-quotient probing. Figure~\ref{fig:r1} shows that reduced
probing preserves the overall operation-wise observed curvature-response distribution and
dominant-operation ordering, although estimation variations become more
visible as the probing ratio decreases. The corresponding downstream
results show a gradual accuracy--efficiency trade-off.

Compared with the current 100\%-probing top-10 reference, 20\%-probing
top-1 decreases FE108 by only 0.5 SR and 0.5 PR, DepthTrack by 0.3
F-score, and LasHeR by 0.6 SR and 0.5 PR. Meanwhile, it reduces the
preprocessing time from 47.8 hours to approximately 9.6 hours (\textbf{5.0$\times$ speedup}). Even the more aggressive 10\%-probing setting retains most of the benefit. Accordingly, we recommend \textbf{20\% probing with top-1} as the efficient
configuration, since it provides a more favorable balance between
estimation cost and tracking accuracy. Overall, these experiments demonstrate that prior significance can be estimated with substantially fewer probing iterations while retaining most of its downstream benefit.

\begin{table*}[t]
\centering
\vskip -0.1in
\caption{
Accuracy--efficiency comparison of prior-significance estimation on
OSTrack-B256. The 100\%-probing top-10 setting serves as the experimental
reference. The 20\%-probing top-1 setting provides a favorable accuracy-efficiency trade-off. Best results are in bold, and the recommended efficiency-oriented setting is underlined.
}
\renewcommand{\arraystretch}{1.05}
\setlength{\tabcolsep}{3.6pt}
\footnotesize
\begin{tabular}{cc|cc|ccc|cc|cc}
\toprule
\multirow{2}{*}{Probing Ratio}
& \multirow{2}{*}{Curvature-Response Budget (Top-$K$)}
& \multicolumn{2}{c|}{FE108}
& \multicolumn{3}{c|}{DepthTrack}
& \multicolumn{2}{c|}{LasHeR}
& \multicolumn{2}{c}{\textbf{Preprocessing Cost}} \\
\cline{3-11}
&
& SR & PR 
& Pr  & Re  & F-score 
& SR  & PR 
& Time  &Speedup \\
\midrule

100\%
& $K=10$
& \textbf{66.8} & \textbf{95.4}
& \textbf{64.1} & \textbf{64.5} & \textbf{64.3}
& \textbf{55.6} & \textbf{69.0}
& 47.8 h & $1.0\times$ \\

100\%
& $K=1$
& 66.5 & 95.0
& 63.9 & 64.3 & 64.1
& 55.2 & 68.7
& 47.8 h & $1.0\times$ \\

50\%
& $K=1$
& 66.5 & 94.9
& 63.9 & 64.2 & 64.1
& 55.3 & 68.7
& 23.9 h & $2.0\times$ \\

20\%
& $K=1$
& \underline{66.3} & \underline{94.9}
& \underline{63.8} & \underline{64.2} & \underline{64.0}
& \underline{55.0} & \underline{68.5}
& 9.6 h & \textbf{$5.0\times$} \\

10\%
& $K=1$
& 65.8 & 94.4
& 63.1 & 63.5 & 63.3
& 54.5 & 67.6
& 4.8 h & $10.0\times$ \\

\cline{1-11}

0\% (FFT)
& -
& 65.2 & 93.5
& 61.6 & 61.4 & 61.5
& 53.2 & 65.8
& 0 h &- \\

\bottomrule
\end{tabular}
\label{tab:r1}
\end{table*}

\subsection{Prior Significance Overhead under Larger Backbones}
In addition to the pre-trained trackers used in this work (e.g., OSTrack-B256 with a ViT-B backbone), we further investigate the FLOPs and memory costs of the curvature-probing-based prior significance estimation under larger backbones, such as ViT-L and Swin-L. Based on empirical measurements conducted on a cluster of 8 NVIDIA RTX 3090 Ti GPUs, the specific computational overhead introduced by prior-significance curvature probing for ViT-L and Swin-L is detailed in Table~\ref{tab:fim_cost_large}. It is important to note that the total cost of the offline prior-significance estimation probing process is data-dependent, scaling with the size of the source RGB tracking data and the number of iterations required for convergence given the model capacity. Consequently, rather than providing a variable total estimate, we report the deterministic per-group and per-iteration costs to accurately reflect the computational costs.

\begin{table*}[thbp]
\centering
\vskip -0.1in
\caption{Computational cost of prior-significance curvature probing on ViT-L and Swin-L (256$\times$256 input resolution).}
\renewcommand{\arraystretch}{0.9}
\setlength{\tabcolsep}{6.5pt}
\footnotesize
\label{tab:fim_cost_large}
\begin{tabular}{c|c|c|cc}
\toprule
\multirow{2}{*}{\textbf{Model}} &\multirow{2}{*}{\textbf{Metric}} &\multirow{2}{*}{\textbf{Granularity}}  & \multicolumn{2}{c}{\textbf{Computational Cost}} \\ 
 & & & \textbf{Per-Group (Operation-wise)} & \textbf{Per-Iteration (Model-wise)} \\ 
\midrule

\multirow{4}{*}{\textbf{ViT-L}} & \textbf{FLOPs} & Theoretical Ops & 170.2 GFLOPs & 49.4 TFLOPs \\ 
\cmidrule{2-5}
 & \textbf{Latency} & Wall-clock Time & 452.5 ms & 131.2 s \\ 
\cmidrule{2-5}
 & \textbf{Memory} & Peak Usage (GPU) &147.6 GB (Batch=200, 8$\times$GPUs) &- \\ 
 \midrule

\multirow{4}{*}{\textbf{Swin-L}} & \textbf{FLOPs} & Theoretical Ops & 323.1 GFLOPs & 93.7 TFLOPs \\ 
\cmidrule{2-5}
 & \textbf{Latency} & Wall-clock Time & 687.8 ms & 206.3 s \\ 
\cmidrule{2-5}
 & \textbf{Memory} & Peak Usage (GPU) &168.6 GB (Batch=200, 8$\times$GPUs) &- \\ 
\bottomrule
\end{tabular}%
\end{table*}

\subsection{Robustness to Incomplete Source RGB Tracking Data}
\label{app:proxy_data}
In practice, the complete source RGB tracking dataset used for the source
tracker may not always be accessible. We therefore further evaluate a
proxy-data setting to examine whether prior significance can still be
estimated from only a partial public source dataset.

Specifically, we compare three configurations:
\textbf{(i)} no prior significance,
\textbf{(ii)} prior significance estimated from the complete RGB-VOT
training set, and
\textbf{(iii)} prior significance estimated using GOT-10k alone as a
public proxy. In the proxy setting, LaSOT, COCO, TrackingNet, and all
generic backbone pre-training data are excluded; GOT-10k is used
exclusively for prior-significance estimation.

As shown in Table~\ref{tab:r2}, using GOT-10k alone consistently
outperforms the no-prior baseline and remains within 0.9 points of the
complete-data setting across all reported metrics. These results
indicate that prior significance can be estimated effectively even when
the original source training set is only partially accessible, and that
a single public tracking dataset can serve as a practical proxy. We emphasize that this is a proxy-data setting rather than a
data-free setting: SRFT still requires source-domain tracking samples
for prior-significance estimation, but does not require access to the
complete original training corpus.

\begin{table}[t]
\centering
\vskip -0.1in
\caption{Proxy-data evaluation of prior-significance estimation on the pre-trained OSTrack-B256.
    ``Full RGB-VOT $D_0$'' uses LaSOT, COCO, TrackingNet, and GOT-10k,
whereas ``GOT-10k only $D_0$'' uses GOT-10k as the sole proxy dataset for prior-significance estimation.}
\renewcommand{\arraystretch}{1.0}
\setlength{\tabcolsep}{3.0pt}
\normalsize
\small
\begin{tabular}{c | c c| c c c |c c}
   \toprule
    \multirow{2}{*}{\textbf{Exp.}} &\multicolumn{2}{c|}{FE108} &\multicolumn{3}{c|}{DepthTrack} &\multicolumn{2}{c}{LasHeR}  \\
    \cline{2-8}
    \multicolumn{1}{c|}{}  &SR &PR &Pr  &Re &F-score  &SR &PR \\
    \hline
    \textbf{FFT}    & 65.2 & 93.5 & 61.6 & 61.4 & 61.5 & 53.2 & 65.8    \\
    \textbf{Full RGB-VOT $D_0$}   &66.8 &95.4 &64.1 &64.5 &64.3 &55.6 &69.0 \\
    \textbf{GOT-10k only $D_0$}   & 66.2 & 95.1 & 63.7 & 63.9 & 63.8 & 55.1 & 68.1   \\
    \bottomrule
\end{tabular}
\label{tab:r2}
\vskip -0.15in
\end{table}

\subsection{Alternative Prior-Significance Estimators}
\label{app:prior_estimators}



Prior significance in SRFT is constructed as an operation-level scalar
sensitivity rather than a matrix-level reconstruction of the source-loss
Hessian. For operation $j$, let
$\mathcal{H}^{(\theta_0^j)}$
denote the corresponding source-loss Hessian block. In practical SRFT,
we estimate its local directional curvature through Rayleigh-quotient
probing. Let $q_{j,(k)}$ denote the $k$-th largest retained observed
curvature response for operation $j$. We summarize the retained
responses as
\begin{equation}
\bar{q}_j
=
\frac{1}{K}
\sum_{k=1}^{K}
q^{j}_{(k)}.
\label{eq:alternative_prior}
\end{equation}
 Here, $\bar{q}_j$ corresponds to the operation-level
prior-significance statistic in Eq.~(8). Before fusion with transfer
significance, these operation-wise scores are rank-normalized to
$[0,1]$ as described in Sec.~3.3.

We emphasize that $\bar{q}_j$ is a practical statistic computed from
observed Rayleigh-quotient curvature responses and should be
distinguished from the theoretical quantity
$\gamma^j=\frac{1}{K}\sum_{k=1}^{K}\lambda_k^j$
in Eq.~(\ref{eq:curvature_scalarization}), where $\lambda_k^j$ denotes
an exact Hessian eigenvalue. Practical SRFT does not explicitly recover
these exact eigenvalues or eigendirections.

This construction intentionally discards within-operation directional
information and retains the relative source sensitivity across
operations. Accordingly, SRFT does not perform direction-dependent
second-order preconditioning or attempt to reconstruct the complete
Hessian geometry. Instead, it requires a compact scalar estimate of how
sensitive each operation is to perturbations of the pre-trained
objective.

To examine whether this operation-level scalarization is effective, we
compare several alternative prior-significance estimators under
identical adaptation settings. Specifically, we consider
\textbf{(i)} parameter-wise diagonal empirical Fisher,
\textbf{(ii)} layer-wise empirical Fisher,
\textbf{(iii)} the largest operation-wise curvature response, and
\textbf{(iv)} the proposed operation-wise top-$K$ curvature mean.
The empirical-Fisher variants are included as representative classical
parameter-importance estimators and are distinct from the source-loss
Hessian formulation adopted by SRFT.

\begin{table*}[t]
\centering
\renewcommand{\arraystretch}{1.0}
\setlength{\tabcolsep}{4.0pt}
\begin{tabular}{l|l|cc|ccc|cc}
\toprule
\multirow{2}{*}{\textbf{Significance Estimator}}
& \multirow{2}{*}{\textbf{Granularity}}
& \multicolumn{2}{c|}{\textbf{FE108}}
& \multicolumn{3}{c|}{\textbf{DepthTrack}}
& \multicolumn{2}{c}{\textbf{LasHeR}}\\
\cline{3-9}
&
& SR & PR
& Pr & Re & F-score
& SR & PR \\
\midrule

FFT
& None
& 65.2 & 93.5
& 61.6 & 61.4 & 61.5
& 53.2 & 65.8\\

Diagonal empirical Fisher
& Parameter-wise
& 65.9 & 94.5
& 63.1 & 63.5 & 63.3
& 54.6 & 67.8 \\

Layer-wise empirical Fisher
& Layer-wise scalar
& 65.6 & 94.1
& 62.7 & 63.1 & 62.9
& 54.2 & 67.2\\

Largest curvature response
& Operation-wise scalar
& 66.5 & 95.0
& 63.9 & 64.3 & 64.1
& 55.2 & 68.7\\

Top-$K$ curvature mean (Ours)
& Operation-wise scalar
& \textbf{66.8} & \textbf{95.4}
& \textbf{64.1} & \textbf{64.5} & \textbf{64.3}
& \textbf{55.6} & \textbf{69.0}\\

\bottomrule
\end{tabular}
\caption{
Comparison of alternative prior-significance estimators.
Diagonal and layer-wise empirical Fisher are included as classical
parameter-importance baselines, whereas the largest-response and
top-$K$ variants are constructed from the source-loss curvature used
by SRFT. All variants use the same initialization, adaptation data,
optimization settings, and evaluation protocol; all other settings
remain unchanged.
}
\label{tab:r10}
\end{table*}

As shown in Table~\ref{tab:r10}, all significance-based
variants improve over unconstrained FFT, while the operation-wise
curvature estimators provide stronger performance than the
empirical-Fisher alternatives. In particular, the
largest-curvature-response variant already remains close to the proposed
top-$K$ construction across the three benchmarks, indicating that the
dominant operation-level curvature response contains substantial
source-sensitivity information. Averaging the top-$K$ observed responses
further provides consistently better or comparable results, yielding
the strongest overall performance on FE108, DepthTrack, and LasHeR.

These results support the operation-level scalar construction adopted
by SRFT. The effectiveness of prior significance does not rely on
reconstructing the complete Hessian or preserving its eigenvectors;
rather, a compact summary of dominant observed local curvature responses
is sufficient to provide effective source-sensitivity information for
significance-guided update regulation

\subsection{Functional Sensitivity Analysis of Prior Significance}
Prior significance is constructed from operation-wise source-loss
curvature to provide a scalar sensitivity score for update-mobility
control. Specifically, we adopt the operation-wise group-diagonal
approximation
\begin{equation}
\mathcal H_{\mathrm{bd}}
=
\operatorname{blkdiag}
\left(
\mathcal H_{11},\ldots,\mathcal H_{NN}
\right),
\end{equation}
where $\mathcal H_{jj}$ denotes the source-loss Hessian block associated
with operation $j$. This approximation retains intra-operation curvature
while omitting cross-operation blocks $\mathcal H_{j\ell}$,
$j\neq\ell$; these omitted interactions are not assumed to be zero.
For a perturbation restricted to operation $j$,
\begin{equation}
\Delta\theta^\top\mathcal H\Delta\theta
=
\Delta\theta_j^\top\mathcal H_{jj}\Delta\theta_j,
\end{equation}
so $\mathcal H_{jj}$ directly characterizes the local source-loss
response to perturbations of that operation.

We distinguish functional interpretability from semantic interpretability. The proposed prior significance quantifies how perturbations to an operation affect the source tracking objective and therefore characterizes functional sensitivity. It does not imply that a high-significance operation corresponds one-to-one to a specific visual concept, since semantic functions are generally distributed across multiple layers and operations.

\noindent \textbf{Module- and layer-wise functional sensitivity.} Directly quantifying the exact discrepancy induced by all omitted
cross-operation blocks would require access to the corresponding
cross-operation curvature structure. We therefore examine the
functional sensitivity of the operation-level significance used by SRFT
through controlled perturbations.

As shown in Table~\ref{tab:r3}, we consider two interventions.
\emph{Flatten-to-mean} removes the relative significance variation
within a selected module while preserving its average regularization
strength, whereas \emph{set-to-zero} removes significance-guided
protection from the selected module or layer group. Performance
degradation closely follows the estimated significance ordering.
Under set-to-zero, FFN-2, QKV, FFN-1, and Projection decrease the
F-score by $2.6$, $2.1$, $1.8$, and $1.6$ points, respectively,
whereas the low-significance normalization operations cause
substantially smaller degradation. The layer-wise results show the same
trend, with early, middle, and late blocks decreasing the F-score by
$2.4$, $1.5$, and $1.1$ points, respectively. The flatten-to-mean
intervention exhibits a consistent pattern. These results show a clear
functional correspondence between the estimated operation-level
significance and downstream sensitivity.

\begin{table}[t]
\centering
\caption{
Module- and layer-wise prior-significance perturbation on DepthTrack.
Normalized significance is obtained by aggregating operation-level
scores within each module or layer group. ``Flatten-to-mean'' removes
within-group significance variation while preserving its mean, whereas
``set-to-zero'' removes significance-guided protection.
$\Delta$F-score is measured relative to FFT + Prior Significance.
}
\renewcommand{\arraystretch}{0.9}
\setlength{\tabcolsep}{1.0pt}
\small
\begin{tabular}{l|cc|ccc|c}
\toprule
\textbf{Perturbed Group}
& \textbf{Norm. Sig.}
& \textbf{Sig. Rank}
& \textbf{Pr}
& \textbf{Re}
& \textbf{F}
& $\boldsymbol{\Delta}$\textbf{F} \\
\midrule
FFT + Prior Significance
& -- & -- & 64.1 & 64.5 & 64.3 & -- \\
\midrule

\multicolumn{7}{c}{\textit{Module-wise flatten-to-mean}} \\
\midrule
Norm-1
& 0.070 & 6 & 63.5 & 64.3 & 63.9 & -0.4 \\ %
QKV
& 0.719 & 2 & 62.8 & 63.6 & 63.2 & -1.1 \\ %
Projection
& 0.580 & 4 & 63.0 & 63.8 & 63.4 & -0.9 \\ %
Norm-2
& 0.142 & 5 & 63.4 & 64.2 & 63.8 & -0.5 \\ %
FFN-1
& 0.627 & 3 & 62.9 & 63.7 & 63.3 & -1.0 \\ %
FFN-2
& 1.000 & 1 & 62.5 & 63.3 & 62.9 & -1.4 \\ %
\midrule

\multicolumn{7}{c}{\textit{Module-wise set-to-zero}} \\
\midrule
Norm-1
& 0.070 & 6 & 63.3 & 64.1 & 63.7 & -0.6 \\ %
QKV
& 0.719 & 2 & 61.8 & 62.6 & 62.2 & -2.1 \\ %
Projection
& 0.580 & 4 & 62.3 & 63.1 & 62.7 & -1.6 \\ %
Norm-2
& 0.142 & 5 & 63.0 & 63.8 & 63.4 & -0.9 \\ %
FFN-1
& 0.627 & 3 & 62.1 & 62.9 & 62.5 & -1.8 \\ %
FFN-2
& 1.000 & 1 & 61.3 & 62.1 & 61.7 & -2.6 \\ %
\midrule

\multicolumn{7}{c}{\textit{Layer-wise set-to-zero}} \\
\midrule
Early blocks (0--3)
& 1.000 & 1 & 61.5 & 62.3 & 61.9 & -2.4 \\ %
Middle blocks (4--7)
& 0.618 & 2 & 62.4 & 63.2 & 62.8 & -1.5 \\ %
Late blocks (8--11)
& 0.465 & 3 & 62.8 & 63.6 & 63.2 & -1.1 \\ %
\bottomrule
\end{tabular}
\label{tab:r3}
\end{table}

\noindent \textbf{Robustness to the retained-response budget $K$.} The top-$K$ statistic is used to summarize the dominant curvature
responses of each operation rather than to reconstruct its complete
curvature block. Its robustness has already been evaluated in
Sec.~4.3 and Table~5 using
\begin{equation}
K\in\{1,2,5,10,20\}.
\end{equation}
Across FE108, DepthTrack, and LasHeR, all reported metrics vary by at
most $0.4$ points, and even $K=1$ remains close to the default
$K=10$ configuration. Together with the operation-wise
curvature-response distributions in Fig.~8, these results indicate
that prior-significance estimation is insensitive to a broad range of
retained curvature-response budgets once the dominant curvature responses are
retained.

Overall, these results show that the operation-level information retained
by prior-significance estimation is functionally meaningful, and robust to
retained curvature-response budget, without requiring the approximation to reconstruct the
complete source-loss curvature geometry.

\noindent\textbf{Cross-architecture validation.}
We further examine whether the prior-significance estimator generalizes
across different tracker architectures. Using the same operation-wise
formulation and retained curvature-response budget $K=10$ without
architecture-specific modification, prior-significance regularization
consistently improves both OSTrack-B256 and SUTrack-B384.
For OSTrack-B256, it improves FE108 from 65.2/93.5 to 66.8/95.4
(SR/PR), DepthTrack from 61.5 to 64.3 F-score, and LasHeR from
53.2/65.8 to 55.6/69.0 (SR/PR). For SUTrack-B384, the corresponding
results improve from 67.1/96.0 to 69.3/97.2 on FE108, from 62.7 to
64.5 F-score on DepthTrack, and from 59.2/73.9 to 61.3/76.4 on
LasHeR. These results indicate that the practical effectiveness of the
operation-level prior-significance estimator is not restricted to a
single tracker architecture.

\subsection{Comparison with Update-Stabilization Strategies}
\label{app:update_stabilization}

Transfer significance is conceptually related to several
update-stabilization strategies, including gradient clipping, adaptive
gradient scaling, layer-wise learning-rate decay, L2-SP, and EWC.
However, these methods differ in both the signal used for update
modulation and the role of the resulting constraint.

To isolate the transfer-significance component, we conduct a controlled
comparison on DepthTrack using the pre-trained OSTrack-B256 checkpoint.
Prior significance is disabled for all variants, and all methods use the
same FFT pipeline, adaptation data, data augmentation, batch size,
training epochs, AdamW optimizer, base learning rate, weight decay,
learning-rate schedule, and evaluation protocol. Only the
update-stabilization strategy is varied. The compared methods are
implemented as follows.

\noindent\textbf{Gradient clipping.}
Let
$g_i=\nabla_{\theta}\mathcal{L}_{\mathrm{trk}}(\theta_i)$
denote the raw gradient at iteration $i$. After back-propagation and
before the AdamW moment update, we clip its global $\ell_2$ norm as
\[
\widetilde{g}_i
=
g_i
\min
\left(
1,
\frac{c}{\|g_i\|_2+\epsilon}
\right),
\qquad
c=1.0,
\]
where $\epsilon$ is used only for numerical stability. The clipped
gradient $\widetilde{g}_i$ is then passed to AdamW.

\noindent\textbf{Adaptive gradient scaling.}
For each parameter $n$, we maintain a running estimate of its gradient
magnitude,
\[
m_{n,i}
=
\beta_g m_{n,i-1}
+
(1-\beta_g)|g_{n,i}|,
\qquad
\beta_g=0.9,
\]
and rescale the current raw gradient as
\[
\widetilde{g}_{n,i}
=
\frac{g_{n,i}}
{m_{n,i}+\epsilon_g},
\qquad
\epsilon_g=10^{-6}.
\]
The rescaled gradient is subsequently passed to AdamW. This baseline
therefore performs magnitude-dependent gradient normalization without
using source-domain importance or the relative target-conditioned
significance employed by SRFT.

\noindent\textbf{Layer-wise learning-rate decay.}
We assign a fixed depth-dependent learning rate to each Transformer
block. For a backbone with $L$ blocks, the learning rate of block
$\ell\in\{0,\ldots,L-1\}$ is
\[
\alpha_{\ell}
=
\alpha_{\mathrm{base}}
\delta_{\mathrm{lr}}^{\,L-1-\ell},
\qquad
\delta_{\mathrm{lr}}=0.9,
\]
where $\alpha_{\mathrm{base}}=1\times10^{-4}$ is used for the deepest
block. Thus, earlier blocks receive progressively smaller learning
rates, while the depth-dependent schedule remains fixed throughout
adaptation.

\noindent\textbf{L2-SP.}
L2-SP~\citep{xuhong2018explicit} constrains the adapted parameters
toward their pre-trained initialization $\theta_0$ through
\[
\mathcal{L}_{\mathrm{L2\text{-}SP}}
=
\mathcal{L}_{\mathrm{trk}}
+
\lambda_{\mathrm{L2\text{-}SP}}
\|\theta-\theta_0\|_2^2,
\qquad
\lambda_{\mathrm{L2\text{-}SP}}=0.01.
\]

\noindent\textbf{EWC.}
For EWC~\citep{kirkpatrick2017overcoming}, we estimate the
parameter-wise diagonal empirical-Fisher importance from the source RGB
tracking data $D_0$ as
\[
F_n^{\mathrm{emp}}
=
\frac{1}{|D_0|}
\sum_{(x,y)\in D_0}
\left(
\frac{
\partial
\mathcal{L}_{\mathrm{trk}}(\theta_0;x,y)
}{
\partial\theta_n
}
\right)^2,
\]
and optimize
\[
\mathcal{L}_{\mathrm{EWC}}
=
\mathcal{L}_{\mathrm{trk}}
+
\lambda_{\mathrm{EWC}}
\sum_n
F_n^{\mathrm{emp}}
(\theta_n-\theta_{0,n})^2,
\qquad
\lambda_{\mathrm{EWC}}=400.
\]
Here, the empirical Fisher is used only as the classical source-importance
estimator adopted by the EWC baseline and should not be confused with
the source-loss Hessian used to formulate prior significance in SRFT.

\noindent\textbf{Transfer significance (Ours).}
Transfer significance is computed parameter-wise from the current
target-domain gradient according to Eq.~(\ref{eq:transfer_significance}).
Unlike the above baselines, it is converted into a relative
target-conditioned sensitivity and dynamically modulates parameter
update mobility during adaptation. Prior significance is disabled in
this experiment so that the comparison isolates the
transfer-significance component.

\begin{table*}[t]
\centering
\caption{
Comparison with representative update-stabilization strategies on
DepthTrack. Prior significance is disabled for all variants, and all
other adaptation and evaluation settings remain unchanged. Best and
second-best results are shown in bold and underlined, respectively.
}
\renewcommand{\arraystretch}{1.0}
\setlength{\tabcolsep}{5pt}
\begin{tabular}{l|l|ccc}
\toprule
\textbf{Method}
& \textbf{Update Criterion}
& Pr
& Re
& F-score \\
\midrule

FFT
& None
& 61.6 & 61.4 & 61.5 \\

Gradient clipping
& Global raw-gradient $\ell_2$ clipping
& 62.1 & 62.0 & 62.0 \\

Adaptive gradient scaling
& Inverse running gradient magnitude
& 62.4 & 62.3 & 62.3 \\

Layer-wise LR decay
& Fixed depth-dependent learning rate
& 62.6 & 62.7 & 62.6 \\

L2-SP
& Distance-to-initialization penalty
& 62.4 & 62.5 & 62.4 \\

EWC
& Source empirical-Fisher importance
& \underline{62.9} & \underline{63.0} & \underline{62.9} \\

Transfer significance (Ours)
& Target-conditioned parameter sensitivity
& \textbf{63.9} & \textbf{64.2} & \textbf{64.1} \\

\bottomrule
\end{tabular}
\label{tab:r13}
\end{table*}

As shown in Table~\ref{tab:r13}, all update-stabilization strategies
improve over unconstrained FFT to varying degrees. Transfer significance
achieves the strongest performance, improving FFT by $2.3$ Pr,
$2.8$ Re, and $2.6$ F-score, and exceeding the strongest alternative,
EWC, by $1.2$ F-score. The comparison indicates that its benefit is not
reproduced by global gradient clipping, magnitude-based rescaling,
fixed layer-wise learning-rate decay, or source-only parameter
anchoring.

These results support the role of transfer significance as a
target-conditioned and dynamically varying update signal. Rather than
applying a fixed threshold, predefined layer schedule, or static
source-domain importance, it adapts parameter modulation to the current
target-domain optimization state.

\begin{figure*}[t]
\begin{center}
\centerline{\includegraphics[width=0.80\textwidth]{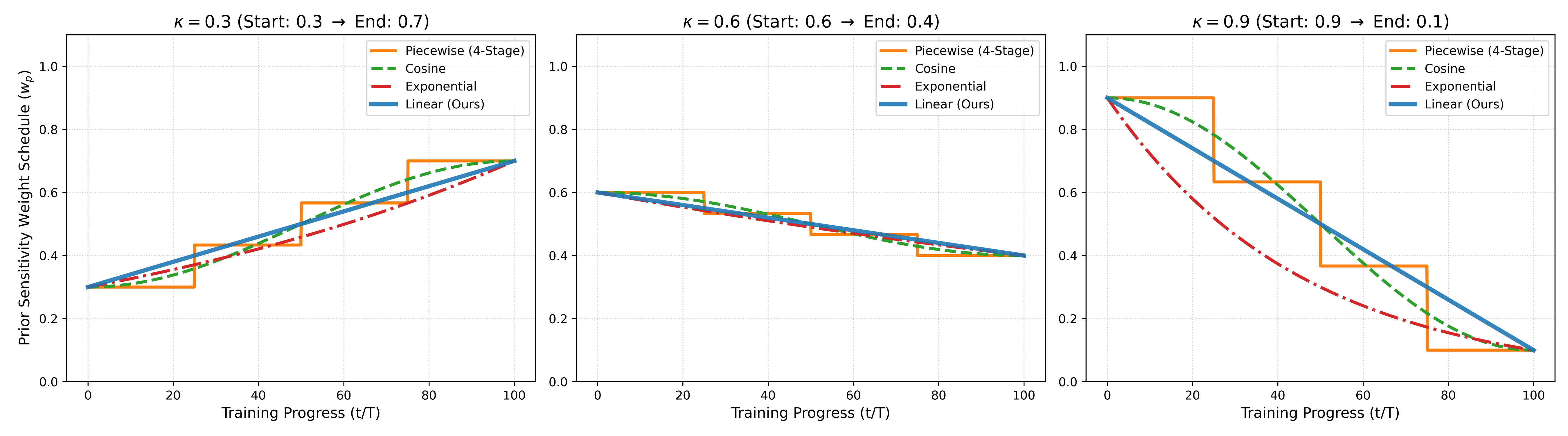}}
\vskip -0.15in
\caption{Schematic illustration of different significance fusion scheduling strategies. We compare the evolution of the prior significance fusion weight $w_p$ over the training process ($\frac{t}{T}$) using Linear (Ours), Cosine, Exponential, and Piecewise functions across varying initial significance harmony coefficient settings ($\kappa \in \{0.3, 0.6, 0.9\}$).}
\label{fig:significance_schedule}
\vskip -0.2in
\end{center}
\end{figure*}

\begin{table*}[thbp]
\centering
\vskip -0.1in
\caption{Comparison of different significance fusion schedules. We compare the prior significance fusion weight $w_p$ over the training process ($\frac{t}{T}$) with Linear (Ours), Cosine, Exponential, and Piecewise functions across varying initial significance harmony coefficient settings ($\kappa \in \{0.3, 0.6, 0.9\}$). \textbf{“PS”} represents prior significance; \textbf{“TS”} is transfer significance.}
\renewcommand{\arraystretch}{1.0}
\setlength{\tabcolsep}{5.0pt}
\label{tab:schedule_ablation}
\begin{tabular}{c|c|c|c|cc|ccc|cc}
\toprule
\textbf{Schedule} & \textbf{PS Weight} & \textbf{TS Weight} &\textbf{$\kappa$} & \textbf{SR} & \textbf{PR} &\textbf{Pr}  &\textbf{Re} &\textbf{F-score}  &\textbf{SR} &\textbf{PR}  \\
\midrule
\multirow{3}{*}{Exponential} & \multirow{3}{*}{$w^p = \kappa \left( \frac{1-\kappa}{\kappa} \right)^{\frac{t}{T}}$} & \multirow{3}{*}{$w^t = 1-w^p$} & 0.3 &65.7 &94.2 &63.8 &64.5  &64.1 &55.4 &68.5  \\
&  & & 0.6 & 67.2 & 96.3 &64.5 &64.9  &64.7 &55.9 &69.7 \\
& &  & 0.9 &66.4 &94.1 &63.7 &64.5  &64.1 &55.5 &68.7 \\
\midrule
\multirow{3}{*}{Cosine} &\multirow{3}{*}{$w^p = (1-\kappa) + \frac{2\kappa - 1}{2} \left( 1 + \cos \left( \frac{t \pi}{T} \right) \right)$} & \multirow{3}{*}{$w^t = 1-w^p$} &0.3 &66.6 &95.0 &63.8 &64.2  &64.0 &55.1 &68.7 \\
& &  & 0.6 & 67.3 & 96.4 &64.6 &65.1  &64.8 &56.1 &69.9 \\
& &  & 0.9 & 66.7 & 95.2 &64.1 &65.0  &64.5 &55.5 &69.1 \\
\midrule
\multirow{3}{*}{\shortstack{Piecewise\\(4-Stage)}} & \multirow{3}{*}{$w_i^p
=
\kappa
+
\frac{1-2\kappa}{3}\left\lfloor\frac{4t}{T}\right\rfloor$} & \multirow{3}{*}{$w^t = 1-w^p$} 
  & 0.3 &66.8 &94.9 &63.8 &64.0  &63.9 &55.0 &68.5 \\
& & & 0.6 &67.1 &96.0 &64.6 &64.8  &64.7 &56.0 &69.9 \\
& & & 0.9 &66.5 &95.3 &64.2 &64.7  &64.4 &55.7 &69.2 \\
\midrule
\multirow{3}{*}{\textbf{Linear (Ours)}} &\multirow{3}{*}{$w^p = \kappa+(1-2\kappa)\frac{t}{T}$} & \multirow{3}{*}{$w^t = 1-w^p$} & 0.3 & 66.9 & 95.1 &63.9 &64.2  &64.0 &55.3 &68.9 \\
& & &0.6 & \textbf{67.4} & \textbf{96.5} &\textbf{64.7} &\textbf{65.4}  &\textbf{65.1} &\textbf{56.3} &\textbf{70.1}\\
& &  & 0.9 & 66.9 & 95.5 &64.7 &64.6  &64.7 &55.8 &69.4 \\
\bottomrule
\end{tabular}
\end{table*}

\subsection{Robustness of Transfer-Significance Estimation}
\label{app:ts_robustness}
Transfer significance is designed to characterize the current local optimization sensitivity of each parameter. Accordingly, SRFT uses the instantaneous squared gradient $g_i^2$,
which responds directly to changes in the current optimization state.
Since transfer significance is subsequently ranked and normalized,
the regularization depends primarily on the relative ordering of
parameter sensitivities rather than their absolute gradient magnitudes.

We further examine the robustness of this estimation under different
temporal gradient statistics and optimization configurations. Unless
otherwise specified, only the factor under investigation is varied,
while all other experimental settings remain unchanged.

\begin{figure}[t]
\centering
\includegraphics[clip,width=0.48\textwidth]
{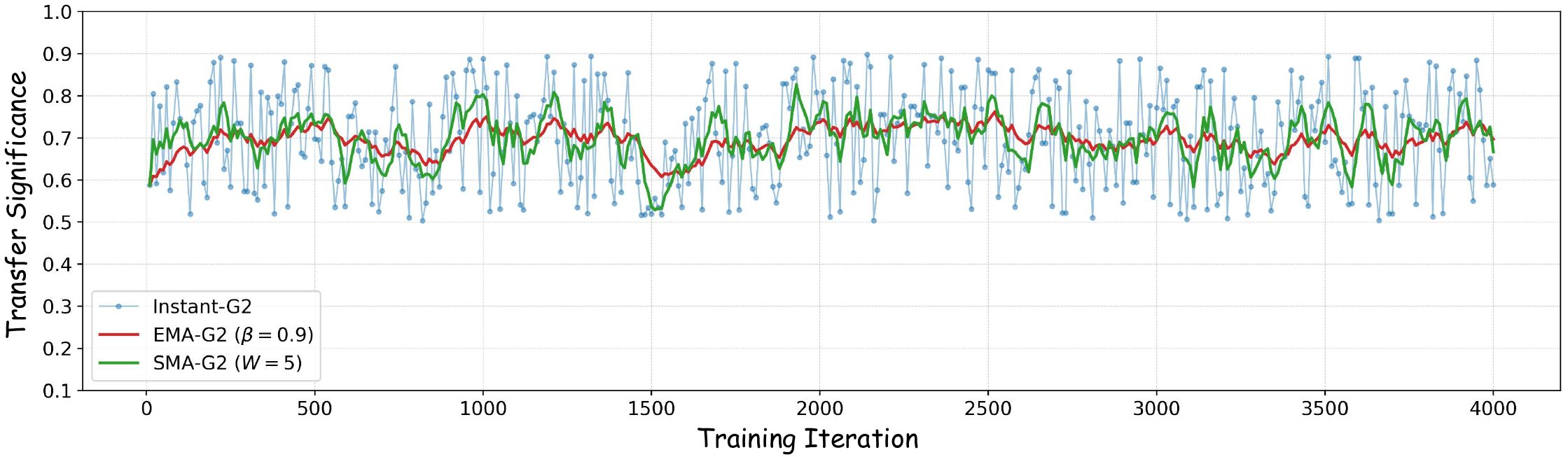}
\vskip -0.15in
\caption{Representative normalized transfer-significance trajectory of Instant-G2, EMA-G2 ($\beta=0.9$), and SMA-G2 ($W=5$) on DepthTrack using OSTrack-B256.}
\vskip -0.10in
\label{fig:r2}
\end{figure}


\noindent\textbf{(1) Temporal gradient statistics.}
We compare the instantaneous estimator with two commonly used temporal
statistics:
\begin{align*}
\textbf{Instant-G2:}\qquad
&s_i^{t}=g_i^2,\\
\textbf{EMA-G2:}\qquad
&s_i^{t}
=
\beta s_{i-1}^{t}
+
(1-\beta)g_i^2,\\
\textbf{SMA-G2:}\qquad
&s_i^{t}
=
\frac{1}{W}
\sum_{k=0}^{W-1}g_{i-k}^{2}.
\end{align*}
EMA-G2 uses $\beta=0.9$, while SMA-G2 adopts a window of $W=5$
optimization steps. All estimators are updated once per optimizer step.

As illustrated in Fig.~\ref{fig:r2}, all three estimators follow a
similar low-frequency trend but exhibit different temporal responses.
Instant-G2 reacts directly to local changes, whereas EMA-G2 and SMA-G2
suppress short-term fluctuations and introduce different degrees of
temporal smoothing.

As reported in Table~\ref{tab:r5}, SMA-G2 performs comparably to
Instant-G2, with differences of at most 0.2 points across all reported
metrics. EMA-G2 with $\beta=0.9$ shows a small degradation of
0.1--0.4 points. These results indicate that transfer-significance
estimation is robust to moderate temporal aggregation, while stronger
historical smoothing does not provide additional performance benefits
and may reduce responsiveness to changes in the current optimization
state. We therefore retain Instant-G2 as the default estimator because
of its competitive accuracy, immediate response, and absence of
additional history-state storage.
\begin{table}[thbp]
\centering
\vskip -0.15in
\caption{Comparison of different temporal gradient statistics.}
\vskip -0.1in
\renewcommand{\arraystretch}{0.9}
\setlength{\tabcolsep}{4.0pt}
\small
\begin{tabular}{c | c c| c c c |c c}
\toprule
\multirow{2}{*}{\textbf{Exp.}}
& \multicolumn{2}{c|}{FE108}
& \multicolumn{3}{c|}{DepthTrack}
& \multicolumn{2}{c}{LasHeR} \\
\cline{2-8}
\multicolumn{1}{c|}{}
& SR & PR & Pr & Re & F-score & SR & PR \\
\hline
\textbf{FFT}
& 65.2 & 93.5 & 61.6 & 61.4 & 61.5 & 53.2 & 65.8 \\
\textbf{Instant-G2}
& \textbf{66.6} & \textbf{94.9}
& \textbf{63.9} & \textbf{64.2} & \textbf{64.1}
& \textbf{55.1} & \textbf{68.4} \\
\textbf{EMA-G2 ($\beta=0.9$)}
& 66.4 & 94.5 & 63.6 & 64.0 & 63.8 & 55.0 & 68.2 \\
\textbf{SMA-G2 ($W=5$)}
& 66.5 & \textbf{94.9}
& \textbf{63.9} & 64.0 & 63.9 & 55.0 & 68.3 \\
\bottomrule
\end{tabular}
\label{tab:r5}
\vskip -0.1in
\end{table}


\noindent\textbf{(2) Effect of batch size.}
To examine sensitivity to the global batch configuration, we vary the
batch size over $\{64,96,128,192\}$ while keeping all other settings
unchanged. As shown in Table~\ref{tab:r6}, the metric ranges across all batch
sizes are only 0.2--0.3 points. Using batch size 192 as the reference,
the largest decrease is 0.3 points on DepthTrack Re, while the largest
increase is 0.1 points on FE108 SR. These results indicate practical
robustness to the tested batch-size range. Although changing batch size
affects gradient magnitude and stochastic variance, the narrow
performance variation suggests that the relative parameter-sensitivity
ordering used by SRFT remains stable.
\begin{figure}[t]
\centering
\includegraphics[clip,width=0.48\textwidth]
{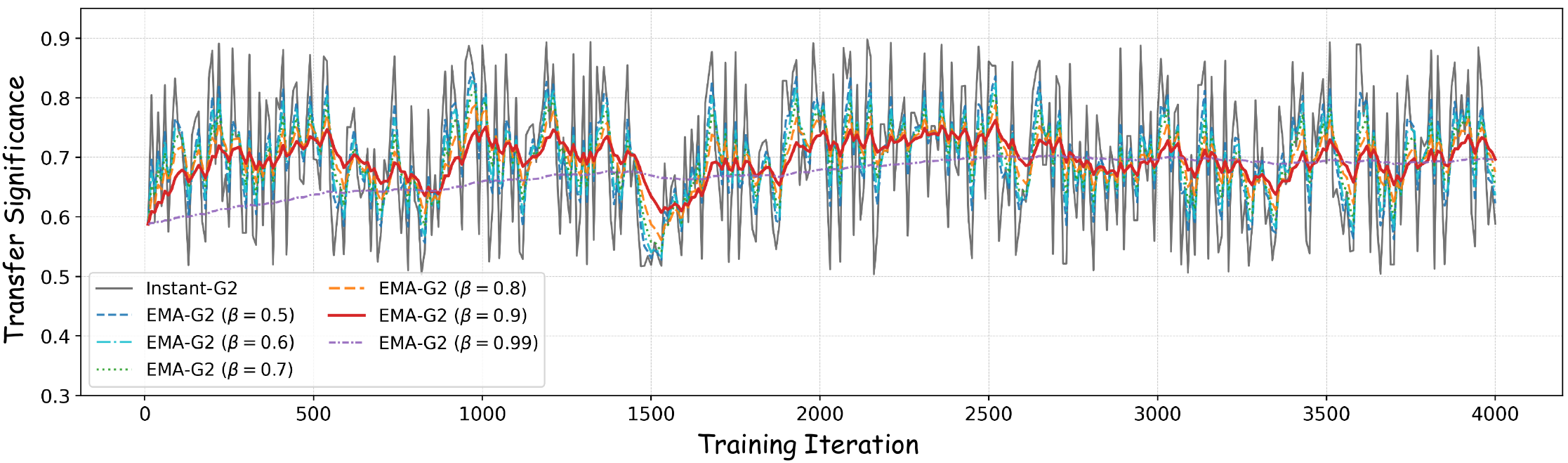}
\vskip -0.15in
\caption{Representative normalized transfer-significance trajectories under different EMA-G2 smoothing coefficients on DepthTrack using OSTrack-B256. }
\vskip -0.2in
\label{fig:r3}
\end{figure}

\begin{table}[thbp]
\centering
\caption{Robustness to global batch size.}
\vskip -0.1in
\renewcommand{\arraystretch}{0.9}
\setlength{\tabcolsep}{5.0pt}
\small
\begin{tabular}{c | c c| c c c |c c}
\toprule
\multirow{2}{*}{\textbf{Batch Size}}
& \multicolumn{2}{c|}{FE108}
& \multicolumn{3}{c|}{DepthTrack}
& \multicolumn{2}{c}{LasHeR} \\
\cline{2-8}
\multicolumn{1}{c|}{}
& SR & PR & Pr & Re & F-score & SR & PR \\
\hline
\textbf{64}
& 66.4 & 94.7 & 63.7 & 63.9 & 63.8 & 54.9 & 68.2 \\
\textbf{96}
& 66.6 & 94.8 & 64.0 & 64.1 & 64.0 & 55.0 & 68.3 \\
\textbf{128}
& \textbf{66.7} & 94.9 & 63.8 & 64.1 & 64.0
& \textbf{55.1} & \textbf{68.4} \\
\textbf{192}
& 66.6 & \textbf{94.9} & \textbf{63.9} & \textbf{64.2}
& \textbf{64.1} & \textbf{55.1} & \textbf{68.4} \\
\bottomrule
\end{tabular}
\label{tab:r6}
\vskip -0.15in
\end{table}

\noindent\textbf{(3) Effect of gradient accumulation.}
We further isolate the influence of micro-batch partitioning by fixing
the effective batch size at $192$ and evaluating
$192\times1$, $96\times2$, $64\times3$, and $48\times4$, where the
first number denotes the micro-batch size and the second denotes the
number of accumulation steps. Transfer significance is computed from
the accumulated gradient immediately before each optimizer update. As shown in Table~\ref{tab:r7}, all reported metrics vary by at most 0.2 points. This indicates that transfer-significance estimation
is largely insensitive to how a fixed effective batch is partitioned
across micro-batches, and is primarily determined by the effective
gradient used for the optimizer update.
\begin{table}[thbp]
\centering
\vskip -0.15in
\caption{Robustness to micro-batch partitioning at a fixed effective
batch size of 192.}
\vskip -0.1in
\renewcommand{\arraystretch}{0.9}
\setlength{\tabcolsep}{3.0pt}
\small
\begin{tabular}{c | c c| c c c |c c}
\toprule
\multirow{2}{*}{\textbf{Batch $\times$ Accum.}}
& \multicolumn{2}{c|}{FE108}
& \multicolumn{3}{c|}{DepthTrack}
& \multicolumn{2}{c}{LasHeR} \\
\cline{2-8}
\multicolumn{1}{c|}{}
& SR & PR & Pr & Re & F-score & SR & PR \\
\hline
\textbf{48$\times$4}
& 66.6 & 94.8 & 64.0 & 64.0 & 64.0 & 55.1 & 68.4 \\
\textbf{64$\times$3}
& 66.6 & 94.9 & 63.8 & 64.2 & 64.1 & 55.0 & 68.3 \\
\textbf{96$\times$2}
& 66.5 & 94.9 & 64.0 & 64.1 & 64.0 & 55.2 & 68.4 \\
\textbf{192$\times$1}
& 66.6 & 94.9 & 63.9 & 64.2 & 64.1 & 55.1 & 68.4 \\
\bottomrule
\end{tabular}
\label{tab:r7}
\vskip -0.15in
\end{table}

\noindent\textbf{(4) Effect of the EMA smoothing coefficient.}
We further evaluate EMA-G2 with
$\beta\in\{0.0,0.5,0.6,0.7,0.8,0.9,0.99\}$, where $\beta=0$
corresponds to Instant-G2. As shown in Table~\ref{tab:r8}, performance remains stable for
$\beta\in[0,0.8]$, with only minor variations across the three
benchmarks. Although $\beta=0.6$ obtains the best numerical value on
several metrics, its difference from Instant-G2 is marginal. In
contrast, $\beta=0.99$ causes a clear degradation, suggesting that
excessive temporal smoothing introduces a substantial lag in the
significance estimate and weakens its responsiveness to the current
optimization state. The representative trajectories in
Fig.~\ref{fig:r3} exhibit the same trend.
\begin{table}[thbp]
\centering
\caption{Sensitivity to the EMA-G2 smoothing coefficient.}
\vskip -0.1in
\renewcommand{\arraystretch}{0.9}
\setlength{\tabcolsep}{3.5pt}
\small
\begin{tabular}{c | c c| c c c |c c}
\toprule
\multirow{2}{*}{\textbf{$\beta$}}
& \multicolumn{2}{c|}{FE108}
& \multicolumn{3}{c|}{DepthTrack}
& \multicolumn{2}{c}{LasHeR} \\
\cline{2-8}
\multicolumn{1}{c|}{}
& SR & PR & Pr & Re & F-score & SR & PR \\
\hline
\textbf{0.0 (Instant-G2)}
& 66.6 & \textbf{94.9} & 63.9 & \textbf{64.2}
& \textbf{64.1} & 55.1 & 68.4 \\
\textbf{0.5}
& 66.4 & 94.6 & 63.9 & \textbf{64.2}
& 64.0 & 55.1 & 68.4 \\
\textbf{0.6}
& 66.6 & \textbf{94.9} & \textbf{64.1} & 64.0
& \textbf{64.1} & \textbf{55.2} & \textbf{68.5} \\
\textbf{0.7}
& \textbf{66.7} & 94.8 & 64.0 & 64.1
& 64.0 & 55.1 & 68.4 \\
\textbf{0.8}
& 66.6 & 94.9 & 63.8 & 64.2
& 64.1 & 55.1 & 68.4 \\
\textbf{0.9}
& 66.4 & 94.5 & 63.6 & 64.0
& 63.8 & 55.0 & 68.2 \\
\textbf{0.99}
& 66.0 & 94.3 & 62.3 & 62.5
& 62.4 & 54.3 & 67.1 \\
\bottomrule
\end{tabular}
\label{tab:r8}
\vskip -0.15in
\end{table}
Overall, transfer-significance estimation is robust to moderate temporal aggregation, global batch size shifts, and gradient accumulation setups. However, excessive temporal smoothing degrades responsiveness to current optimization states. We thus default to Instant-G2 for its competitive accuracy, immediate response, and zero history-state overhead.

\begin{figure}[thbp]
\centering
\includegraphics[clip, width=0.48\textwidth]{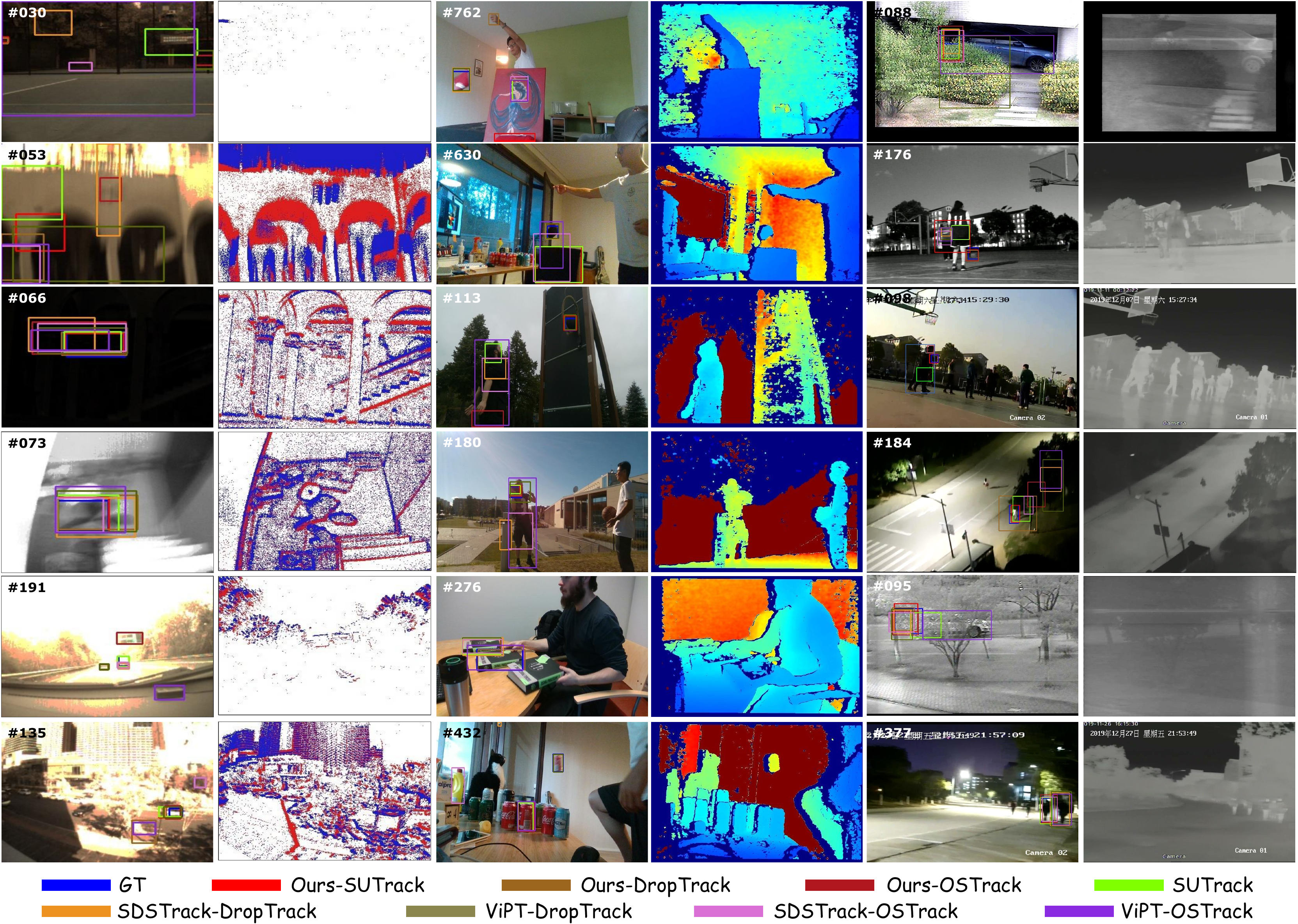}
\vskip -0.1in
\caption{Visual comparisons of the tracking performance of different methods on the (\textbf{Left}) RGB-Event, (\textbf{Middle}) RGB-Depth and (\textbf{Right}) RGB-Thermal datasets.}
\label{fig:visual}
\vskip -0.2in
\end{figure}

\vspace{-0.4cm}
\subsection{Different Significance Fusion Scheduling Strategies}
\label{sec:sign_fusion}
Our design is motivated by the stability–plasticity trade-off in cross-domain fine-tuning: effective transfer requires balancing stability (preserving critical pre-trained knowledge) and plasticity (adapting to the target domain). In the paper, we explicitly define a dynamic linear schedule to harmonize the two significance terms: at the beginning of training, the prior-significance weight is $\kappa$ and the transfer-significance weight is $1-\kappa$. As training progresses, the prior weight decreases linearly to $1-\kappa$, while the transfer weight increases to $\kappa$. This is formalized by Eq. (14), where the combined significance is a convex combination with linear dependence on training progress $\frac{t}{T}$. 

As summarized in the Fig.~\ref{fig:significance_schedule} and Tab.~\ref{tab:schedule_ablation}, we additionally compare exponential, cosine, piecewise (4-stage), and linear (ours) schedules on FE108, DepthTrack, and LasHeR datasets under significance harmony coefficient $\kappa \in \{0.3, 0.6, 0.9\}$. To ensure a fair comparison, all schedules are constrained to share the
same initial and final significance weights, differing only in how the
weights evolve during training. Overall, the linear schedule is consistently best or tied-best. When $\kappa=0.6$ (our default), both the weight evolution and the final performance are very close across different schedules, indicating the schedule choice is not overly sensitive in this moderate regime. However, when $\kappa$ becomes too small or too large, the alternative schedules tend to be more fragile. Mechanistically, exponential/cosine may over-emphasize one significance at certain phases (e.g., too aggressive early transfer or overly conservative late adaptation), whereas piecewise scheduling introduces abrupt weight jumps that can destabilize optimization. In contrast, the linear schedule provides a monotonic and smooth transition with a constant rate of change, which empirically leads to more stable and robust behavior when $\kappa$ deviates from the moderate setting.

\vspace{-0.4cm}
\section{Visual Comparison} \label{sec:a-4}
Some representative visualization results are illustrated in Fig.~\ref{fig:visual}. For clarity, all visual instances are explicitly labeled by the predicted bounding box. Horizontally, each instance is represented by a pair of pictures - the right image shows the main view, while the left displays the auxiliary modality. Our method demonstrates a marked improvement over state-of-the-art multi-modality methods, especially in handling complex scenarios where cross-modal integration plays a crucial role. It excels in tracking in conditions like motion blur, occlusions, and low illumination, where other methods struggle. These results demonstrate that our method improves tracking accuracy and provides a more stable and efficient way to integrate auxiliary data.

\end{document}